\documentclass[lettersize,journal]{IEEEtran}

\usepackage{amsmath}
\usepackage{amssymb}
\usepackage{amsfonts}
\usepackage{url}

\usepackage{xcolor}
\usepackage{adjustbox}
\usepackage{colortbl}
\usepackage{tabularx}
\usepackage{booktabs} 
\usepackage{graphicx}
\usepackage{subfig}
\usepackage[linesnumbered, vlined, ruled]{algorithm2e}
\SetArgSty{textup}
\usepackage{enumerate}
\usepackage{tcolorbox}

\usepackage{color, colortbl}
\usepackage{braket}
\definecolor{MyHiLiRow}{gray}{0.9}

\usepackage{seqsplit}
\usepackage[whole]{bxcjkjatype}

\usepackage{nameref}
\usepackage{xr}
\newcommand{\argmax}{\mathop{\rm argmax}\limits}
\newcommand{\argmin}{\mathop{\rm argmin}\limits}

\newcommand\tabblue[1]{{\color[HTML]{1f77b4}{#1}}}
\newcommand\taborange[1]{{\color[HTML]{ff7f0e}{#1}}}

\newcommand{\Enote}[2]{$#1$e$#2$}
\newcommand{\en}[2]{$#1$e$#2$}
\newcommand{\PSsymbol}{~\makebox[1em]{\texttt{+}}~}
\newcommand{\MSsymbol}{~\makebox[1em]{\texttt{-}}~}
\newcommand{\EQsymbol}{~\makebox[1em]{$\approx$}~}
\newcommand{\sgnPS}{$\phantom{+}$}
\newcommand{\sgnMS}{$-$}
\definecolor{c1}{RGB}{150,150,150}
\definecolor{c2}{RGB}{220,220,220}
\definecolor{c3}{RGB}{255,255,255}
\definecolor{c4}{RGB}{255,255,255}

\usepackage{lscape}
\usepackage{pdflscape}
\usepackage{mathtools}

\begin{document}

%\title{A Sample Article Using IEEEtran.cls\\ for IEEE Journals and Transactions}
\title{Speeding up Local Search for the Indicator-based Subset Selection Problem by a Candidate List Strategy}

\author{
Keisuke~Korogi,~and~Ryoji~Tanabe,~\IEEEmembership{Member,~IEEE}
\thanks{K. Korogi is with Graduate School of Environment and Information Sciences, Yokohama National University, Yokohama, Japan. (e-mail: keisuke.korogi.52@gmail.com).}
\thanks{R. Tanabe is with Faculty of Environment and Information Sciences, Yokohama National University, Yokohama, Japan. (e-mail: rt.ryoji.tanabe@gmail.com).}
}





\maketitle

\begin{abstract}

In evolutionary multi-objective optimization, the indicator-based subset selection problem involves finding a subset of points that maximizes a given quality indicator.
Local search is an effective approach for obtaining a high-quality subset in this problem.
However, local search requires high computational cost, especially as the size of the point set and the number of objectives increase.
To address this issue, this paper proposes a candidate list strategy for local search in the indicator-based subset selection problem.
In the proposed strategy, each point in a given point set has a candidate list.
%Each point can be swapped only with unselected points in its associated candidate list.
During search, each point is only eligible to swap with unselected points in its associated candidate list.
This restriction drastically reduces the number of swaps at each iteration of local search.
We consider two types of candidate lists: nearest neighbor and random neighbor lists.
This paper investigates the effectiveness of the proposed candidate list strategy on various Pareto fronts.
The results show that the proposed strategy with the nearest neighbor list can significantly speed up local search on continuous Pareto fronts without significantly compromising the subset quality.
The results also show that the sequential use of the two lists can address the discontinuity of Pareto fronts.

\end{abstract}


%\section{Introduction}
%\IEEEPARstart{T}{his} file is intended to serve as a ``sample article file''

\section{Introduction}
\label{sec:introduction}

%\noindent \textit{General context.}
%
%\IEEEPARstart{T}{his} paper considers multi-objective optimization that simultaneously minimizes $d$ objective functions $f_1, \dots, f_d$.
\IEEEPARstart{M}{ulti-objective} optimization aims to simultaneously minimize $d$ objective functions $f_1, \dots, f_d$.
In general, no absolute optimal solution can minimize all $d$ objective functions.
Thus, the ultimate goal of multi-objective optimization is to find a Pareto optimal solution preferred by a decision maker \cite{Miettinen98}.
An a posteriori decision making is generally performed when the decision maker's preference information is not available, where she/he selects a single solution from a non-dominated solution set.
Evolutionary multi-objective optimization (EMO) \cite{Deb01} is an effective approach for obtaining a non-dominated solution set that approximates the Pareto front (PF) in the objective space.
For simplicity, this paper denotes a $d$-dimensional objective vector $f(x)$ of a solution $x$ as a point $p$.
This paper also considers only the objective space $V \subseteq \mathbb{R}^d$.

%$d$-dimensional objective space $V \subseteq \mathbb{R}^d$, 
% Given a set of $n$ non-dominated points $P \subseteq V$
Given a non-dominated point set $P \subseteq V$ of size $n$, a quality indicator $\mathcal{I}$, and a positive integer $k$, the indicator-based subset selection problem (ISSP) \cite{BasseurDGL16} aims to find a subset $S^* \subset P$ of size $k$ that maximizes $\mathcal{I}$, where $k < n$.
As discussed in \cite{BasseurDGL16,Falcon-CardonaC20}, the ISSP can be generally found in  the context of EMO. %, including postprocessing of an unbounded external archive and environmental selection in an EMO algorithm. 
For example, previous studies \cite{Lopez-IbanezKL11,BringmannFK14ppsn,IshibuchiSMN16} reported the effectiveness of an unbounded external archive that maintains all non-dominated solutions found so far.
Although the size of the unbounded external archive is generally large at the end of the search, it is difficult for the decision maker to examine such a large number of solutions.
Thus, it is necessary to select a small number of representative solutions from the archive for her/him.
This postprocessing procedure is identical to the ISSP.
% In addition, environmental selection in indicator-based EMO algorithms (e.g., IBEA \cite{ZitzlerK04} and HypE \cite{BaderZ11}) can be considered as the ISSP \cite{Falcon-CardonaC20}. 
%
In addition, the ISSP appears in environmental selection in indicator-based EMO algorithms \cite{Falcon-CardonaC20}.
Let $P$ and $Q$ be the population and offspring, respectively.
Let also $k$ and $n$ be the sizes of $P$ and the union $P \cup Q$, respectively.
In environmental selection in indicator-based EMO algorithms (e.g., HypE \cite{BaderZ11}),\footnote{Environmental selection in decomposition-based EMO algorithms~\cite{Li25} is based on scalarizing functions and generally does not use any quality indicator.
Thus, the ISSP does not appear in decomposition-based EMO algorithms, and their selection methods (e.g., \cite{LiZKLW14,WuLKZZ17}) cannot be applied to the ISSP.} the best $k$ individuals in terms of a quality indicator are needed to be selected from $n$ individuals in $P \cup Q$.
This is exactly the same as the ISSP.
Any subset selection method for the ISSP can be incorporated into indicator-based EMO algorithms in a plug-in manner.
Thus, designing an effective subset selection method is beneficial from the perspective of indicator-based EMO algorithms.
%

%(e.g., IBEA \cite{ZitzlerK04} and HypE \cite{BaderZ11}). 

Representative inexact approaches for the ISSP include  genetic algorithm  \cite{IshibuchiSMN16,IshibuchiSTN09}, local search (LS) \cite{BasseurDGL16,BradstreetBW06,NanSIH21,NanSIH23}, and greedy search \cite{BasseurDGL16,GuerreiroFP16,ShangIC21,ChenIS22}.
The computation of a quality indicator $\mathcal{I}$ is generally expensive as the number of objectives $d$ and the subset size $k$ increase.
For this reason, the computationally cheap greedy search approach is the most popular in the ISSP.
In fact, Basseur et al. \cite{BasseurDGL16} investigated the performance of LS but focused only on small-size ISSP instances with $n \leq 1\,000$.
%Bradstreet et al. \cite{BradstreetBW06} reported the poor performance of local search for a large $k$ mainly due to its high computational cost.
Bradstreet et al. \cite{BradstreetBW06} reported that LS performs poorly for a large $k$ due to its high computational cost.
%Nan et al. \cite{NanSIH21} also terminated the LS procedure early at $1\,000$ iterations to avoid the time-consuming process.
Nan et al. \cite{NanSIH21} terminated the LS procedure early at $1\,000$ iterations to avoid a time-consuming process.
In \cite{NanSIH23}, they also terminated the LS procedure when the hypervolume improvement falls below a predetermined threshold.
In both cases, there is no guarantee that LS finds a local optimal subset.

%In fact, the previous study \cite{NanSIH21} reported that LS performs worse than greedy search in terms of both the approximation performance and the computational cost.

%\noindent \textit{Motivation.}
%\cite{BradstreetBW06,BasseurDGL16} .

%For the sake of speeding up LS, two approaches can be considered.
However, as demonstrated in \cite{BasseurDGL16,BradstreetBW06}, LS for the ISSP requires high computational cost but can find a better subset than greedy search.
Thus, an effective inexact method for the ISSP can be designed by reducing the computational cost in LS.
Roughly speaking, two approaches can be considered to speed up LS.
One is to improve a method for evaluating a solution. %, i.e., a subset in the ISSP.
Iterative stochastic search methods (including LS) require many solution evaluations.
If the evaluation of a solution is time-consuming, it incurs a high computational cost.
For example, the computation of the hypervolume indicator \cite{ZitzlerT98} is expensive as the number of objectives $d$ increases.
To address this issue, Bradstreet et al. \cite{BradstreetWB07} proposed a fast method for computing the hypervolume of a subset $S$ in LS.
Shang et al. \cite{ShangIC21} also proposed the use of the R2 indicator \cite{HansenJ98,ShangIZL18} instead of the hypervolume indicator.\footnote{Strictly speaking, they addressed the ISSP using the R2 indicator, not the ISSP using the hypervolume indicator.}
Here, the R2 indicator is computationally cheaper than the hypervolume indicator.

The other is to reduce the neighborhood size in LS.
In other words, this approach reduces the number of candidate solutions to be evaluated by an objective function for each iteration. 
A candidate list strategy \cite{HoosS2004} is one of the most classical reduction approaches in the context of combinatorial optimization.
As an example, let us consider LS with 2-opt moves in the travelling salesperson problem (TSP).
Note that we below describe the candidate list strategy for the TSP just to explain the working concept of the candidate list strategy.
Note also that it is impossible to re-use the candidate list strategy for the TSP on the ISSP in a straightforward manner.
%However, it is not possible to apply the candidate list strategy for the TSP directly to the ISSP.
%
In the TSP, the simplest candidate list for each city includes its $l$ nearest cities in terms of the Euclidean distance, where $l$ is typically 10--40.
%Altough some types of candidate lists have been proposed for the TSP  \cite{Helsgaun00}, the simplest one includes 
Although LS scans the 2-opt neighborhood of the current tour for each iteration, the neighborhood size becomes significantly large as the number of cities increases.
%Fortunately, it has been empirically observed that an edge included in the optimal tour almost always connects two nodes close to each other \cite{Helsgaun00}.
%
Here, it has been empirically observed that exchanging an edge connecting  two nodes far from each other is unlikely to improve the tour length \cite{HoosS2004}.
For this reason, even if LS restricts the neighborhood to candidate lists, it does not significantly affect the quality of tours.

%If the ISSP shares the same property with the TSP, a candidate list strategy can reduce the number of examined solutions that do not lead to improvements.

While the first approach to speed up LS has been studied for the ISSP as described above, the second approach reducing the neighborhood size has attracted less attention in the context of the ISSP.
LS for the ISSP could potentially be sped up by not examining ``unpromising subsets" that do not lead to improvements.
However, how to detect such ``unpromising subsets" is not straightforward in the ISSP.
A subset (i.e., a solution in the ISSP) is generally represented by a 0-1 vector.
Each position in the binary vector indicates whether the corresponding point is selected.
Unfortunately, this binary vector itself does not provide a clue for restricting the neighborhood.

%\noindent \textit{Contribution.}
%(candidate list strategy) 
%
Motivated by the above discussion, first, this paper analyzes the property of the ISSP to demonstrate the rationale for the neighborhood restriction.
Then, based on the analysis, this paper proposes a candidate list strategy to speed up LS in the ISSP. 
Let $l$ be a size of the candidate list.
For each point $p$ in the current subset $S$, the proposed candidate list strategy swaps only $l$ unselected points in the candidate list of $p$.
First-improvement LS with the 2-swap operator needs to examine $k (n-k)$ subsets at each iteration in the worst case \cite{BasseurDGL16}.
In contrast, the proposed candidate list strategy can reduce this to $kl$ subsets.
%Thus, the proposed candidate list strategy can speed up LS by not performing unpromising swaps. 
Thus, the proposed candidate list strategy can speed up LS by saving unpromising swaps. 
Compared to the above-mentioned existing methods in the first approach, the advantage of the proposed candidate list strategy is generality with respect to quality indicators.
For example, the methods \cite{ShangIC21,BradstreetWB07} can be applied to only the ISSP using the hypervolume indicator.
In contrast, the proposed candidate list strategy can be applied to the ISSP using any quality indicator.
%While they require quality indicator-specific 

We consider two types of candidate lists: nearest neighbor and random neighbor lists.
The concept of the nearest neighbor list is the same as that in the TSP.
For each point $p$, its nearest neighbor list includes $l$ nearest points in the objective space.
In contrast, the random list includes $l$ randomly selected points.
We point out that LS with the nearest neighbor list does not work well on discontinuous PFs, e.g., the PF of DTLZ7 \cite{DebTLZ05}.
To address this issue, we propose a two-phase candidate list strategy that sequentially uses the random and nearest neighbor lists.
We investigate the effectiveness of the proposed candidate list strategy on the ISSP of seven quality indicators, including the hypervolume \cite{ZitzlerT98} indicator. %.various PFs

%\noindent \textit{Outline.}
%
The rest of this paper is organized as follows. 
Section \ref{sec:preliminaries} provides some preliminaries.
Section \ref{sec:analysis_cand_list} analyzes the property of the ISSP, where we focus on the distance between points for a successful swap.
Section \ref{sec:proposed_method} proposes the candidate list strategy.
Section \ref{sec:setting} describes the experimental setup.
Section \ref{sec:results} shows analysis results.
%Finally, 
Section \ref{sec:conclusion} concludes this paper.

% \noindent \textit{Code availability.}
% % %
% If this paper is accepted, we will upload the code used in this work to GitHub.
% The code of LS with the proposed candidate list strategy is available at \url{https://github.com/rogi52/issp_localsearch}

\section{Preliminaries}
\label{sec:preliminaries}

\subsection{Multi-objective optimization}
\label{sec:mo}

Here, we consider the minimization of $d$ objective functions $f = (f_1, \dots, f_d)$.
$V \subseteq \mathbb{R}^d$ represents a $d$-dimensional objective space.
As described in Section \ref{sec:introduction}, we denote an objective vector $f(x)$ as a point $p$, i.e., $p=f(x)$.

%Let us consider two points $p$ and $q$ in $V$.
A point $p$ is said to dominate another point $q$ if $p_i \leq q_i$ for all $i \in \{1, \dots, d\}$ and $p_i < q_i$ for at least one index $i$.
We denote this Pareto dominance relation as $p \prec q$.
In addition, $p$ is said to weakly dominate $q$ if $p_i \leq q_i$ for all $i \in \{1, \dots, d\}$.
If $p^*\in V$ is not dominated by any point in $V$, $p^*$ is called a Pareto optimal point.
The set of all Pareto optimal points $\{p^* \in  V \,|\, \nexists p \in  V: p \prec p^* \}$ is called the PF.
%An EMO algorithm aims to a solution set that approximates the PF in $V$.

%\subsection{Indicators in multi-objective optimization}
\subsection{Quality indicators}
\label{sec:moIndicator}

Let $P \subset V$ be a non-dominated point set of size $n$  found by an EMO algorithm.
It is desirable that $P$ approximates the PF well.
Let $\Omega$ be the set of all non-dominated point sets in the objective space $V$.
A quality indicator $\mathcal{I}: \Omega \rightarrow \mathbb{R}$ evaluates the quality of $P$ in terms of at least one of the convergence, uniformity, and spread \cite{KnowlesC02,ZitzlerTLFF03,LiY19}.
Here, a combination of the uniformity and the spread is called ``diversity" in the EMO community.
Since we focus only on the distribution of non-dominated points throughout this paper, we do not consider the cardinality.
This paper also considers only unary quality indicators.

Representative quality indicators include hypervolume (HV) \cite{ZitzlerT98}, inverted generational distance (IGD) \cite{CoelloS04}, IGD plus (IGD$^+$) \cite{Ishibuchi15}, the additive $\epsilon$-indicator ($\epsilon$) \cite{ZitzlerTLFF03}, R2 \cite{HansenJ98}, new R2 indicator (NR2) \cite{Shang18}, and $s$-energy \cite{HardinS04}.
For their details, see \cite{LiY19,TanabeI20b}.
Briefly speaking, HV calculates the volume of the region dominated by the points in $P$ and bounded by the reference point $r \in V$.
IGD measures the average distance from each reference point $s \in S \subset V$ to its nearest point in $P$, where $S$ is a set of reference points uniformly distributed on the PF.
Since IGD is not Pareto-compliant, IGD can mislead the results \cite{SchutzeELC12,Falcon-CardonaE22}.
IGD$^+$ is a weakly Pareto-compliant version of IGD that uses a modified distance function.
The $\epsilon$ indicator measures the minimum shift such that each point in $P$ weakly dominates at least one reference point in $S$.
R2 calculates the average minimum values of the weighted Tchebycheff function values of $P$ with respect to a weight vector set $W$.
NR2 is an improved version of R2 that more closely approximates the HV value of $P$.
The $s$-energy indicator was originally proposed in the literature of mathematics.
The $s$-energy indicator evaluates the uniformity of the points in $P$.

Let us consider the ranking of all point sets in $V$ by a quality indicator $\mathcal{I}$.
If the ranking by $\mathcal{I}$ is consistent with the Pareto dominance relation, $\mathcal{I}$ is said to be Pareto-compliant \cite{KnowlesTZ06}.
Since HV is one of the most popular Pareto-compliant indicators, HV is a reasonable first choice. 
However, HV does not accurately evaluate the uniformity of a point set in most cases \cite{TanabeI20b,AugerBBZ09,JiangOZF14}.
Other quality indicators are generally used in combination to complement the results of HV.

As reviewed in \cite{TanabeL24}, some preference-based quality indicators (e.g., R-HV and R-IGD~\cite{LiDY18}) have been proposed for benchmarking preference-based EMO algorithms~\cite{BechikhKSG15}.
Preference-based quality indicators take the decision maker's preference information into account in quality assessment.
Thus, they differ from the general quality indicators (e.g., HV and IGD), which do not consider a priori preference information from the decision maker.
Addressing preference-based quality indicators is beyond the scope of this paper.

\subsection{Indicator-based subset selection problem}
%\subsection{ISSP}
\label{sec:issp}

%The ISSP aims to find a subset $S$ of a non-dominated point set $P$ that maximizes a quality indicator $\mathcal{I}$ as follows:
Given a $d$-dimensional objective space $V \subseteq \mathbb{R}^d$, the ISSP is to find a subset $S$ with the maximum quality indicator value $\mathcal{I}(S)$:
\begin{align}
\label{eqn:issp}
    S^* = \argmax_{\substack{S \subset P, |S| = k}} \mathcal{I}(S), 
\end{align}
where $P$ is a set of $n$ non-dominated points, and $k$ is the size of $S$.
Note that $k<n$.
The quality indicator $\mathcal{I}$ to be minimized (e.g., IGD and R2) can be reformulated as $-\mathcal{I}$ without loss of generality.
In equation \eqref{eqn:issp}, the number of all possible subsets is $\binom{n}{k}$.
Technically, a subset $S$ can be represented by an $n$-dimensional $0$-$1$ vector $u=(u_1, \dots, u_n)^{\top}$, where $\sum^{n}_{i=1} u_i = k$.
For $i \in \{1, \dots, n\}$, if $u_i=1$, $S$ includes the $i$-th point $p_i$ in $P$.
For example, when $u=(0, 1, 1, 0, 1)^{\top}$ for $n=5$ and $k=3$, $S=\{p_2, p_3, p_5\}$.

%If $u_i=0$, $S$ includes the $i$-th point $p_i$ in $P$.
% % \cite{BaderZ11,GuerreiroFP16}

The hypervolume subset selection problem (HSSP) \cite{BaderZ11,GuerreiroFP16} is a special case of the ISSP using HV as $\mathcal{I}$.
The HSSP has been well studied in the EMO community.
In addition, the ISSPs using $\epsilon$, R2, and IGD have also been addressed in \cite{BringmannFK14ppsn,ShangIC21}, and \cite{ChenIS22}, respectively.
The ISSP is an NP-hard combinatorial optimization problem as the same as a general subset selection problem \cite{BringmannCE17}.
Only for $d=2$, the optimal subset $S^*$ of the ISSP using HV and $\epsilon$ can be found by dynamic programming \cite{BringmannFK14}.

The definition of the ISSP in equation \eqref{eqn:issp} can be applied to any constrained multi-objective optimization \cite{LiCFY19,LiLLY25} when $P$ consists of feasible non-dominated points.
Therefore, conventional methods for the ISSP can perform subset selection of such $P$ without any change.
The existence of constraints could introduce a discontinuity in the PF, which is addressed in this paper.

%Subset selection of a set containing both feasible and infeasible points has never been studied in the context of the ISSP, and thus it is beyond the scope of this paper.

\IncMargin{0.5em}
\begin{algorithm}[t]
%\scriptsize
%\footnotesize
\small
%\SetAlgoLined
\SetSideCommentRight
%\KwData{this text}
%\KwResult{how to write algorithm with \LaTeX2e }
%$t \leftarrow 1$, initialize the solution of an ISSP $\vector{z} = (z_1, ..., z_{n})^{\top}$\;
Initialize $S \subset P$ randomly\;
\While{There exists a pair of points that improves $\mathcal{I}$}{
  \For{$s \in S$}{
    \For{$p \in P \setminus S$}{
      $S' \leftarrow S \setminus \{s\} \cup \{p\}$\;
      \lIf{$\mathcal{I}(S') > \mathcal{I}(S)$}{
        $S \leftarrow S'$
      }
    }   
  }
  %%   \While{$j < n$ or }{
  %%     }
    %% \If{$HV(\vector{S}\setminus \{s_i\})$}{
    %% $\vector{P} \leftarrow \vector{P} \cup \{\vector{x}\}$ and $\vector{Q} \leftarrow \vector{Q} \setminus \{\vector{x}\}$\;
    %%   }
%  }
%  $t \leftarrow t + 1$\;
}
\caption{LS for the ISSP \cite{BasseurDGL16}}
\label{alg:fils}
\end{algorithm}\DecMargin{0.5em}

\subsection{Local Search for the ISSP}
% \label{sec:issp}

Algorithm \ref{alg:fils} shows first-improvement LS for the ISSP presented in \cite{BasseurDGL16}.
As mentioned in \cite{BasseurDGL16}, a similar LS method was proposed in \cite{BradstreetWB07}.

At the beginning of the search, a subset $S$ is randomly initialized (line 1).
Then, the following steps are repeatedly performed until there does not exist a pair of points that improves $\mathcal{I}$.
For each iteration, a new subset is generated by swapping each point $s$ in $S$ and each unselected point $p$ in $P \setminus S$ (line 5).
Thus, LS in Algorithm \ref{alg:fils} uses the 2-swap neighborhood.
If the new subset $S'$ is better than the current subset $S$ in terms of $\mathcal{I}$, $S$ is replaced with $S'$ (line 6).
Note that the final subset found by running LS is a local optimum for the 2-swap neighborhood.

Although the initial $S$ is randomly generated (line 1), it can be set to a subset found by greedy search as demonstrated in \cite{BasseurDGL16}.
In \cite{NanSIH21}, the initial $S$ is generated such that $k$ points in $S$ are uniformly distributed in the objective space $V$.

\begin{figure}[t]
\captionsetup[subfloat]{farskip=2pt,captionskip=1pt}
\centering
\newcommand{\width}{0.24}
\subfloat[HV]{\includegraphics[width=\width\textwidth]{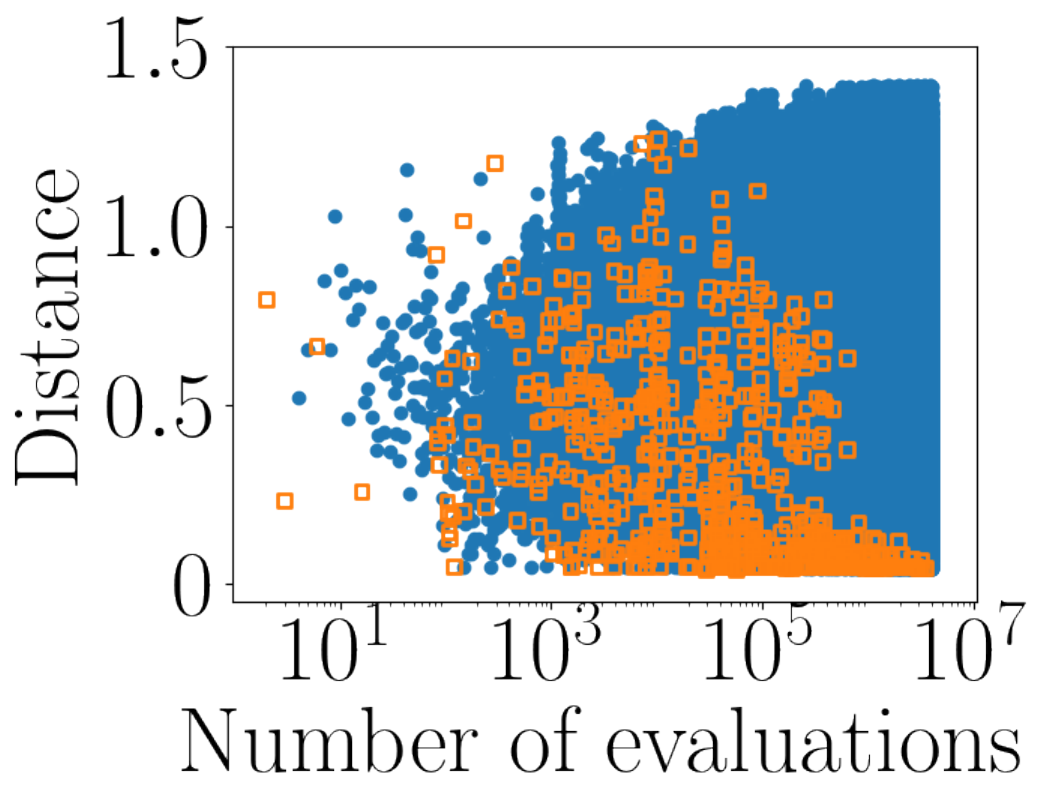}}
\subfloat[IGD]{\includegraphics[width=\width\textwidth]{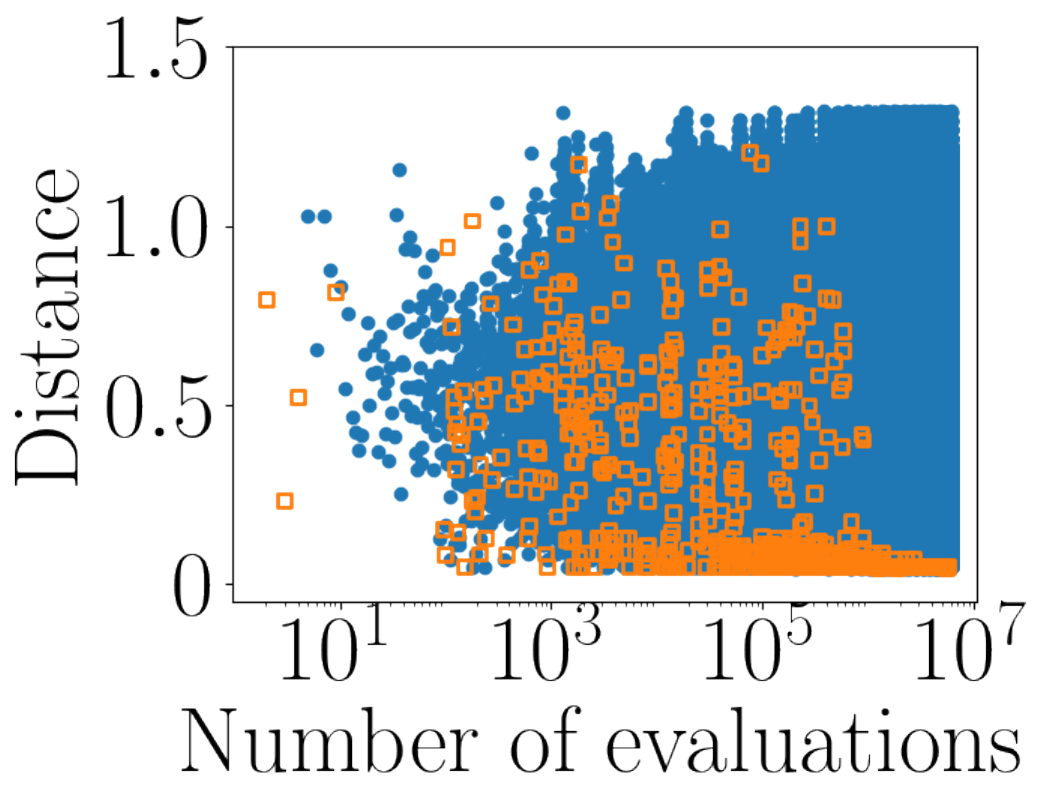}}\\
\subfloat[IGD$^+$]{\includegraphics[width=\width\textwidth]{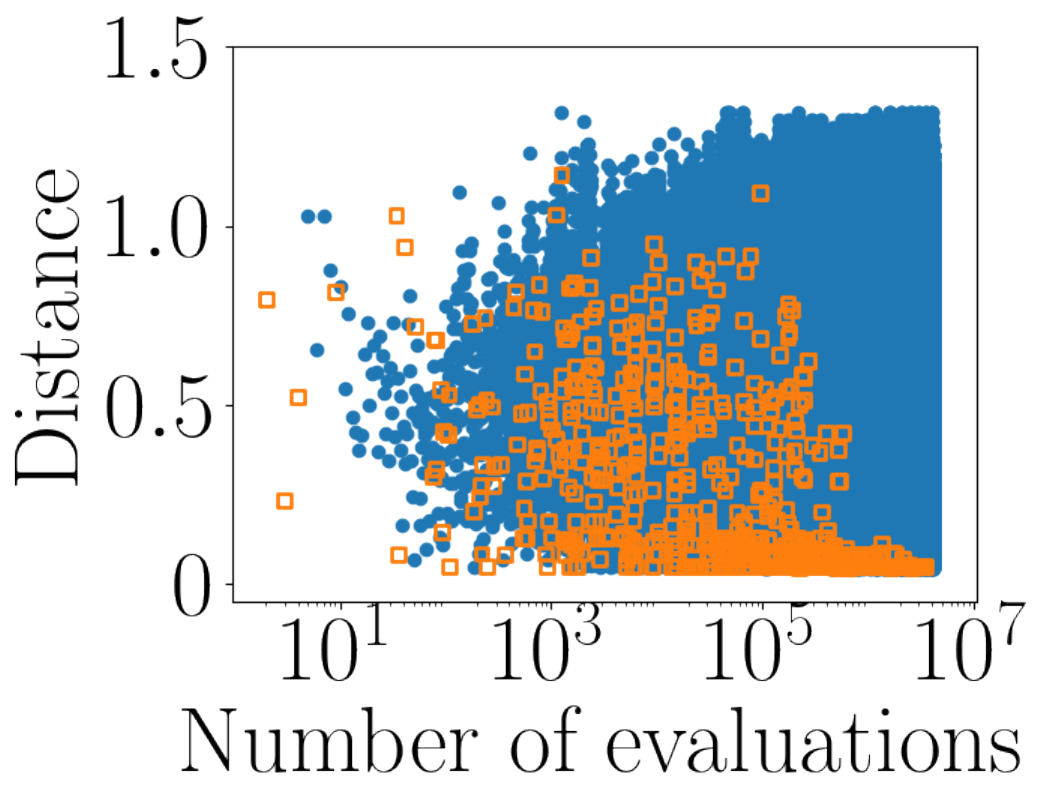}}
\subfloat[$\epsilon$]{\includegraphics[width=\width\textwidth]{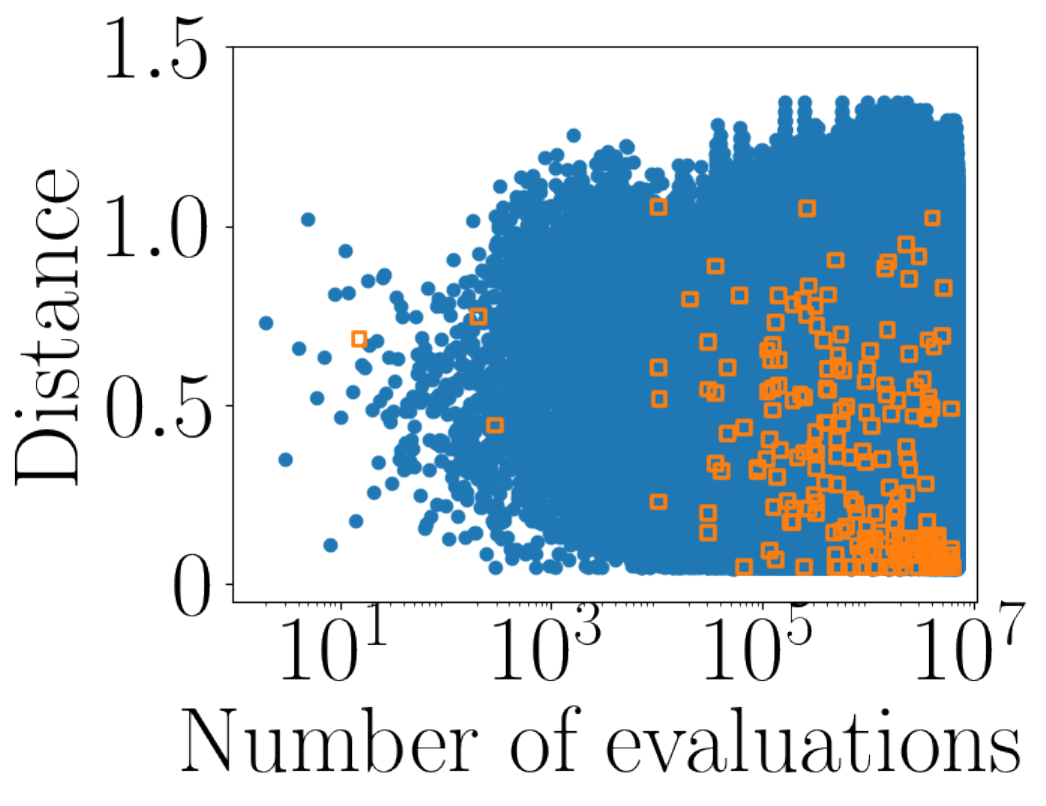}}\\
\subfloat[R2]{\includegraphics[width=\width\textwidth]{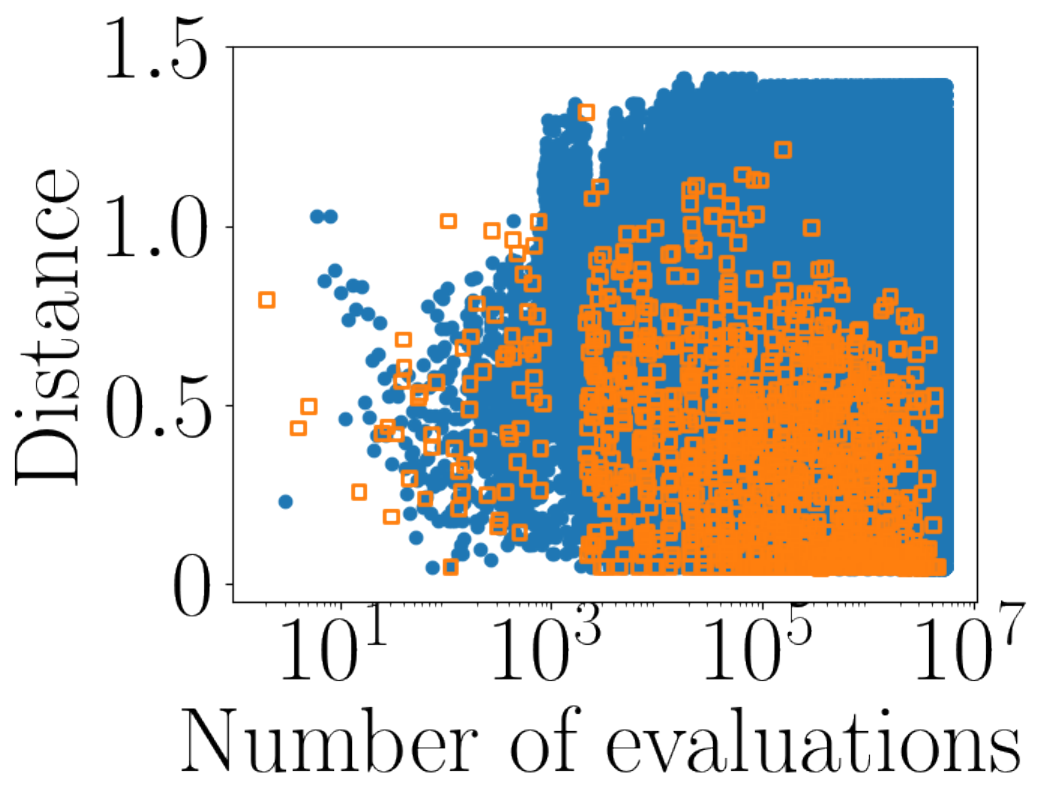}}
\subfloat[NR2]{\includegraphics[width=\width\textwidth]{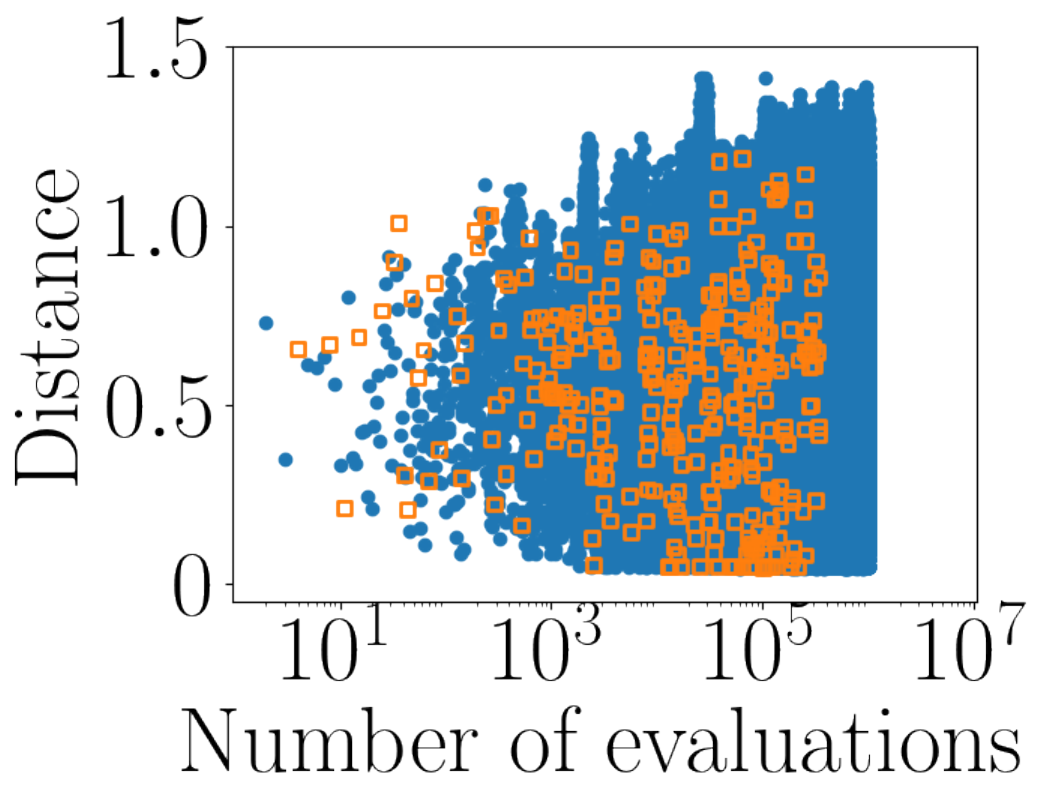}}\\
\subfloat[$s$-energy]{\includegraphics[width=\width\textwidth]{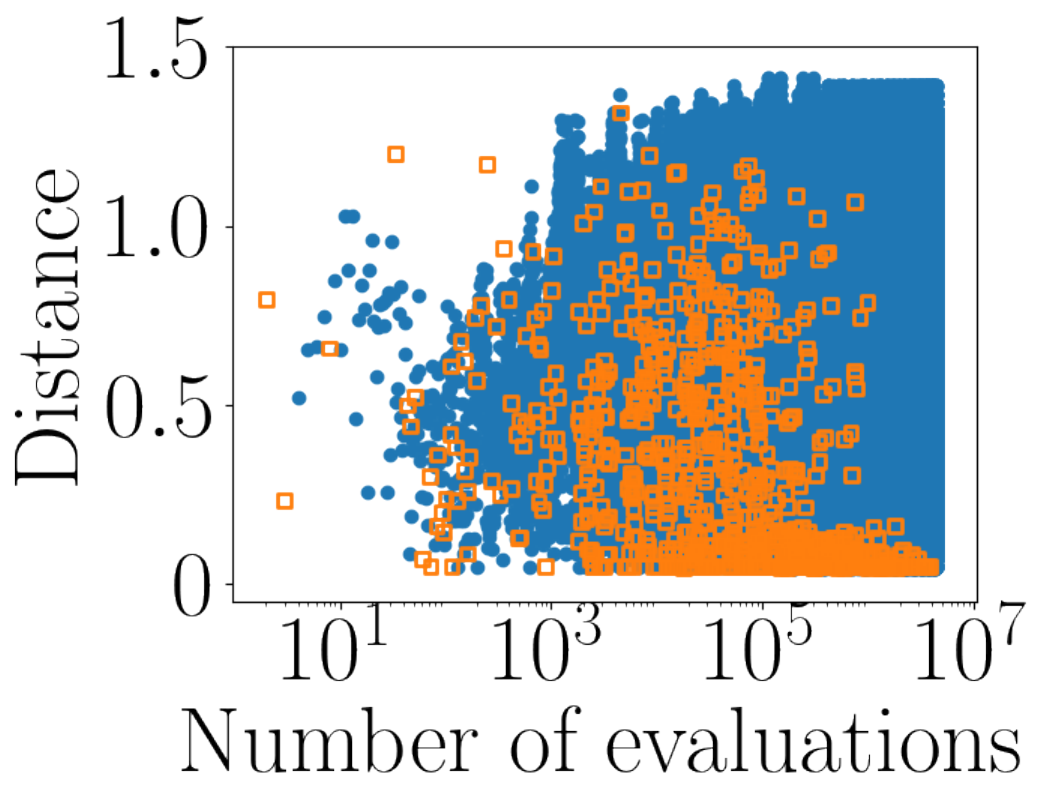}}
\caption{
Distribution of \texttt{dist}$^{\mathrm{s}}$ and \texttt{dist}$^{\mathrm{us}}$ values in a representative run of LS, where \taborange{$\square$} indicates a successful swap, and \tabblue{$\bullet$} indicates an unsuccessful swap.
The results are shown for the ISSPs using the seven quality indicators.
}
   \label{fig:dist_s_and_dist_us}
\end{figure}

\section{Analysis of the distance between two exchanged points}
\label{sec:analysis_cand_list}

%This section investigates the property of the ISSP. % to discuss the effectiveness of the neighborhood restriction in LS.
%The research question is \textit{how far apart two points  in a successful swap  in LS are}.
This section investigates the property of the ISSP to answer the following research question:

%\textbf{$\bullet$ RQ1} \textit{how far apart are two points in a successful swap  in LS?}.

\begin{enumerate}[RQ1:]
\item How far apart are two points in a successful swap in LS?
\end{enumerate}

\noindent Here, in lines 5--6 in Algorithm \ref{alg:fils}, we say that a swap of two points $s$ and $p$ in LS is successful when $\mathcal{I}(S') > \mathcal{I}(S)$.
Recall that $\mathcal{I}$ is to be maximized.
Otherwise, we say that the swap is unsuccessful.
Below, we focus on the distance between two points when their swap is successful or unsuccessful.

Let \texttt{dist}$^{\mathrm{s}}$ be the Euclidean distance between $s$ and $p$ for the successful swap.
In addition, let \texttt{dist}$^{\mathrm{us}}$ be that for the unsuccessful swap.
Note that both \texttt{dist}$^{\mathrm{s}}$ and \texttt{dist}$^{\mathrm{us}}$ are calculated in the objective space, not the solution space. 
Fig. \ref{fig:dist_s_and_dist_us} shows \texttt{dist}$^{\mathrm{s}}$ and \texttt{dist}$^{\mathrm{us}}$ for each evaluation of a subset by $\mathcal{I}$ until the end of the search on the ISSP using the seven quality indicators ( HV, IGD, IGD$^+$, $\epsilon$, R2, NR2, and $s$-energy) described in Section \ref{sec:issp}.
We set $n=5\,000$ and $d=4$.
%We used the linear PF.
We describe the detail of the experimental setup in Section \ref{sec:setting} later.
% Figs. \ref{supfig:dist_s_and_dist_us_nonconvex}--\ref{supfig:dist_s_and_dist_us_convex} in the supplementary file shows the results on other ISSP instances.
% We do not describe Figure XXX due to the paper length limitation, but it is similar to Figure \ref{fig:dist_s_and_dist_us}.
Fig. \ref{fig:dist_s_and_dist_us} shows the results on the linear PF.
% The results on the nonconvex and convex PFs are found in Figs. \ref{supfig:dist_s_and_dist_us_nonconvex} and \ref{supfig:dist_s_and_dist_us_convex}, respectively.
% Since Figs. \ref{supfig:dist_s_and_dist_us_nonconvex} and \ref{supfig:dist_s_and_dist_us_convex} are similar to Fig. \ref{fig:dist_s_and_dist_us}, we do not describe them.
%
In addition, Section \ref{supsec:improvement} investigates the relationship between the improvement of quality indicator values and $\texttt{dist}^{{\mathrm s}}$.
Section \ref{supsec:success_rate} also investigates the relationship between the success rate of swapping two points and $\texttt{dist}^{{\mathrm s}}$.

As shown in the results of HV in Fig. \ref{fig:dist_s_and_dist_us}(a), \texttt{dist}$^{\mathrm{s}}$ values are relatively large within $10^6$ evaluations.
HV generally prefers a diverse distribution of points \cite{TanabeI20b,AugerBBZ09,JiangOZF14}.
When a subset $S$ is randomly initialized, the diversity of $S$ is poor in most cases.
For these reasons, swapping two points far away from each other is successful at an early stage.
The previous study \cite{NanSIH21} proposed a variant of LS for the ISSP that swaps a point in $S$ and the farthest point from $P \setminus S$ for the first 50 iterations.
Although the previous study \cite{NanSIH21} did not show the rationale of this strategy, it is consistent with our observation.

In contrast to the results at an early stage, as seen from Fig. \ref{fig:dist_s_and_dist_us}(a), \texttt{dist}$^{\mathrm{s}}$ values are small for more than $10^6$ evaluations, where \texttt{dist}$^{\mathrm{s}} = 0.129$ even for the maximum case.
This result indicates that the HV value of $S$ generally cannot be improved by swapping two points far away from each other at a later stage of the search.
This observation suggests that the number of evaluations of unpromising new subsets can be reduced by swapping a point in $S$ only with its near points.

Although we discussed only the results for HV shown in Fig. \ref{fig:dist_s_and_dist_us}(a), the results for IGD, IGD$^+$, and $s$-energy are similar to the results for HV.
In contrast, the results for R2, NR2, and $\epsilon$ are different from the results for HV.
As described in Section \ref{sec:moIndicator}, the calculation of R2, NR2, and $\epsilon$ include the ``\texttt{min}" and ``\texttt{max}" operations.
A contribution of each point in $S$ can influence some quality indicators with the ``\texttt{sum}" operation in a cumulative manner.
In contrast, only $d$ extreme points are likely to have an impact on some quality indicators with the ``\texttt{min}" and ``\texttt{max}" operations.
In this case, other points do not contribute to the quality indicator value.
Thus, the values of quality indicators that use the ``\texttt{min}" and ``\texttt{max}" operations may be improved or worsened only when changing points extremely.
This is the reason why swapping two points far away from each other is effective on the ISSP using R2, NR2, and $\epsilon$ even at a later stage of the search, as shown in Figs. \ref{fig:dist_s_and_dist_us}(d) and (e).

%For these reasons, the ISSP using $\epsilon$ and R2 may not share the same property with the ISSP using other quality indicators.

Figs. \ref{supfig:dist_s_and_dist_us_nonconvex} and \ref{supfig:dist_s_and_dist_us_convex} in the supplementary file show the results for the nonconvex and convex PFs, respectively.
Although we do not describe Figs. \ref{supfig:dist_s_and_dist_us_nonconvex} and \ref{supfig:dist_s_and_dist_us_convex} in detail, these results are similar to those in Fig. \ref{fig:dist_s_and_dist_us}, including the unexpected behavior of LS on the ISSP using R2, NR2, and $\epsilon$ (see the discussion above).
Next, we fix the setting of the ISSP as follows: the PF $=$ linear, $d = 4$, $n = 5000$, $k = 100$, and $\mathcal{I} = $ HV. 
Then, we investigate the influence of PF, $d$, $n$, and $k$ by varying each individually.
Fig. \ref{fig:dist_various_settings:pf} shows the results for PF $\in \{$linear, convex, nonconvex, inverted-linear, inverted-convex, inverted-nonconvex$\}$. 
Fig. \ref{fig:dist_various_settings:d} shows the results for $d \in \{2, 3, 4, 5, 6\}$.
Fig. \ref{fig:dist_various_settings:n} also shows the results for $n \in \{3\,000, 4\,000, 5\,000, 6\,000, 7\,000\}$.
In addition, Fig. \ref{fig:dist_various_settings:k} shows the results for $k \in \{50, 75, 100, 125, 150\}$. 
We do not describe these supplemental figures in detail here, but similar conclusions can be drawn from them.
Finally, we performed five independent runs of LS, each with a different initial subset.
The results in Fig. \ref{fig:dist_various_settings:seed} exhibit that all five runs show similar behavior.
In summary, based on these results, 
%we confirmed that the findings in this section are generally observed in various ISSPs.
we confirm that the conclusions in this section are universally applicable.
%Here, note that we do not claim that 

%\begin{tcolorbox}[title=Answers to RQ1, sharpish corners, top=2pt, bottom=2pt, left=4pt, right=4pt, boxrule=0.5pt]
%\textbf{Answers to RQ1:}
\begin{tcolorbox}[sharpish corners, top=2pt, bottom=2pt, left=4pt, right=4pt, boxrule=0.0pt, colback=black!4!white,leftrule=0.75mm,]
\textbf{Answers to RQ1:}
    Our results show that the distance between two points in a successful swap \texttt{dist}$^{\mathrm{s}}$ depends on the stage of the search and the type of quality indicator.
    On the ISSP using the HV, IGD, IGD$^+$, and $s$-energy, while \texttt{dist}$^{\mathrm{s}}$ is large at an early stage, \texttt{dist}$^{\mathrm{s}}$ is small at a later stage.
    Thus, our results suggest that restricting the neighborhood of each point in $S$ can possibly speed up LS on these ISSPs, at least at a later stage.
    %However, our results show that \texttt{dist}$^{\mathrm{s}}$ is large on the ISSP using $\epsilon$ and R2 even at a later stage.
    %Thus, the neighborhood restriction may not be effective for the ISSP using $\epsilon$ and R2.
\end{tcolorbox}

%As shown in Figure \ref{fig:dist_s_and_dist_us}(XXX), \textbf{TODO: Describe the results of $\epsilon$}.

\section{Proposed method}
\label{sec:proposed_method}

This section introduces the proposed candidate list strategy.
First, Section \ref{sec:clist_ls} describes LS with the candidate list strategy.
%Then, we consider two types of candidate lists: the nearest neighbor list and the random neighbor list.
Then, Sections \ref{sec:nbhd_list} and \ref{sec:rand_list} describe the nearest neighbor and random neighbor lists, respectively.
Finally, Section \ref{sec:clist_2face} proposes the two-phase candidate list strategy.

%Below, we discuss the originality of the proposed candidate list strategy.
As noted in Section \ref{sec:introduction}, the candidate list strategy has not been previously studied in the context of the ISSP.
Thus, the proposed candidate list strategy for the ISSP provides a new perspective.
Note that the concept of the candidate list strategy itself is general and not novel.
No work can claim that using the candidate list strategy itself is the original contribution.
Instead, the main point is how to make use of such a general idea on a particular problem domain.
In fact, how to restrict the neighborhood is not obvious on an unseen problem, including the ISSP.
%Restricting the neighborhood in a previously unexplored problem, such as the ISSP, is not a straightforward task.
We designed the proposed candidate list strategy based on the analysis in Section \ref{sec:analysis_cand_list}, where the analysis is another key  contribution of this paper.

%Clearly, it is impossible to straightforwardly re-use the candidate list strategy for the TSP on the ISSP.

% In addition, defining the neighborhoods of local search on an unseen problem domain is not obvious and requires an analysis.
% In fact, we defined the neighborhoods based on the on the analysis in Section III.

\subsection{Local search with the candidate list strategy}
\label{sec:clist_ls}

Algorithm \ref{alg:fils_clist} shows LS with a candidate list.
The candidate list $L_p$ is a set of $l$ possible points that are swapped with $p \in P$.
Here, $l$ is the size of the candidate list.
Thus, there are $n$ candidate lists $L_{p_1}, \dots, L_{p_n}$ for $n$ points $p_1, \dots, p_n$, respectively.
%Note that either the nearest neighbor list or the random neighbor list can be incorporated into Algorithm \ref{alg:fils_clist} as the candidate list.
%There is only one difference between Algorithms \ref{alg:fils} and \ref{alg:fils_clist}.
The only difference between Algorithms \ref{alg:fils} and \ref{alg:fils_clist} lies in line 4.
For each selected point $s$ in the current subset $S$, Algorithm \ref{alg:fils} swaps $s$ and an unselected point $p$ in $P$.
In contrast, Algorithm \ref{alg:fils_clist} swaps $s$ and $p$ only in the corresponding candidate list $L_{s}$.

Recall that $n$ is the size of $P$, and $k$ is the size of $S$.
For each iteration, LS in Algorithm \ref{alg:fils} swaps selected $k$ points in $S$ for unselected $n-k$ points in $P \setminus S$.
Thus, each LS iteration requires $k(n-k)$ evaluations of a quality indicator $\mathcal{I}$ \cite{BasseurDGL16}.
In contrast, for each iteration, LS with the candidate list strategy in Algorithm \ref{alg:fils_clist} swaps $k$ points in $S$ only for $l$ points from the corresponding candidate lists in the maximum case.
This means that LS with the candidate list strategy scans only the neighborhood of size $k l$ even in the worst case.
%In summary, the number of evaluations by $\mathcal{I}$ for each iteration can be significantly reduced from $k(n-k)$ to $kl$.
In summary, the candidate list strategy can significantly reduce the number of evaluations of $\mathcal{I}$ at each iteration from $k(n-k)$ to $kl$.
However, unlike Algorithm \ref{alg:fils}, Algorithm \ref{alg:fils_clist} needs to construct $n$ candidate lists for $n$ points in  $P$ at the beginning of the search.
Thus, Algorithm \ref{alg:fils_clist} requires additional time and space costs.

The remaining question is how to generate the candidate list.
This paper proposes two types of candidate lists: the nearest neighbor and  random neighbor lists.
First, Section \ref{sec:nbhd_list} introduces the nearest neighbor list that includes only the $l$ nearest points in the objective space.
We emphasize that the term ``near" is in the objective space, not in the search space of the ISSP.\footnote{For example, as mentioned in Section \ref{sec:issp}, a solution of the ISSP can be represented by an $n$-dimensional 0-1 vector. However, we are not interested in how near two 0-1 vectors are in terms of the Hamming distance.}
Then, Section \ref{sec:rand_list} introduces the random neighbor list that includes only $l$ randomly selected points at the beginning of the search.
If the aim is only to reduce the computational cost of LS, this can be achieved by using any neighbor list.
However, this paper aims to reduce the computational cost of LS without significantly compromising the quality of subsets on various ISSPs.
For this purpose, the two neighbor lists are necessary.

\def\HiLi{\leavevmode\rlap{\hbox to \hsize{\color{black!9}\leaders\hrule height .8\baselineskip depth .5ex\hfill}}}

\IncMargin{0.5em}
\begin{algorithm}[t]
%\scriptsize
%\footnotesize
\small
%\SetAlgoLined
\SetSideCommentRight
Initialize $S \subset P$ randomly\;
\While{There exists a pair of points that improves $\mathcal{I}$}{
\For{$s \in S$}{
  \HiLi \For{$p \in L_s \setminus S$}{
      $S' \leftarrow S \setminus \{s\} \cup \{p\}$\;
      \lIf{$\mathcal{I}(S') > \mathcal{I}(S)$}{
        $S \leftarrow S'$
      }
    }   
  }
}
\caption{LS with the proposed candidate list strategy}
\label{alg:fils_clist}
%\label{alg:fils_clist_rlist_function}
\end{algorithm}\DecMargin{0.5em}

\IncMargin{0.5em}
\begin{algorithm}[t]
%\scriptsize
%\footnotesize
\small
%\SetAlgoLined
\SetSideCommentRight
\For{$p \in P$}{
  $L^{\mathrm{N}}_p \leftarrow \emptyset$\;
  \While{$|L^{\mathrm{N}}_p| < l$}{
     %\vspace{5pt}\scriptsize{$L^{\mathrm{N}}_p \leftarrow L^{\mathrm{N}}_p \cup \left\{\argmin_{q \in P \setminus \{p\} \setminus L^{\mathrm{N}}_p} \mathrm{dist}(p,q)\right\}$}\;
     $L^{\mathrm{N}}_p \leftarrow L^{\mathrm{N}}_p \cup \left\{\argmin_{q \in P \setminus \{p\} \setminus L^{\mathrm{N}}_p} \mathrm{dist}(p,q)\right\}$\;
  }
}
\caption{A method for generating $L^{\mathrm{N}}$}
\label{alg:make_clist}
\end{algorithm}\DecMargin{0.5em}

\begin{figure}[t]
\centering
\subfloat[Continuous PF.]{  \includegraphics[width=0.23\textwidth]{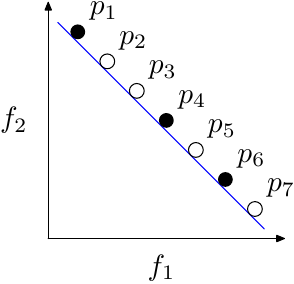}}
\subfloat[Discontinuous PF.]{  \includegraphics[width=0.23\textwidth]{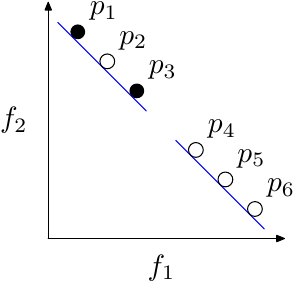}}
\caption{Examples of distributions of points on continuous and discontinuous PFs.}
\label{fig:example_point_exchange}
\end{figure}

\subsection{Nearest neighbor list}
\label{sec:nbhd_list}

For each $p \in P$, the nearest neighbor list $L^{\mathrm{N}}_p$ is a set of  the $l$ nearest points to $p$ in the objective space.
Algorithm \ref{alg:make_clist} shows a method for generating the nearest neighbor list. 
For each $p \in P$, $L^{\mathrm{N}}_p$ is initialized to an empty set (line 2).
Then, the nearest point to $p$ is repeatedly added to $L^{\mathrm{N}}_p$ until $|L^{\mathrm{N}}_p|=l$ (line 4).
Here, $\mathrm{dist}(p, q)$ represents the Euclidean distance between $p$ and $q$ in the objective space.

Fig. \ref{fig:example_point_exchange} shows examples of distributions of points on continuous and discontinuous linear PFs.
In the example of Fig. \ref{fig:example_point_exchange}(a), the current subset $S$ is $\{p_1, p_4, p_6\}$.
%When $l=2$, the nearest neighborhood list $L^{\mathrm{N}}_{p_6}$ of $p_6$ is $\{p_5, p_7\}$.
When $l=2$, the nearest neighbor lists of $p_1$, $p_4$, and $p_6$ are  $L^{\mathrm{N}}_{p_1} = \{p_2, p_3\}$, $L^{\mathrm{N}}_{p_4} = \{p_3, p_5\}$, and $L^{\mathrm{N}}_{p_6} = \{p_5, p_7\}$, respectively.
Let us now consider the swap operation for $p_1$.
In LS in Algorithm \ref{alg:fils}, $p_1$ can be swapped with the four points ($p_2$, $p_3$, $p_5$, and $p_7$).
In contrast, in LS with the nearest neighbor list in Algorithm \ref{alg:fils_clist}, $p_1$ can be swapped only with the two nearest points ($p_2$, $p_3$).

As observed in Section \ref{sec:analysis_cand_list}, swapping two points far away from each other in the objective space is unlikely to improve the quality indicator value of $S$ at a later stage of the search.
In contrast, Algorithm \ref{alg:fils_clist} using the nearest neighbor list $L^{\mathrm{N}}$ swaps only two points close to each other in the objective space. 
Thus, we believe that Algorithm \ref{alg:fils_clist} using $L^{\mathrm{N}}$ can save unpromising swaps.
Note that Algorithm \ref{alg:fils_clist} using $L^{\mathrm{N}}$ can indirectly swap two points far away from each other by repeatedly swapping two points close to each other on a continuous PF.
For example, when $l=2$ in Fig. \ref{fig:example_point_exchange}(a), a swap of $p_4$ and $p_7 \notin L^{\mathrm{N}}_{p_4}$ can be performed by swapping 1) $p_4$ and $p_5 \in L^{\mathrm{N}}_{p_4}$, 2) $p_6$ and $p_7 \in L^{\mathrm{N}}_{p_6}$, and 3) $p_5$ and $p_6 \in L^{\mathrm{N}}_{p_5}$.

% Algorithm \ref{alg:make_clist} needs to construct a candidate list for each $p \in P$ at the beginning of the search.
% The time and space complexities of this constructing procedure are $O((d + \log n) n^2)$ and $O((l + d)n)$, respectively.
%Below, we discuss the complexity of Algorithm \ref{alg:make_clist}.
The time and space complexities of Algorithm \ref{alg:make_clist} are $O((d + \log n) n^2)$ and $O((l + d)n)$, respectively.
First, the distance calculation $\mathrm{dist}(p, q)$ is needed for each $p \in P$ and $q \in P \setminus \{p\}$.
Next, for each $p \in P$, these values are sorted to find $l$ nearest points from $p$.
Note that the time complexity can be reduced to $O(dn^2)$ by simply selecting $l$ smallest values.

In our experiments, constructing the nearest neighbor list requires only a few seconds for $n \leq 10^4$ even in the worst case.
We also observed that the computation time for constructing the nearest neighbor list is much smaller than that for the search by LS.
Thus, the construction of the nearest neighbor list is computationally cheap in practice, except for a special case (e.g., stopping LS immediately after  list construction).

\subsection{Random neighbor list}
\label{sec:rand_list}

For each $p\in P$, the random neighbor list $L^{\mathrm{R}}_p$ consists a set of $l$ randomly selected points from $P \setminus \{p\}$.
The random neighbor list is generated only once at the beginning of the search.
As noted in Section \ref{sec:nbhd_list}, using the nearest neighbor list $L^{\mathrm{N}}$ allows LS to perform an indirect swap of two points far away from each other on a \textit{continuous} PF.
However, this is not always true for a \textit{discontinuous} PF.

The PF in Fig. \ref{fig:example_point_exchange}(b) consists of two subsets. 
When $S=\{p_1, p_3\}$ and $l=2$, $L^{\mathrm{N}}_{p_1}=\{p_2, p_3\}$ and $L^{\mathrm{N}}_{p_3}=\{p_1, p_2\}$.
In this case, LS using the nearest neighbor list cannot swap $p \in \{p_1, p_3\}$ for $q \in \{p_4, p_5, p_6\}$.
In other words, LS using the nearest neighbor list can swap points only on the same subset of the PF.
For this reason, Algorithm \ref{alg:fils_clist} using the nearest neighbor list is likely to perform worse than Algorithm \ref{alg:fils} on ISSP instances with discontinuous PFs in terms of the quality of subsets.
In contrast, Algorithm \ref{alg:fils_clist} using the random neighbor list can swap points on different subsets of the PF.
Therefore, we believe that the random neighbor list is effective on an ISSP instance with a discontinuous PF.

The expected time complexity of randomly selecting $l$ unique points from $P$ is only $O(ln)$.
Thus, the time complexity of constructing the random neighbor list is significantly lower than that of constructing the nearest neighbor list described in Section \ref{sec:nbhd_list}.

\subsection{Sequential use of the two neighbor lists}
\label{sec:clist_2face}

As discussed in Section \ref{sec:rand_list}, LS using the nearest neighbor list is likely to perform poorly on an ISSP instance with a discontinuous PF.
However, we believe that LS using the random neighbor list can achieve only a poor-quality subset.
%This is because this version of LS is unlikely to swap two points close to each other in the objective space.
This is simply because this version of LS can swap only randomly selected points at the beginning of the search.
To address these issues, this section considers the sequential use of the random neighbor and nearest neighbor lists.
It is expected that the drawbacks of the two neighbor lists complement each other by using them sequentially.
Note that the use of multiple neighbor lists has been well studied in the context of variable neighborhood search \cite{MladenovicH97}.
Our contribution here is to propose an efficient way to combine two specific lists for the ISSP.
% combination 

Algorithm \ref{alg:fils_tp_clist} shows LS using the two neighbor lists.
First, the subset $S$ is randomly initialized (line 1).
Then, Algorithm \ref{alg:fils_clist} using the random neighbor list $L^{\mathrm{R}}$ is applied to $S$ (lines 2--3).
Finally, $S$ is further improved by Algorithm \ref{alg:fils_clist} using the nearest neighbor list $L^{\mathrm{N}}$ (lines 4--5).
Here, in Algorithm \ref{alg:fils_tp_clist}, $S$ is not re-initialized at line 1 in Algorithm \ref{alg:fils_clist}.
Since Algorithm \ref{alg:fils_tp_clist} uses the two lists, it requires the total time complexity of constructing them described in Sections \ref{sec:nbhd_list} and \ref{sec:rand_list}, respectively.

Our results in Section \ref{sec:analysis_cand_list} showed that swapping two points far away from each other can improve the quality indicator value of a subset at an early stage in LS (Algorithm \ref{alg:fils}).
Our results also showed that swapping two points close to each other is effective at a later stage.
We believe that Algorithm \ref{alg:fils_tp_clist} can behave similarly to these successful swapping operations in Algorithm \ref{alg:fils}.
The random neighbor list allows LS to swap two points far away from each other.
In contrast, the nearest neighbor list restricts LS to swap only two points close to each other.

We restrict the neighborhood at all stages by using the two neighbor lists.
This is to reduce the number of examined solutions that do not lead to improvements.
It may be possible not to use the nearest neighbor list at ``the early stage" and to use it at ``the later stage".
However, this raises the question of how to define ``the early stage".
For example, it is possible to switch the search strategy after $u$ iterations.
However, the best $u$ value clearly depends on a problem, and finding it on a real-world problem is difficult.
Designing an adaptive switching strategy can be an avenue for future research.

\def\HiLiRAND{\leavevmode\rlap{\hbox to \hsize{\color{magenta!9}\leaders\hrule height .8\baselineskip depth .5ex\hfill}}}
\def\HiLiNBR{\leavevmode\rlap{\hbox to \hsize{\color{cyan!9}\leaders\hrule height .8\baselineskip depth .5ex\hfill}}}

\def\HiLiRAND{\leavevmode\rlap{\hbox to \hsize{\color{magenta!30}\leaders\hrule height .8\baselineskip depth .5ex\hfill}}}
\def\HiLiNBHD{\leavevmode\rlap{\hbox to \hsize{\color{cyan!30}\leaders\hrule height .8\baselineskip depth .5ex\hfill}}}
\IncMargin{0.5em}
\begin{algorithm}[t]
%\scriptsize
%\footnotesize
\small
%\SetAlgoLined
\SetSideCommentRight
Initialize $S$ randomly\;
\HiLiRAND{}$L^{\mathrm{R}}\leftarrow $ Generate the random neighbor list of $P$\; %(Section \ref{sec:rand_list})\;
\HiLiRAND{}$S \leftarrow $ Apply Algorithm \ref{alg:fils_clist} using $L^{\mathrm{R}}$ to $S$\;
\HiLiNBHD{}$L^{\mathrm{N}}\leftarrow $ Generate the nearest neighbor list of $P$\; %(Section \ref{sec:nbhd_list})\;
\HiLiNBHD{}$S \leftarrow $ Apply Algorithm \ref{alg:fils_clist} using $L^{\mathrm{N}}$ to $S$\;
\caption{LS using the two neighbor lists}
\label{alg:fils_tp_clist}
\end{algorithm}\DecMargin{0.5em}

\section{Experimental setup}
\label{sec:setting}

This section describes the experimental setup.
We conducted all experiments on a workstation with an AMD Ryzen Threadripper PRO 3975WX (32-core, 3.5GHz) processor and 512GB of RAM using Ubuntu 22.04.
We implemented all LS methods in \texttt{C++}.
% \footnote{The implementation is available at
% https://github.com/XXXXXXXXX.}
\footnote{The implementation is available at \url{https://github.com/rogi52/issp_ls_clist}}
%\footnote{If this paper is accepted, we will upload the code to GitHub.}
The programs were compiled using GNU \texttt{C++} compiler 11.1.0 with optimization level O2.

This paper considers the ISSPs of the following seven quality indicators: HV, IGD, IGD$^+$, R2, NR2, $\epsilon$, and $s$-energy.
The pagmo \cite{BiscaniI20} implementation of the WFG algorithm \cite{WhileBB12} was used to calculate HV.
For HV and NR2, the reference point $r \in \mathbb{R}^d$ was set to $r=(1.1, ..., 1.1)^{\top}$.
We set the size of the reference point set in IGD and IGD$^+$ to $1\,000$.
We also used a weight vector set of size $1\,000$ for R2 and NR2.

As in \cite{NanSIH23}, $k$ was set to $100$ unless otherwise noted.
% As in \cite{NanSIH23}, we used the following six PFs: a linear PF (DTLZ1), concave PF (DTLZ2), convex PF (convDTLZ2), and their inverted versions.
% However, our preliminary results showed that the results on the six PFs are very similar.
% For this reason, this paper shows the results only on the linear PF.
% In addition, we used a discontinuous PF (DTLZ7).
Section \ref{sec:results} shows the results on the linear continuous PF in DTLZ1 and discontinuous PF in DTLZ7.
In our preliminary experiment, as in \cite{NanSIH23}, we used the following six continuous PFs: a linear PF (DTLZ1), concave PF (DTLZ2), convex PF (convDTLZ2), and their inverted versions.
However, our preliminary results showed that the results for the six PFs are similar.
For this reason, this paper shows the results only for the linear PF.

\definecolor{c1}{RGB}{150,150,150}
\definecolor{c2}{RGB}{220,220,220}

%\begin{landscape}
\begin{table*}[t]
\captionsetup[subfloat]{farskip=2pt,captionskip=1pt}
\setlength{\tabcolsep}{5.0pt} % Default value: 6pt
  \renewcommand{\arraystretch}{0.9} 
\centering
  \caption{\small Comparison of the four LS methods on the ISSP instances with the linear PF and HV, IGD, R2, and $\epsilon$. The average quality indicator values of the subsets found by each LS method are shown. The symbols\PSsymbol and\MSsymbol indicate that the corresponding LS with the candidate list strategy (LS-N, LS-R, or LS-RN) performs significantly better and significantly worse than conventional LS according to the Wilcoxon rank-sum test with $\alpha < 0.05$, respectively. The symbol\EQsymbol indicates no significant difference between the results of the corresponding LS with the candidate list strategy and conventional LS.
  }
  \label{tab:comparison_qi}  
%{\scriptsize
{\footnotesize
%{\small
%%%%%%%%%%%
\subfloat[HV]{
\begin{tabular}{ccccc}
\toprule
$n$ & LS & LS-N & LS-R & LS-RN\\  
\midrule
 $1$K & \cellcolor{c1}\Enote{1.37}{+00} & \cellcolor{c2}\Enote{1.37}{+00}\MSsymbol (\sgnPS\Enote{6.42}{-03}\%) & \Enote{1.37}{+00}\MSsymbol (\sgnPS\Enote{7.56}{-02}\%) &               \Enote{1.37}{+00}\MSsymbol (\sgnPS\Enote{8.36}{-03}\%) \\
 $2$K & \cellcolor{c1}\Enote{1.37}{+00} & \cellcolor{c2}\Enote{1.37}{+00}\MSsymbol (\sgnPS\Enote{3.57}{-03}\%) & \Enote{1.37}{+00}\MSsymbol (\sgnPS\Enote{9.92}{-02}\%) &               \Enote{1.37}{+00}\MSsymbol (\sgnPS\Enote{5.14}{-03}\%) \\
 $3$K & \cellcolor{c1}\Enote{1.37}{+00} & \cellcolor{c2}\Enote{1.37}{+00}\MSsymbol (\sgnPS\Enote{6.44}{-03}\%) & \Enote{1.37}{+00}\MSsymbol (\sgnPS\Enote{1.08}{-01}\%) &               \Enote{1.37}{+00}\MSsymbol (\sgnPS\Enote{6.96}{-03}\%) \\
 $4$K & \cellcolor{c1}\Enote{1.37}{+00} & \cellcolor{c2}\Enote{1.37}{+00}\MSsymbol (\sgnPS\Enote{6.10}{-03}\%) & \Enote{1.37}{+00}\MSsymbol (\sgnPS\Enote{1.20}{-01}\%) &               \Enote{1.37}{+00}\MSsymbol (\sgnPS\Enote{6.38}{-03}\%) \\
 $5$K & \cellcolor{c1}\Enote{1.37}{+00} &               \Enote{1.37}{+00}\MSsymbol (\sgnPS\Enote{7.72}{-03}\%) & \Enote{1.37}{+00}\MSsymbol (\sgnPS\Enote{1.25}{-01}\%) & \cellcolor{c2}\Enote{1.37}{+00}\MSsymbol (\sgnPS\Enote{6.19}{-03}\%) \\
 $6$K & \cellcolor{c1}\Enote{1.37}{+00} & \cellcolor{c2}\Enote{1.37}{+00}\MSsymbol (\sgnPS\Enote{7.19}{-03}\%) & \Enote{1.37}{+00}\MSsymbol (\sgnPS\Enote{1.36}{-01}\%) &               \Enote{1.37}{+00}\MSsymbol (\sgnPS\Enote{7.54}{-03}\%) \\
 $7$K & \cellcolor{c1}\Enote{1.37}{+00} & \cellcolor{c2}\Enote{1.37}{+00}\MSsymbol (\sgnPS\Enote{5.78}{-03}\%) & \Enote{1.37}{+00}\MSsymbol (\sgnPS\Enote{1.38}{-01}\%) &               \Enote{1.37}{+00}\MSsymbol (\sgnPS\Enote{7.17}{-03}\%) \\
 $8$K & \cellcolor{c1}\Enote{1.37}{+00} & \cellcolor{c2}\Enote{1.37}{+00}\MSsymbol (\sgnPS\Enote{6.00}{-03}\%) & \Enote{1.37}{+00}\MSsymbol (\sgnPS\Enote{1.41}{-01}\%) &               \Enote{1.37}{+00}\MSsymbol (\sgnPS\Enote{8.97}{-03}\%) \\
 $9$K & \cellcolor{c1}\Enote{1.37}{+00} & \cellcolor{c2}\Enote{1.37}{+00}\MSsymbol (\sgnPS\Enote{6.95}{-03}\%) & \Enote{1.37}{+00}\MSsymbol (\sgnPS\Enote{1.40}{-01}\%) &               \Enote{1.37}{+00}\MSsymbol (\sgnPS\Enote{9.03}{-03}\%) \\
$10$K & \cellcolor{c1}\Enote{1.37}{+00} &               \Enote{1.37}{+00}\MSsymbol (\sgnPS\Enote{8.56}{-03}\%) & \Enote{1.37}{+00}\MSsymbol (\sgnPS\Enote{1.40}{-01}\%) & \cellcolor{c2}\Enote{1.37}{+00}\MSsymbol (\sgnPS\Enote{8.53}{-03}\%) \\
\toprule
\end{tabular}
}
\\
%%%%%%%%%%%
\subfloat[IGD]{
\begin{tabular}{ccccc}
\toprule
$n$ & LS & LS-N & LS-R & LS-RN\\  
\midrule
 $1$K & \cellcolor{c1}\Enote{8.07}{-02} &               \Enote{8.10}{-02}\MSsymbol (\sgnPS\Enote{4.25}{-01}\%) & \Enote{8.24}{-02}\MSsymbol (\sgnPS\Enote{2.14}{+00}\%) & \cellcolor{c2}\Enote{8.08}{-02}\MSsymbol (\sgnPS\Enote{1.52}{-01}\%) \\
 $2$K & \cellcolor{c1}\Enote{8.06}{-02} &               \Enote{8.09}{-02}\MSsymbol (\sgnPS\Enote{3.24}{-01}\%) & \Enote{8.31}{-02}\MSsymbol (\sgnPS\Enote{3.06}{+00}\%) & \cellcolor{c2}\Enote{8.08}{-02}\MSsymbol (\sgnPS\Enote{2.02}{-01}\%) \\
 $3$K & \cellcolor{c1}\Enote{7.98}{-02} & \cellcolor{c2}\Enote{8.00}{-02}\MSsymbol (\sgnPS\Enote{2.33}{-01}\%) & \Enote{8.31}{-02}\MSsymbol (\sgnPS\Enote{4.09}{+00}\%) &               \Enote{8.00}{-02}\MSsymbol (\sgnPS\Enote{2.91}{-01}\%) \\
 $4$K & \cellcolor{c1}\Enote{7.96}{-02} &               \Enote{7.98}{-02}\MSsymbol (\sgnPS\Enote{2.99}{-01}\%) & \Enote{8.30}{-02}\MSsymbol (\sgnPS\Enote{4.31}{+00}\%) & \cellcolor{c2}\Enote{7.97}{-02}\MSsymbol (\sgnPS\Enote{1.99}{-01}\%) \\
 $5$K & \cellcolor{c1}\Enote{7.94}{-02} &               \Enote{7.96}{-02}\MSsymbol (\sgnPS\Enote{3.66}{-01}\%) & \Enote{8.28}{-02}\MSsymbol (\sgnPS\Enote{4.34}{+00}\%) & \cellcolor{c2}\Enote{7.96}{-02}\MSsymbol (\sgnPS\Enote{2.53}{-01}\%) \\
 $6$K & \cellcolor{c1}\Enote{7.92}{-02} &               \Enote{7.95}{-02}\MSsymbol (\sgnPS\Enote{3.74}{-01}\%) & \Enote{8.28}{-02}\MSsymbol (\sgnPS\Enote{4.51}{+00}\%) & \cellcolor{c2}\Enote{7.94}{-02}\MSsymbol (\sgnPS\Enote{2.04}{-01}\%) \\
 $7$K & \cellcolor{c1}\Enote{7.92}{-02} &               \Enote{7.94}{-02}\MSsymbol (\sgnPS\Enote{2.84}{-01}\%) & \Enote{8.27}{-02}\MSsymbol (\sgnPS\Enote{4.47}{+00}\%) & \cellcolor{c2}\Enote{7.93}{-02}\MSsymbol (\sgnPS\Enote{1.83}{-01}\%) \\
 $8$K & \cellcolor{c1}\Enote{7.91}{-02} &               \Enote{7.94}{-02}\MSsymbol (\sgnPS\Enote{3.30}{-01}\%) & \Enote{8.28}{-02}\MSsymbol (\sgnPS\Enote{4.59}{+00}\%) & \cellcolor{c2}\Enote{7.93}{-02}\MSsymbol (\sgnPS\Enote{1.91}{-01}\%) \\
 $9$K & \cellcolor{c1}\Enote{7.90}{-02} &               \Enote{7.93}{-02}\MSsymbol (\sgnPS\Enote{3.67}{-01}\%) & \Enote{8.27}{-02}\MSsymbol (\sgnPS\Enote{4.72}{+00}\%) & \cellcolor{c2}\Enote{7.92}{-02}\MSsymbol (\sgnPS\Enote{2.89}{-01}\%) \\
$10$K & \cellcolor{c1}\Enote{7.90}{-02} &               \Enote{7.93}{-02}\MSsymbol (\sgnPS\Enote{3.70}{-01}\%) & \Enote{8.28}{-02}\MSsymbol (\sgnPS\Enote{4.87}{+00}\%) & \cellcolor{c2}\Enote{7.92}{-02}\MSsymbol (\sgnPS\Enote{2.52}{-01}\%) \\
\toprule
\end{tabular}
}
\\
\subfloat[R2]{
\begin{tabular}{ccccc}
\toprule
$n$ & LS & LS-N & LS-R & LS-RN\\  
\midrule
 $1$K & \cellcolor{c1}\Enote{2.17}{-02} &               \Enote{2.17}{-02}\MSsymbol (\sgnPS\Enote{1.19}{-01}\%) & \Enote{2.18}{-02}\MSsymbol (\sgnPS\Enote{4.96}{-01}\%) & \cellcolor{c2}\Enote{2.17}{-02}\MSsymbol (\sgnPS\Enote{1.08}{-01}\%) \\
 $2$K & \cellcolor{c1}\Enote{2.15}{-02} &               \Enote{2.16}{-02}\MSsymbol (\sgnPS\Enote{3.12}{-01}\%) & \Enote{2.17}{-02}\MSsymbol (\sgnPS\Enote{1.00}{+00}\%) & \cellcolor{c2}\Enote{2.16}{-02}\MSsymbol (\sgnPS\Enote{2.95}{-01}\%) \\
 $3$K & \cellcolor{c1}\Enote{2.13}{-02} & \cellcolor{c2}\Enote{2.14}{-02}\MSsymbol (\sgnPS\Enote{3.52}{-01}\%) & \Enote{2.18}{-02}\MSsymbol (\sgnPS\Enote{1.98}{+00}\%) &               \Enote{2.14}{-02}\MSsymbol (\sgnPS\Enote{3.55}{-01}\%) \\
 $4$K & \cellcolor{c1}\Enote{2.13}{-02} &               \Enote{2.14}{-02}\MSsymbol (\sgnPS\Enote{4.63}{-01}\%) & \Enote{2.19}{-02}\MSsymbol (\sgnPS\Enote{2.78}{+00}\%) & \cellcolor{c2}\Enote{2.14}{-02}\MSsymbol (\sgnPS\Enote{4.31}{-01}\%) \\
 $5$K & \cellcolor{c1}\Enote{2.13}{-02} &               \Enote{2.14}{-02}\MSsymbol (\sgnPS\Enote{5.38}{-01}\%) & \Enote{2.19}{-02}\MSsymbol (\sgnPS\Enote{2.76}{+00}\%) & \cellcolor{c2}\Enote{2.14}{-02}\MSsymbol (\sgnPS\Enote{4.79}{-01}\%) \\
 $6$K & \cellcolor{c1}\Enote{2.12}{-02} &               \Enote{2.14}{-02}\MSsymbol (\sgnPS\Enote{5.99}{-01}\%) & \Enote{2.19}{-02}\MSsymbol (\sgnPS\Enote{3.32}{+00}\%) & \cellcolor{c2}\Enote{2.13}{-02}\MSsymbol (\sgnPS\Enote{5.62}{-01}\%) \\
 $7$K & \cellcolor{c1}\Enote{2.12}{-02} &               \Enote{2.13}{-02}\MSsymbol (\sgnPS\Enote{6.77}{-01}\%) & \Enote{2.22}{-02}\MSsymbol (\sgnPS\Enote{4.47}{+00}\%) & \cellcolor{c2}\Enote{2.13}{-02}\MSsymbol (\sgnPS\Enote{6.22}{-01}\%) \\
 $8$K & \cellcolor{c1}\Enote{2.12}{-02} & \cellcolor{c2}\Enote{2.13}{-02}\MSsymbol (\sgnPS\Enote{6.88}{-01}\%) & \Enote{2.20}{-02}\MSsymbol (\sgnPS\Enote{3.85}{+00}\%) &               \Enote{2.14}{-02}\MSsymbol (\sgnPS\Enote{7.65}{-01}\%) \\
 $9$K & \cellcolor{c1}\Enote{2.12}{-02} &               \Enote{2.13}{-02}\MSsymbol (\sgnPS\Enote{7.43}{-01}\%) & \Enote{2.21}{-02}\MSsymbol (\sgnPS\Enote{4.46}{+00}\%) & \cellcolor{c2}\Enote{2.13}{-02}\MSsymbol (\sgnPS\Enote{6.23}{-01}\%) \\
$10$K & \cellcolor{c1}\Enote{2.12}{-02} &               \Enote{2.13}{-02}\MSsymbol (\sgnPS\Enote{8.97}{-01}\%) & \Enote{2.21}{-02}\MSsymbol (\sgnPS\Enote{4.57}{+00}\%) & \cellcolor{c2}\Enote{2.13}{-02}\MSsymbol (\sgnPS\Enote{8.18}{-01}\%) \\
\toprule
\end{tabular}
}
\\
\subfloat[$\epsilon$]{
\begin{tabular}{ccccc}
\toprule
$n$ & LS & LS-N & LS-R & LS-RN\\  
\midrule
 $1$K &               \Enote{8.22}{-02} & \Enote{1.04}{-01}\MSsymbol (\sgnPS\Enote{2.71}{+01}\%) & \cellcolor{c2}\Enote{8.11}{-02}\PSsymbol (\sgnMS\Enote{1.31}{+00}\%) & \cellcolor{c1}\Enote{8.07}{-02}\PSsymbol (\sgnMS\Enote{1.90}{+00}\%) \\
 $2$K & \cellcolor{c1}\Enote{9.30}{-02} & \Enote{9.78}{-02}\MSsymbol (\sgnPS\Enote{5.24}{+00}\%) & \cellcolor{c2}\Enote{9.48}{-02}\EQsymbol (\sgnPS\Enote{1.94}{+00}\%) &               \Enote{9.58}{-02}\EQsymbol (\sgnPS\Enote{3.05}{+00}\%) \\
 $3$K &               \Enote{8.67}{-02} & \Enote{9.30}{-02}\MSsymbol (\sgnPS\Enote{7.34}{+00}\%) & \cellcolor{c2}\Enote{8.57}{-02}\EQsymbol (\sgnMS\Enote{1.09}{+00}\%) & \cellcolor{c1}\Enote{8.43}{-02}\EQsymbol (\sgnMS\Enote{2.71}{+00}\%) \\
 $4$K & \cellcolor{c1}\Enote{7.87}{-02} & \Enote{8.74}{-02}\MSsymbol (\sgnPS\Enote{1.11}{+01}\%) &               \Enote{8.50}{-02}\MSsymbol (\sgnPS\Enote{7.97}{+00}\%) & \cellcolor{c2}\Enote{7.88}{-02}\EQsymbol (\sgnPS\Enote{1.94}{-01}\%) \\
 $5$K & \cellcolor{c1}\Enote{7.28}{-02} & \Enote{8.50}{-02}\MSsymbol (\sgnPS\Enote{1.68}{+01}\%) &               \Enote{7.97}{-02}\MSsymbol (\sgnPS\Enote{9.57}{+00}\%) & \cellcolor{c2}\Enote{7.72}{-02}\MSsymbol (\sgnPS\Enote{6.14}{+00}\%) \\
 $6$K & \cellcolor{c1}\Enote{7.36}{-02} & \Enote{9.81}{-02}\MSsymbol (\sgnPS\Enote{3.33}{+01}\%) &               \Enote{8.55}{-02}\MSsymbol (\sgnPS\Enote{1.61}{+01}\%) & \cellcolor{c2}\Enote{8.48}{-02}\MSsymbol (\sgnPS\Enote{1.52}{+01}\%) \\
 $7$K & \cellcolor{c1}\Enote{7.74}{-02} & \Enote{9.29}{-02}\MSsymbol (\sgnPS\Enote{1.99}{+01}\%) &               \Enote{9.01}{-02}\MSsymbol (\sgnPS\Enote{1.63}{+01}\%) & \cellcolor{c2}\Enote{8.39}{-02}\EQsymbol (\sgnPS\Enote{8.36}{+00}\%) \\
 $8$K & \cellcolor{c1}\Enote{7.75}{-02} & \Enote{8.84}{-02}\MSsymbol (\sgnPS\Enote{1.41}{+01}\%) &               \Enote{8.77}{-02}\MSsymbol (\sgnPS\Enote{1.31}{+01}\%) & \cellcolor{c2}\Enote{8.46}{-02}\EQsymbol (\sgnPS\Enote{9.14}{+00}\%) \\
 $9$K & \cellcolor{c1}\Enote{7.25}{-02} & \Enote{8.78}{-02}\MSsymbol (\sgnPS\Enote{2.12}{+01}\%) &               \Enote{8.79}{-02}\MSsymbol (\sgnPS\Enote{2.13}{+01}\%) & \cellcolor{c2}\Enote{8.23}{-02}\MSsymbol (\sgnPS\Enote{1.36}{+01}\%) \\
$10$K & \cellcolor{c1}\Enote{7.44}{-02} & \Enote{9.13}{-02}\MSsymbol (\sgnPS\Enote{2.27}{+01}\%) &               \Enote{8.60}{-02}\MSsymbol (\sgnPS\Enote{1.56}{+01}\%) & \cellcolor{c2}\Enote{8.27}{-02}\MSsymbol (\sgnPS\Enote{1.11}{+01}\%) \\
\toprule
\end{tabular}
}
\\
%\vspace{-20pt}
}
\end{table*}
%\end{landscape}

\begin{figure*}[t]
\captionsetup[subfloat]{farskip=2pt,captionskip=1pt}
\centering
\newcommand{\widthvar}{0.235}
% \subfloat[HV ($\mathcal{I}(S^{\mathrm{bsf}})$)]{  
% \includegraphics[width=0.24\textwidth]{graph/con_scale_n/scale_n_pf-linear_d4_v-qi_I-hv.pdf}
% }
% \subfloat[IGD ($\mathcal{I}(S^{\mathrm{bsf}})$)]{  
% \includegraphics[width=\widthvar\textwidth]{graph/con_scale_n/scale_n_pf-linear_d4_v-qi_I-igd.pdf}
% }
% \subfloat[R2 ($\mathcal{I}(S^{\mathrm{bsf}})$)]{  
% \includegraphics[width=\widthvar\textwidth]{graph/con_scale_n/scale_n_pf-linear_d4_v-qi_I-r2.pdf}
% }
%  \subfloat[$\epsilon$ ($\mathcal{I}(S^{\mathrm{bsf}})$)]{  
%  \includegraphics[width=\widthvar\textwidth]{graph/con_scale_n/scale_n_pf-linear_d4_v-qi_I-epsilon.pdf}
%  }
%  \\
\includegraphics[width=0.4\textwidth]{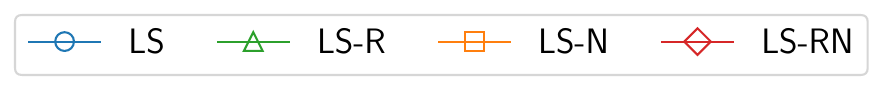}\\%[-1em]
\subfloat[HV (Num. evals.)]{  
\includegraphics[width=\widthvar\textwidth]{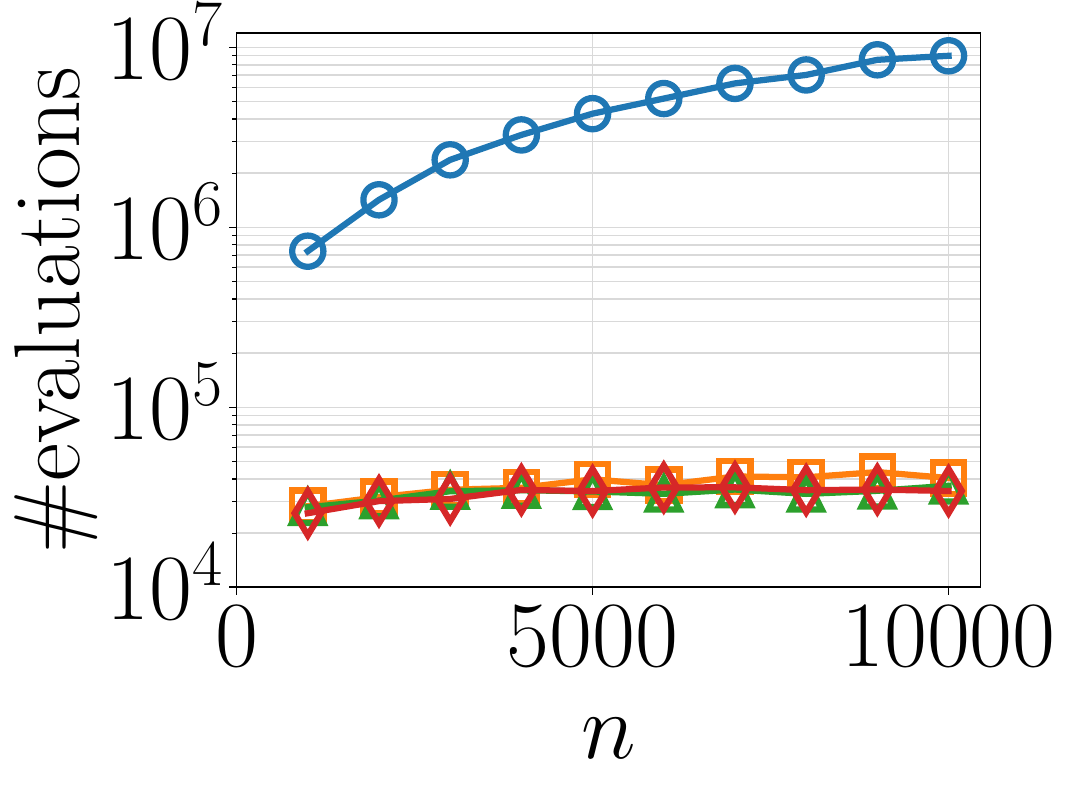}
}
\subfloat[IGD (Num. evals.)]{  
\includegraphics[width=\widthvar\textwidth]{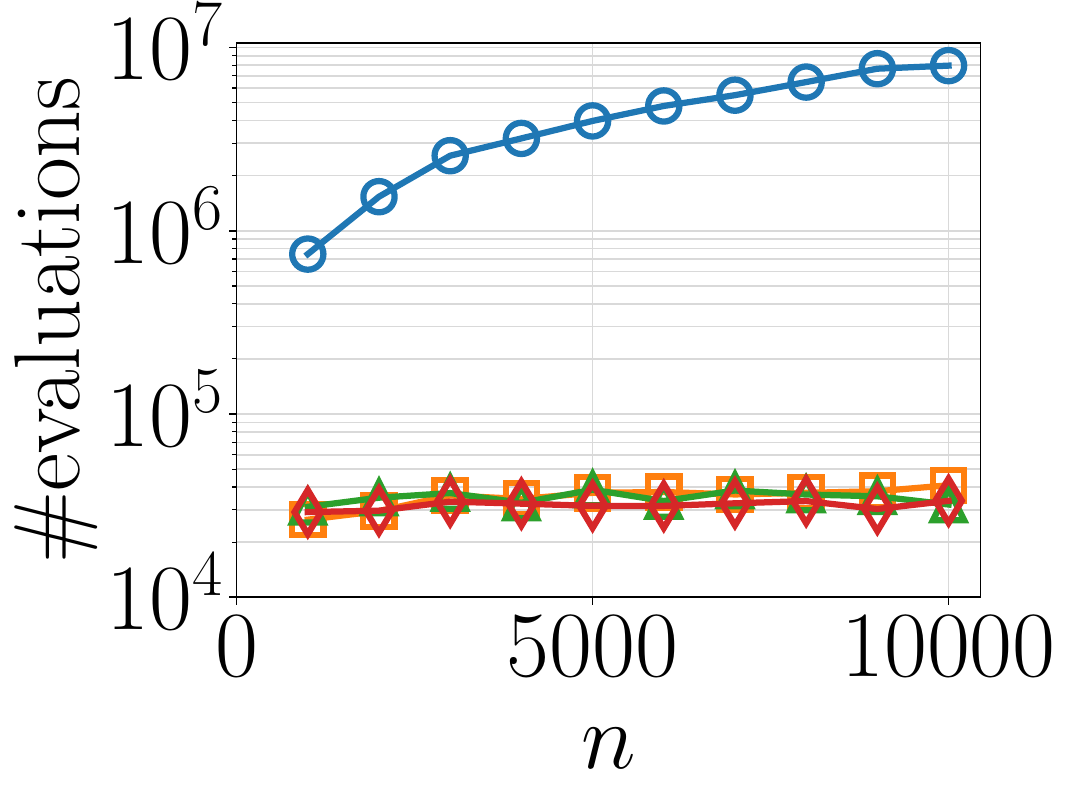}
}
\subfloat[R2 (Num. evals.)]{  
\includegraphics[width=\widthvar\textwidth]{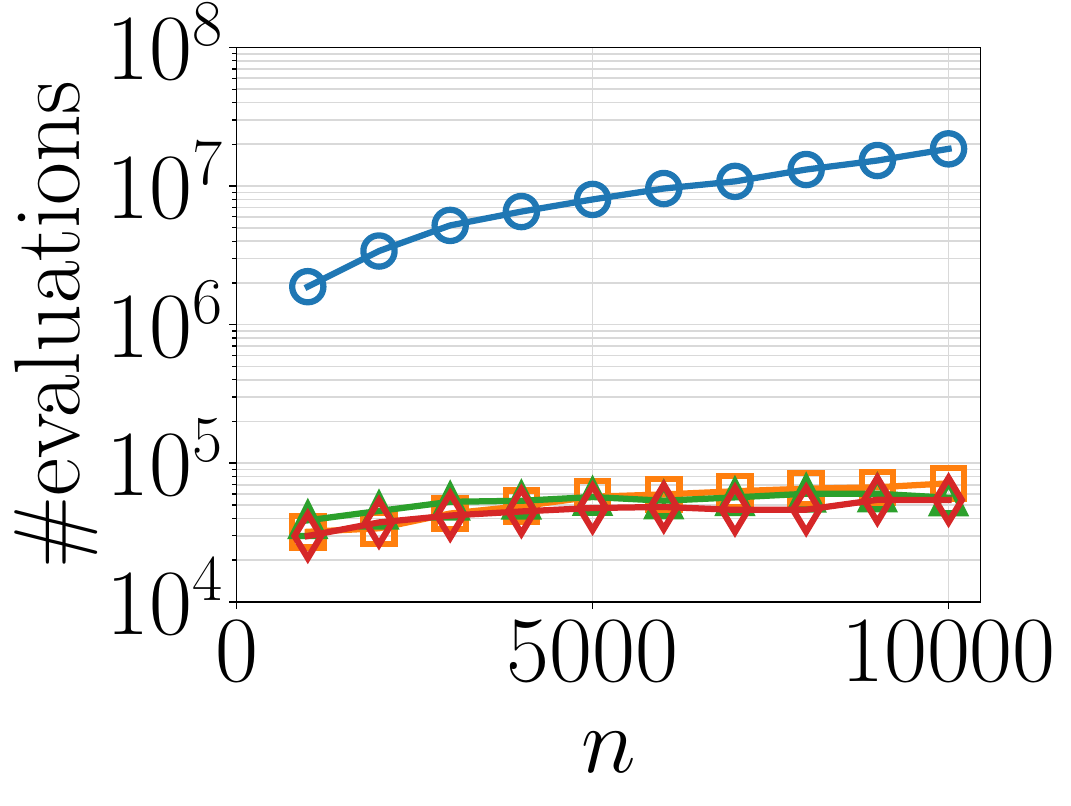}
}
 \subfloat[$\epsilon$ (Num. evals.)]{  
 \includegraphics[width=\widthvar\textwidth]{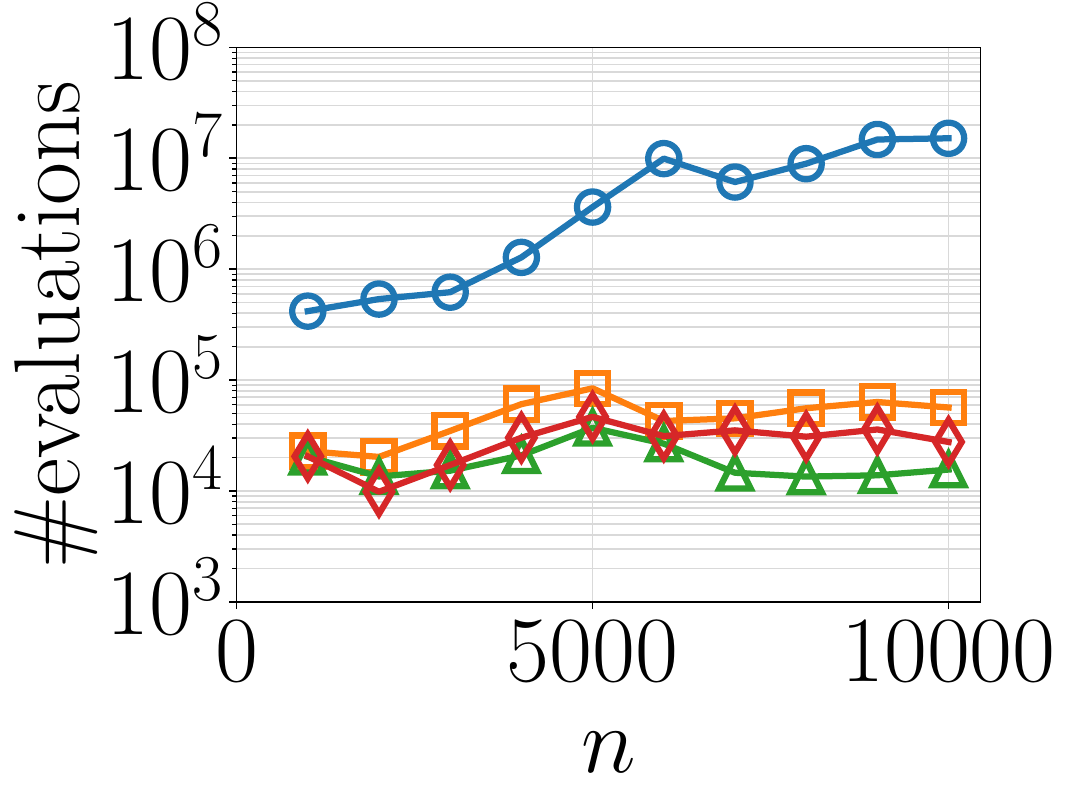}
 }
 \\
\subfloat[HV (Time)]{  
\includegraphics[width=\widthvar\textwidth]{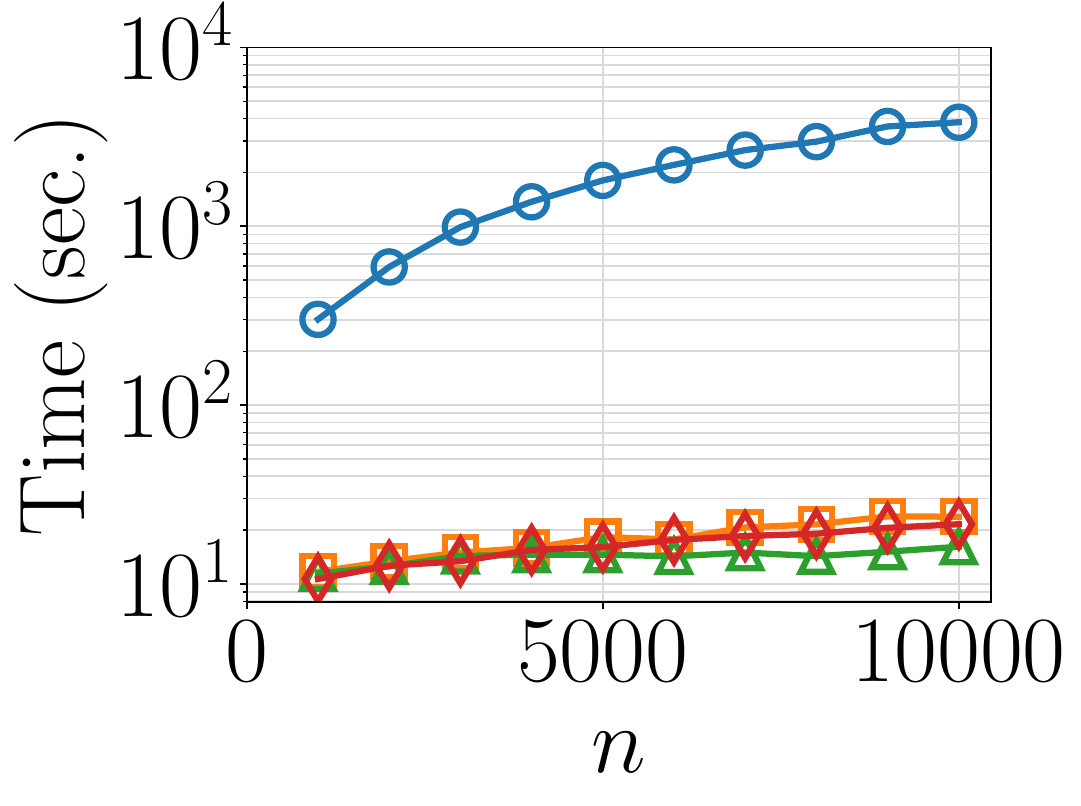}
}
\subfloat[IGD (Time)]{  
\includegraphics[width=\widthvar\textwidth]{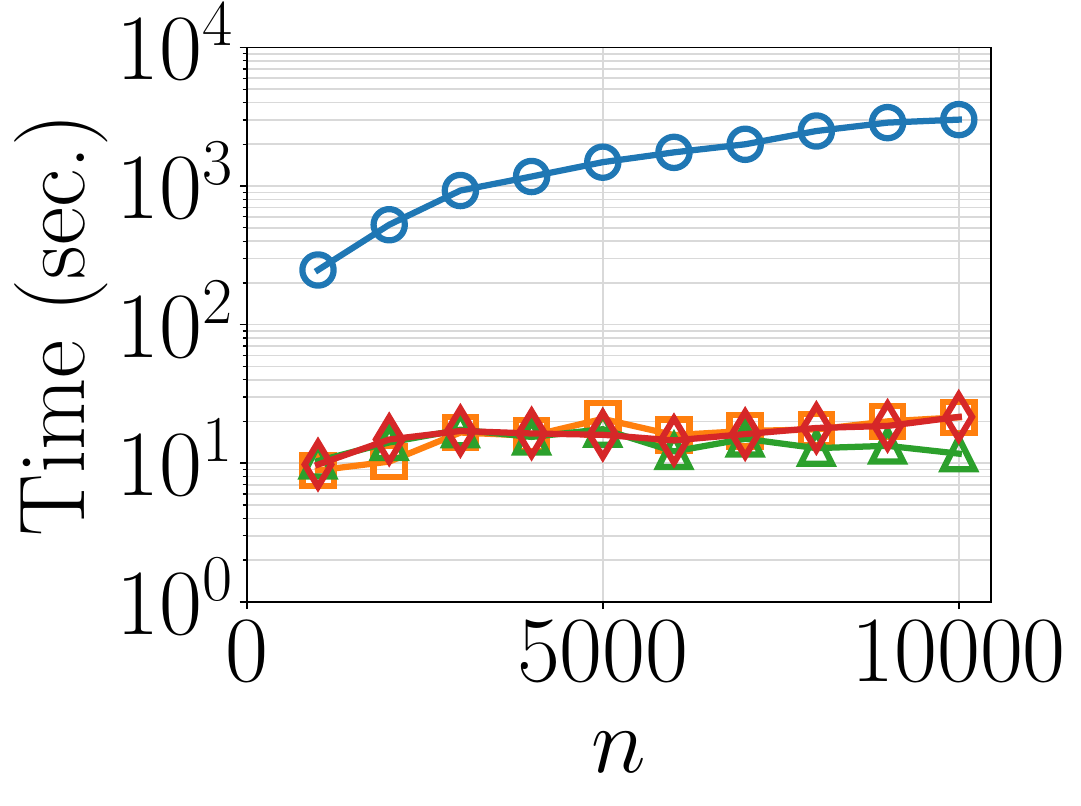}
}
\subfloat[R2 (Time)]{  
\includegraphics[width=\widthvar\textwidth]{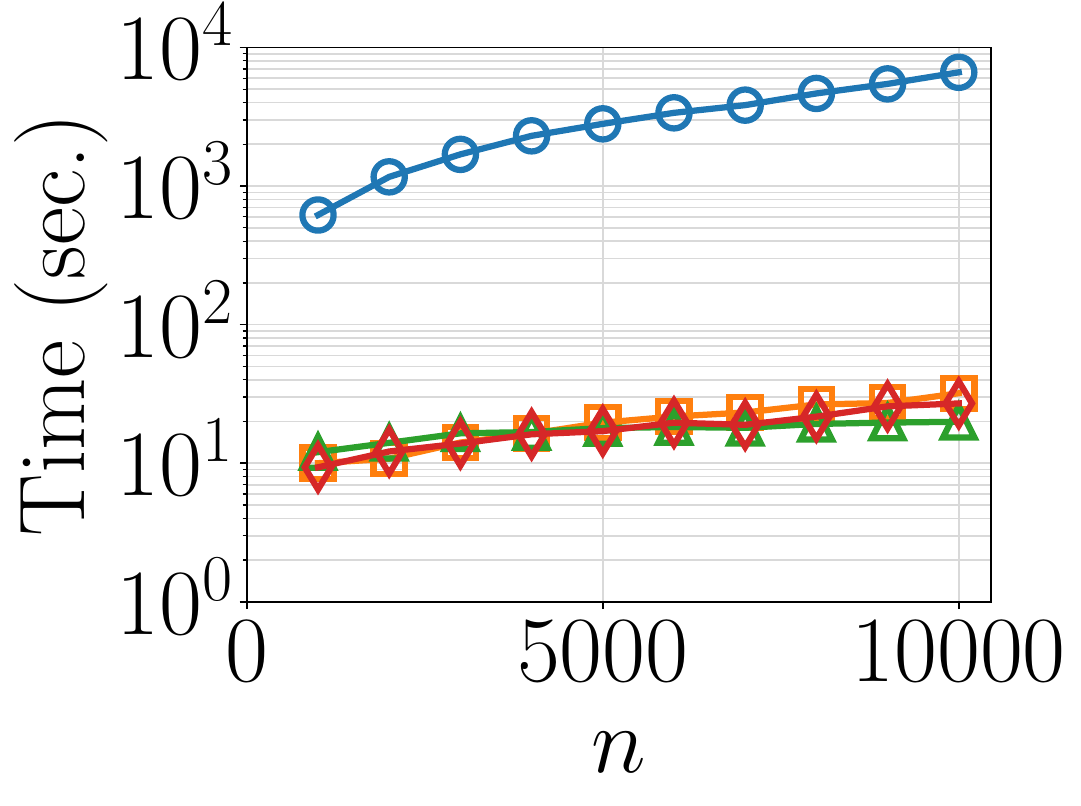}
}
 \subfloat[$\epsilon$ (Time)]{  
 \includegraphics[width=\widthvar\textwidth]{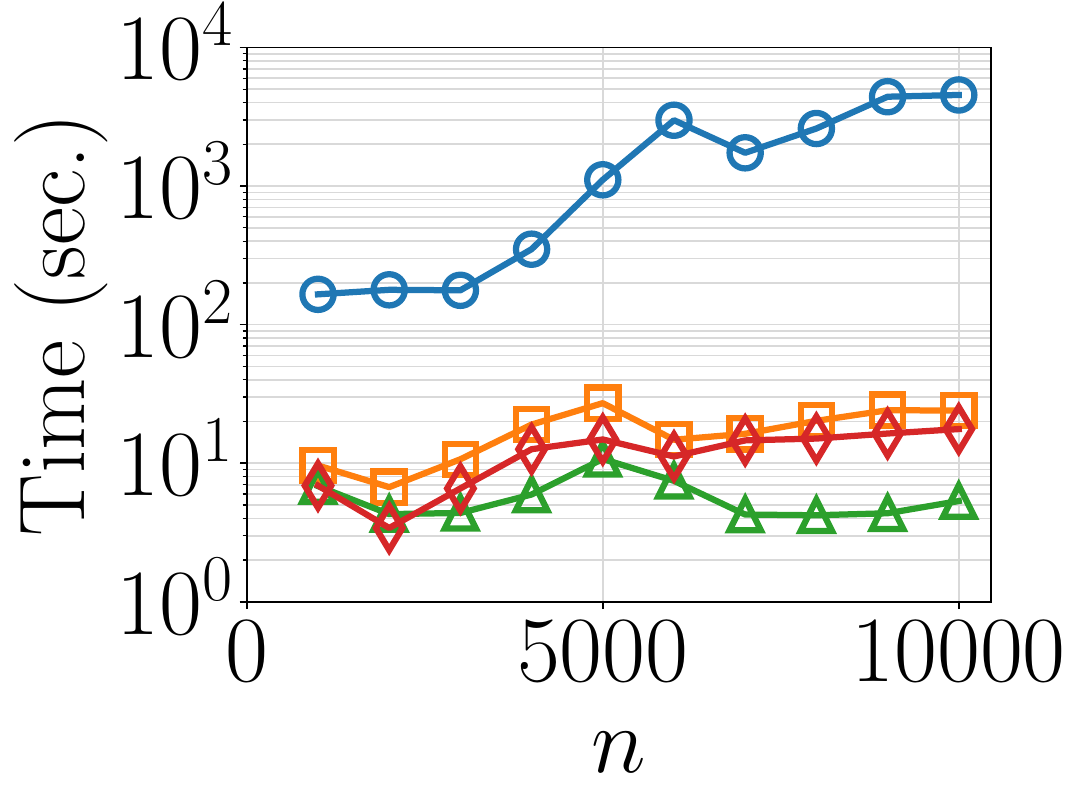}
 }
 \caption{Results of the four LS methods on the ISSP with the linear PF using HV, IGD, R2, and $\epsilon$. The average number of subset evaluations by $\mathcal{I}$ and the wall-clock time of the run are shown in figures (a)--(d) and (e)--(h), respectively. %The results for $d=4$ are shown.
 }
   \label{fig:indicator_effect}
\end{figure*}

We generated a non-dominated point set $P$ of size $n$ for each PF.
First, $P$ was uniformly generated on a linear PF using the method proposed in \cite{BlankDDBS21}, where $\sum^d_{i=1}p_i=1$ for each $p$ in $P$.
Then, $P$ was translated for each PF using the method presented in \cite{TianXZCJ18}.
We set the point set size $n$ to $1\,000, 2\,000, \dots, 10\,000$.
We also set the number of objectives $d$ to $2$, $4$, and $6$.
This work used neither the maximum number of subset evaluations by a quality indicator nor the cut-off time as a termination condition.
%All LS methods are terminated only when they find local optima.
Instead, all LS methods were terminated when there does not exist a pair of points that improves a given quality indicator $\mathcal{I}$.
We performed 31 runs of each LS method.

Section \ref{sec:results} investigates the performance of the following four LS methods:
\begin{description}
\item[\textbf{LS}] \hspace{0.3em} Conventional LS (Section \ref{sec:preliminaries}),
\item[\textbf{LS-N}] \hspace{0.3em} LS using the nearest neighbor list (Section \ref{sec:nbhd_list}),
\item[\textbf{LS-R}] \hspace{0.3em} LS using the random neighbor list (Section \ref{sec:rand_list}),
\item[\textbf{LS-RN}] \hspace{0.3em} LS using the two neighbor lists (Section \ref{sec:clist_2face}).
\end{description}
%

%We set the size of the two neighbor lists to $20$.
\noindent Let $l^{\mathrm{N}}$ be the size of the nearest neighbor list $L^{\mathrm{N}}$.
Let also $l^{\mathrm{R}}$ be the size of the  random neighbor list $L^{\mathrm{R}}$.
%($l^{\mathrm{N}} \coloneqq |l^{\mathrm{N}}|, l^{\mathrm{R}} \coloneqq |l^{\mathrm{R}}|$).
This work used $l^{\mathrm{N}} = 40$ for LS-N, $l^{\mathrm{R}} = 40$ for LS-R, and $l^{\mathrm{N}} = l^{\mathrm{R}} = 20$ for LS-RN ($l^{\mathrm{N}} + l^{\mathrm{R}} = 40$) unless otherwise noted.

\section{Results}
\label{sec:results}

This section describes our analysis results.
Through experiments, we address the following four research questions (RQ2--RQ5) in Sections \ref{sec:eval_scalability}--\ref{sec:analysis_l}, respectively.
Section \ref{sec:further_analysis} shows further investigations.

\begin{enumerate}[RQ1:]
%\begin{enumerate}[\textbf{RQ1}:]
\setcounter{enumi}{1}
\item Can the proposed candidate list strategy speed up LS on the ISSP with a continuous PF?
\item How does the performance of the candidate list strategy scale with respect to the number of objectives $d$ and the subset size $k$?
\item Can the proposed candidate list strategy handle a discontinuous PF?
\item How does the list size $l$ influence the effectiveness of the proposed candidate list strategy?
\end{enumerate}

\subsection{Effectiveness of the candidate list strategy} % as $n$ and $d$ increase}
\label{sec:eval_scalability}

First, we investigate how the performance of LS using the proposed candidate list strategy scales with respect to $n$. % while fixing $d$ and $k$.
For each run of an LS method, we measure $i)$ a quality indicator value $\mathcal{I}(S^{\mathrm{bsf}})$ of the best subset found so far $S^{\mathrm{bsf}}$, $ii)$ the number of subset evaluations by $\mathcal{I}$, and $iii)$ the wall-clock time of the run.

\subsubsection{Quality of subsets found by LS}

Table \ref{tab:comparison_qi} shows the results of the four LS methods (LS, LS-N, LS-R, and LS-RN) on the linear PF in terms of the quality indicator value $\mathcal{I}(S)$ on the ISSP using HV, IGD, R2, and $\epsilon$.
Table \ref{subtab:comparison_qi} shows the results for all seven quality indicators, including IGD$^+$, NR2, and $s$-energy.
Table \ref{subtab:comparison_qi} also shows the the standard deviation of $\mathcal{I}(S^{\mathrm{bsf}})$.
Since Table \ref{subtab:comparison_qi} is similar to Table \ref{tab:comparison_qi}, we do not describe Table \ref{subtab:comparison_qi}.
As seen from Table \ref{tab:comparison_qi}, the four LS methods find subsets with very similar quality.

Below, we discuss the relative performance of LS-N, LS-R, and LS-RN compared to conventional LS in terms of the quality of the subsets.
Let $I^{\mathrm{mean}}_{\mathrm{LS}}$ be the mean of the quality indicator values of the best subsets found so far by an LS method over 31 runs. 
We calculated the relative error for conventional LS and the three proposed methods (LS-N, LS-R, and LS-RN) as follows: $(I_{\mathrm{LS}}^{\mathrm{mean}} - I_{\mathrm{LS}'}^{\mathrm{mean}}) / |I_{\mathrm{LS}}^{\mathrm{mean}}|$, where LS$'$ is either LS-N, LS-R, or LS-RN.
A large relative error suggests that the corresponding LS method finds worse subsets than the conventional LS method.
Note that a negative relative error value can be observed if the corresponding LS method finds better subsets than the conventional LS method.
We conducted the Wilcoxon rank-sum test with $\alpha < 0.05$.
Table \ref{subtab:lsrn-based_stat_linear} shows the results of the statistical tests for LS-RN with respect to LS-N and LS-R.
We do not explain Table \ref{subtab:lsrn-based_stat_linear} in detail, but Table \ref{subtab:lsrn-based_stat_linear} shows that LS-RN performs better than or is comparable to LS-N and LS-R in most cases.

As shown in Tables \ref{tab:comparison_qi} (a), (b), and (c), LS-N and LS-RN perform significantly worse than conventional LS on the ISSP with HV, IGD, and R2.
However, the relative error is $0.897\%$ only in the maximum case.
Here, the previous study \cite{ShangIC21} concluded that the relative error in the range $0.0677\%$ -- $3.5109\%$ is considered ``very small".
Based on this, our results indicate that LS-N and LS-RN can find well-approximated subsets.
As expected, LS-R is outperformed by LS, LS-N, and LS-RN.
%It is natural that swapping only randomly selected points is not effective.
This outcome is natural because swapping only randomly selected points is not effective.
As discussed in Section \ref{sec:clist_2face}, swapping two points close to each other is more effective than swapping randomly selected points on a continuous PF in terms of the quality of the subsets.
However, the relative error for LS-R is $4.87\%$ even in the maximum case. 

As seen from Table \ref{tab:comparison_qi}(d), the results for $\epsilon$ are different from those for other quality indicators.
LS-N, LS-R, and LS-RN achieve significantly worse subsets than LS in terms of $\epsilon$.
For example, the relative error for LS-N is $27\%$ in the maximum case.
%
%LS-N outperforms LS-R in terms of the quality of subsets in most cases.
%It is natural that swapping only randomly selected points is not effective.
LS-R obtains a better quality indicator value than LS-N, except for the results for $n=5\,000$.
In contrast, LS-RN finds better subsets than both LS-N and LS-R in most cases.
This result indicates that the performance of LS-RN is less sensitive to the type of quality indicator compared to LS-N and LS-R.

\subsubsection{Speed improvement by the candidate list}

Fig. \ref{fig:indicator_effect} shows the results of each LS method on the ISSP with the linear PF using HV, IGD, R2, and $\epsilon$.
Fig. \ref{fig:indicator_effect} shows the average of the number of subset evaluations by $\mathcal{I}$ and the wall-clock time of the run. % over 31 runs.
Fig. \ref{supfig:indicator_effect} shows the results for IGD$^+$, NR2, and $s$-energy.
Since the results for IGD$^+$, NR2, and $s$-energy are similar to those for IGD, they are not shown in Fig.~\ref{fig:indicator_effect}.

As seen from the results of LS-N, LS-R, and LS-RN in Figs. \ref{fig:indicator_effect}(a)--(d), the number of subset evaluations is significantly reduced by using the candidate list strategy for all quality indicators.
For example, on the ISSP using HV, the reduction rate of the number of subset evaluations by LS-RN is $96.5$\% for $n=1\,000$ and $99.6$\% for $n=10\,000$.
Thus, the results show that the candidate list strategy can speed up LS in terms of the number of subset evaluations, especially for the ISSP with a large $n$.
Except for $\epsilon$, the results of LS-N, LS-R, and LS-RN in Fig. \ref{fig:indicator_effect} are similar.
This is because the number of examined subsets at each iteration of LS-N, LS-R, and LS-RN is almost the same due to the property of the candidate list strategy.
While LS scans the neighborhood of size $k(n-k)$ at each iteration, LS-N, LS-R, and LS-RN scan only the neighborhood of size $kl$, even in the worst case.
Recall that LS-N, LS-R, and LS-RN use the same $l$.
% In Fig. \ref{fig:indicator_effect}, $l^{\mathrm{N}}$ in LS-N and $l^{\mathrm{R}}$ in LS-R are the same.
% The sum of $l^{\mathrm{N}}$ and $l^{\mathrm{R}}$ in LS-RN is also the same as $l^{\mathrm{N}}$ and $l^{\mathrm{R}}$.
As discussed in Section \ref{sec:analysis_cand_list}, $\epsilon$ has a different property compared to other quality indicators.
This makes the difference between the results of the three LS methods for $\epsilon$.

As shown in Figs. \ref{fig:indicator_effect}(e)--(h), the candidate list strategy can also speed up LS in terms of the computation time.
For example, LS-RN is approximately $29$--$175$ times faster than LS on the ISSP using HV in terms of the computation time.
As seen from Fig. \ref{fig:indicator_effect}, LS-RN is slightly faster than LS-N in terms of both number of subset evaluations and computation time.
This is due to the effect of the first LS-R phase in LS-RN.

In summary, our results show that using the candidate list strategy slightly degrades the quality of subsets found by LS but drastically reduces the computational cost.
It should be noticed that LS with the candidate list strategy can find a better subset through repeatedly performing restarts.

%\begin{tcolorbox}[title=Answers to RQ2, sharpish corners, top=2pt, bottom=2pt, left=4pt, right=4pt, boxrule=0.5pt]
%\textbf{Answers to RQ1:}
\begin{tcolorbox}[sharpish corners, top=2pt, bottom=2pt, left=4pt, right=4pt, boxrule=0.0pt, colback=black!4!white,leftrule=0.75mm,]
\textbf{Answers to RQ2:}
Our results of LS-RN and LS-N demonstrate that the candidate list strategy  can drastically speed up LS in terms of both  number of subset evaluations and  computation time without significantly compromising the quality of subsets on the ISSP using HV, IGD, IGD$^+$, R2, NR2, and $s$-energy.
The candidate list strategy is effective especially when $n$ is large.
%LS-RN performs better than LS-N in terms of the three measurements $i)$--$iii)$.
LS-RN performs better than LS-N in terms of both quality of subsets and computational cost.
However, both LS-RN and LS-N perform poorly in terms of the quality of subsets on the ISSP using $\epsilon$ due to its property observed in Section \ref{sec:analysis_cand_list}.
% However, LS using the candidate list strategy performs poorly in terms of the quality of subsets on the ISSP using $\epsilon$ and R2 due to their property observed in Section \ref{sec:analysis_cand_list}.
Unlike LS-N and LS-RN, LS-R achieves only poor subsets in most cases.
This observation suggests the importance of the nearest neighbor list.  
    
\end{tcolorbox}

%\subsection{Scalability of the candidate list strategy to the number of objectives $d$ and the subset size $k$}
\subsection{Scalability of the candidate list strategy to $d$ and $k$}
\label{sec:scale_k_and_d}

\subsubsection{Scalability to the number of objectives $d$}

\begin{figure}[t]
\centering
\newcommand{\widthvar}{0.235}
\includegraphics[width=0.35\textwidth]{graph/legend1.pdf}\\[-1em]
\subfloat[HV (Num. evals.)]{  
\includegraphics[width=\widthvar\textwidth]{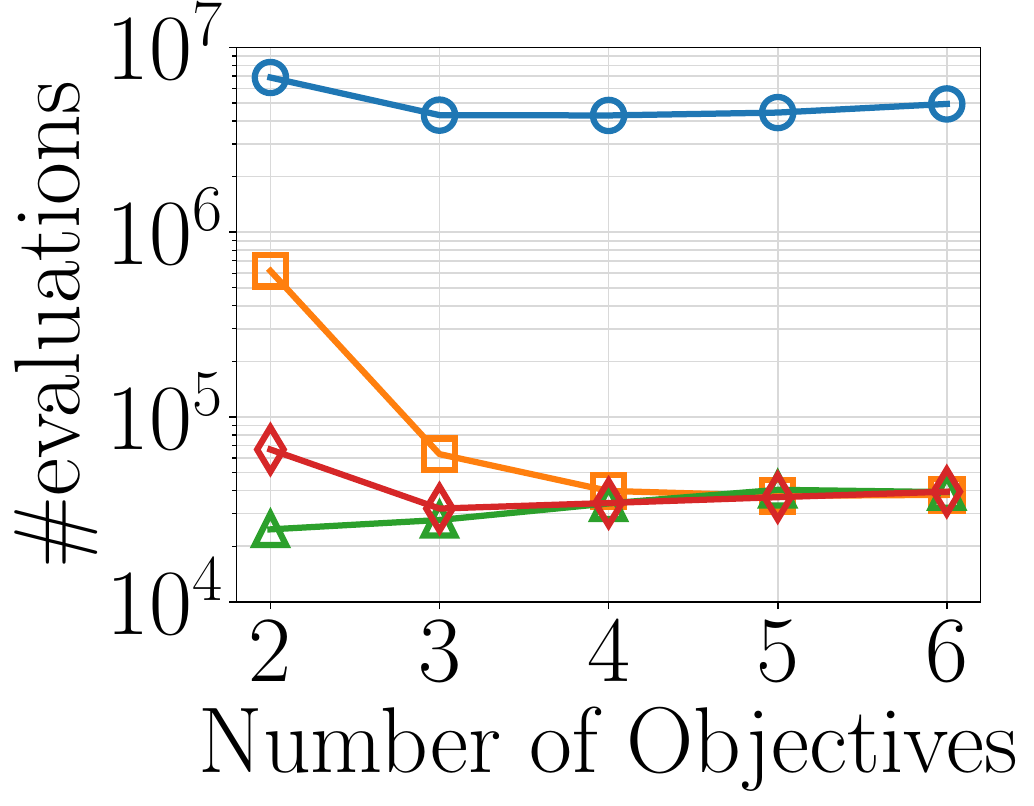}
}
 \subfloat[$\epsilon$ (Num. evals.)]{  
 \includegraphics[width=\widthvar\textwidth]{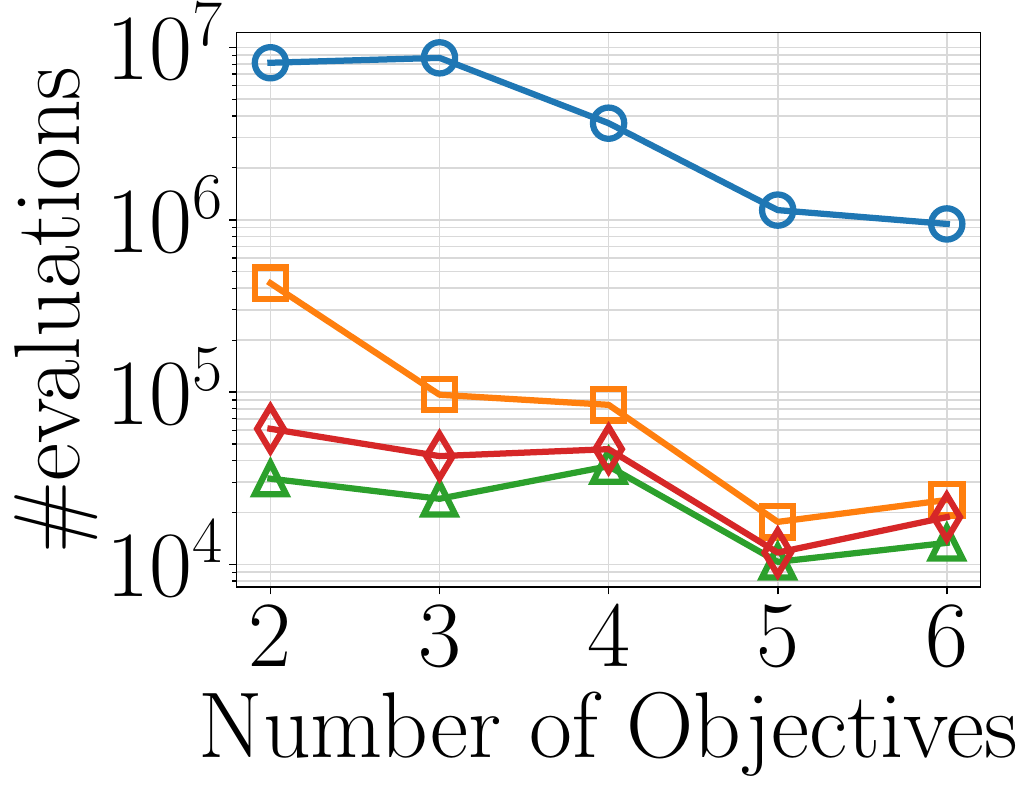}
 }
 \\
\caption{
Scalability of the four LS methods to the number of objectives $d$ on the ISSP with the linear PF using HV and $\epsilon$.
%The results of the four LS methods on the ISSP with the linear PF using HV and $\epsilon$.
%, where (\tabblue{$\bigcirc$}: LS, \tabgreen{$\triangle$}: LS-R, \taborange{$\square$}: LS-N, and \tabred{$\Diamond$}: LS-RN.
%The results for $n=5\,000$ are shown.
}
   \label{fig:scale_d}
\end{figure}

Fig. \ref{fig:scale_d} shows the average results of the four LS methods on the ISSP with $n=5\,000$, $k=100$, and $d \in \{2, 3, 4, 5, 6\}$ using HV and $\epsilon$.
Since the results are consistent across values of $n$, we show only those for $n=5\,000$. 
For all seven quality indicators, Table \ref{suptable:scale_d} shows the results in terms of the quality indicator value.
In addition, Fig. \ref{supfig:scale_d} shows the results in terms of the number of subset evaluations and the computation time.
We do not describe Table \ref{suptable:scale_d} in detail, but Table \ref{suptable:scale_d} is similar to Table \ref{tab:comparison_qi}.
Thus, the results show that LS-RN and LS-N find slightly worse subsets than the conventional LS in most cases, except on the ISSP using $\epsilon$.
We also do not describe Fig. \ref{supfig:scale_d} in detail, but Fig. \ref{supfig:scale_d} is similar to Fig. \ref{fig:scale_d}.
%We also do not describe Fig. \ref{supfig:scale_d} in detail, but the results for IGD, IGD$^+$, R2, NR2, and $s$-energy are similar to the results for HV.
%The results for R2 are also similar to the results for $\epsilon$.

%Table \ref{suptable:_scale_d} is similar to Table \ref{tab:comparison_qi}.
%Similar to Table \ref{tab:comparison_qi}, Table \ref{suptable:scale_d} shows the relative error compared to conventional LS in terms of the quality of subsets.

% We do not describe Table \ref{suptable:scale_d} in detail, but interestingly, LS-R finds a good subset in terms of HV and IGD as $d$ increases.
% However, LS-R performs poorly for R2 and $\epsilon$.
% LS-N finds better subsets than LS-R on the ISSP using HV, IGD, and $\epsilon$ in most cases.
% LS-RN outperforms LS-N and LS-R for any quality indicator and any $d$.

As seen from Fig. \ref{fig:scale_d}, LS-R terminates the search earlier than LS-N and LS-RN for smaller values of $d$.
This suggests that LS-R achieves fewer successful point swaps compared to LS-N and LS-RN.
%the number of successful swaps of points in LS-R is smaller than that in LS-N and LS-RN.
%
In Fig. \ref{fig:scale_d}, LS-R, LS-N, and LS-RN perform similarly in terms of the number of subset evaluations as $d$ increases.
Contrary to intuition, the number of subset evaluations in all four LS methods decreases as $d$ increases.
As shown in Figs. \ref{supfig:scale_d}(a)--(g), similar results can be observed on the ISSPs using other quality indicators.
Similar to Figs. \ref{fig:indicator_effect}(e)--(h), Figs. \ref{supfig:scale_d}(h)--(n) show the results regarding the computation time.
The results for HV in Fig. \ref{supfig:scale_d}(h) show that all four LS methods require higher computation time as $d$ increases.
However, this observation is not inconsistent with Fig. \ref{fig:scale_d}(a). 
This is because the computation time of HV grows exponentially with $d$.

%\vspace{1em}
\begin{figure}[t]
\centering
\newcommand{\widthvar}{0.235}
\includegraphics[width=0.35\textwidth]{graph/legend1.pdf}\\[-1em]
\subfloat[HV (Num. evals.)]{  
\includegraphics[width=\widthvar\textwidth]{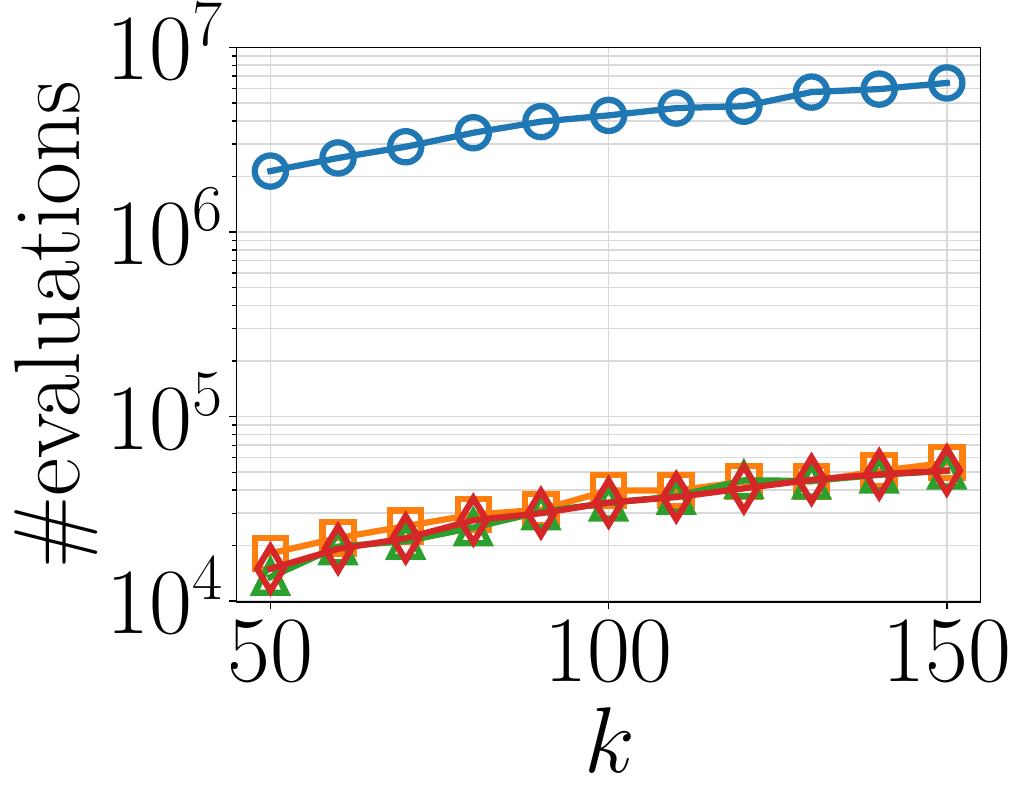}
}
 \subfloat[$\epsilon$ (Num. evals.)]{  
 \includegraphics[width=\widthvar\textwidth]{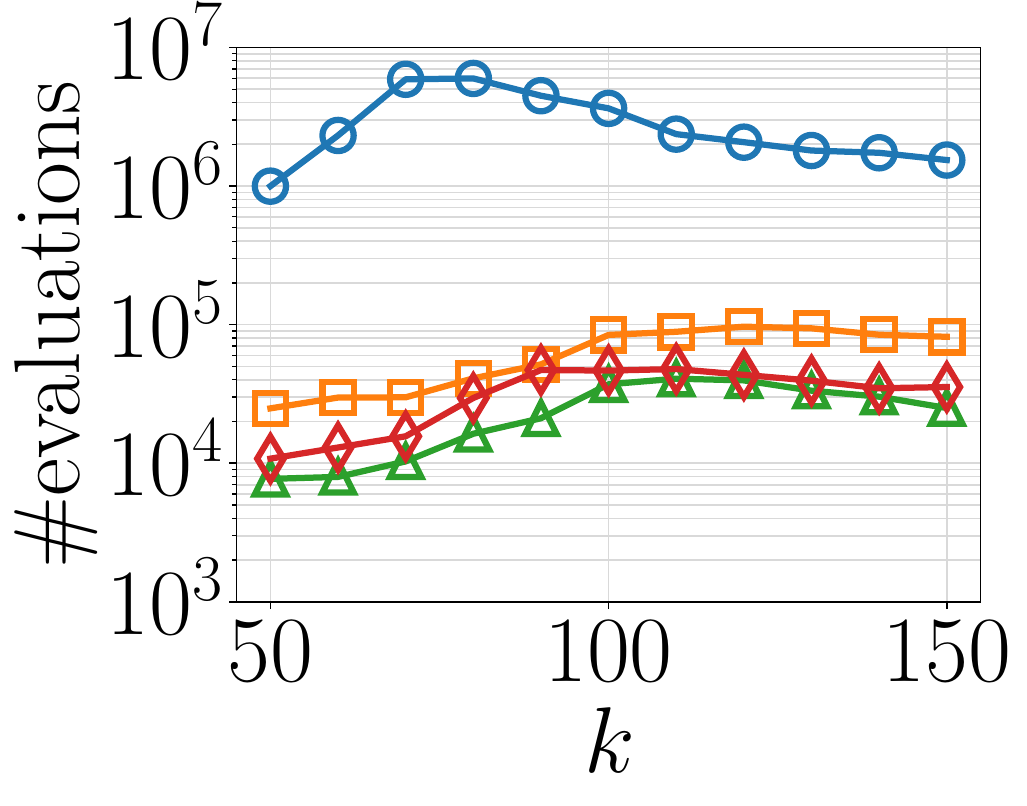}
 }
 \\
\caption{
Scalability of the four LS methods to the subset size $k$ on the ISSP with the linear PF using HV and $\epsilon$.
%, where (\tabblue{$\bigcirc$}: LS, \tabgreen{$\triangle$}: LS-R, \taborange{$\square$}: LS-N, and \tabred{$\Diamond$}: LS-RN.
%The results for $d=4, n=5\,000$ are shown.
}
   \label{fig:scale_k}
\end{figure}

\begin{figure*}[t]
\captionsetup[subfloat]{farskip=2pt,captionskip=1pt}
\centering
\newcommand{\widthvar}{0.235}
\includegraphics[width=0.4\textwidth]{graph/legend1.pdf}\\ %[-1em]
\subfloat[HV (Num. evals.)]{  
\includegraphics[width=\widthvar\textwidth]{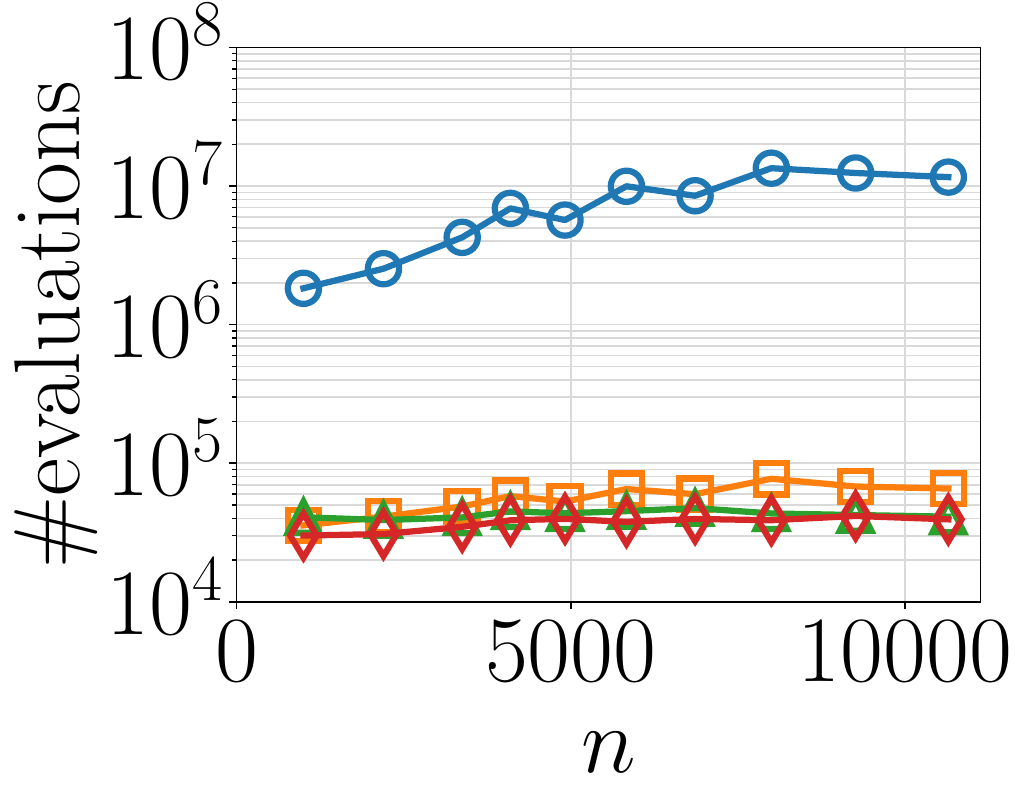}
}
\subfloat[IGD (Num. evals.)]{  
\includegraphics[width=\widthvar\textwidth]{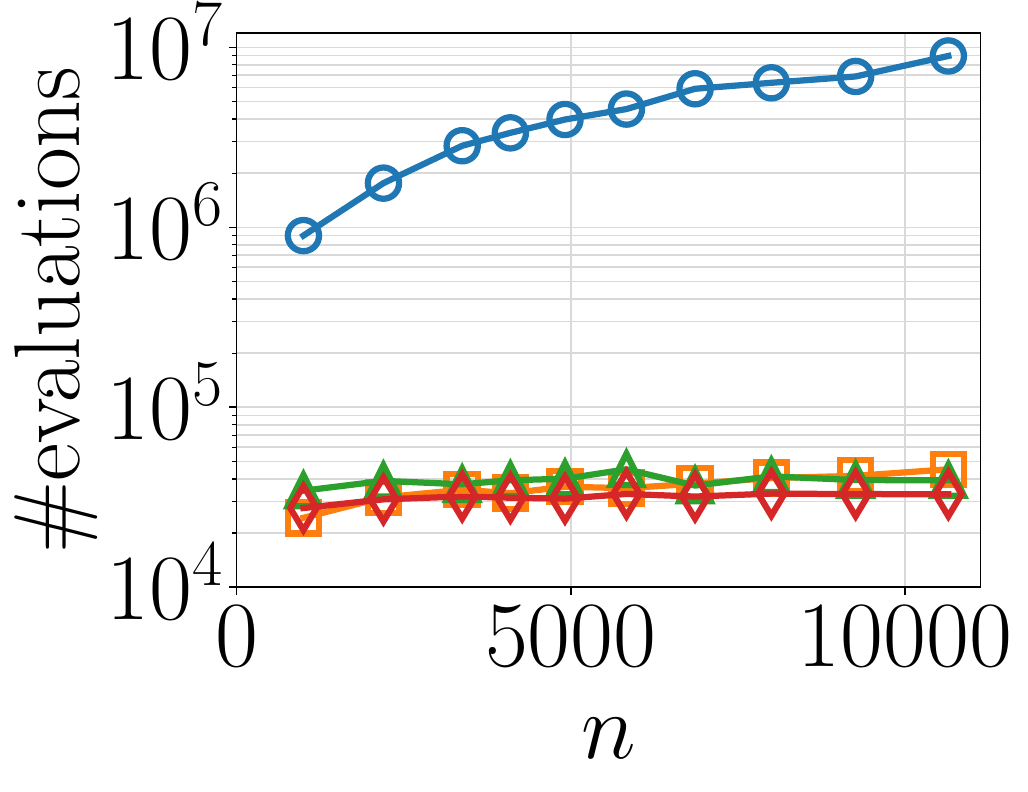}
}
\subfloat[R2 (Num. evals.)]{  
\includegraphics[width=\widthvar\textwidth]{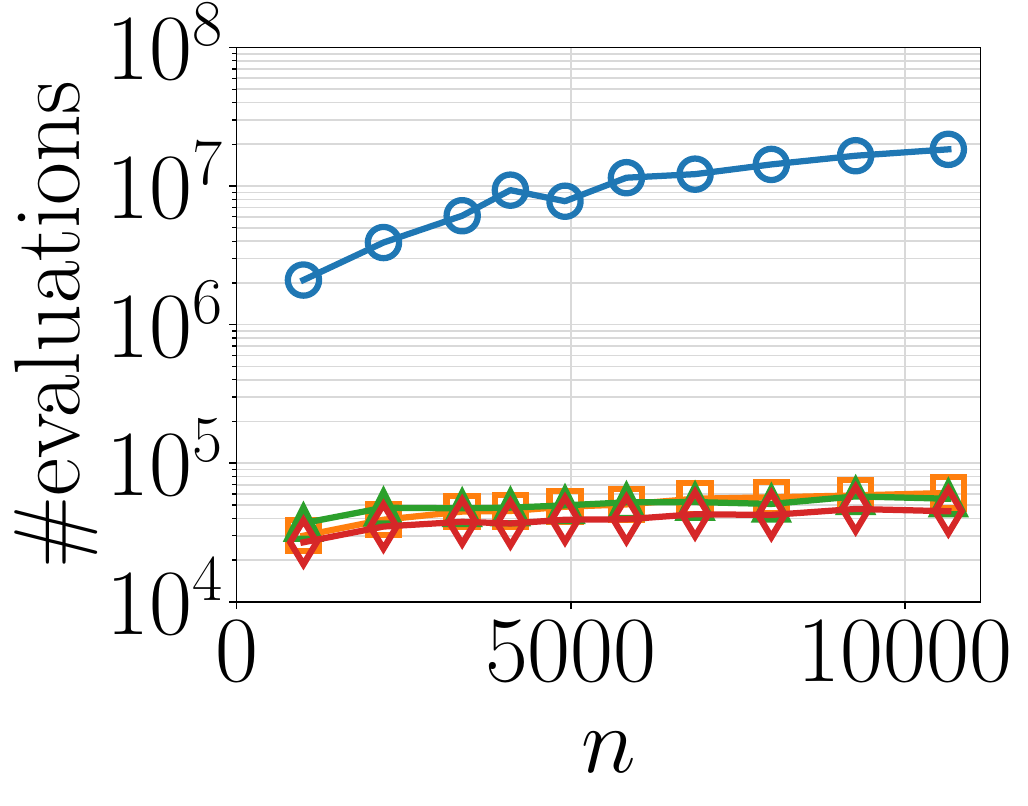}
}
 \subfloat[$\epsilon$ (Num. evals.)]{  
 \includegraphics[width=\widthvar\textwidth]{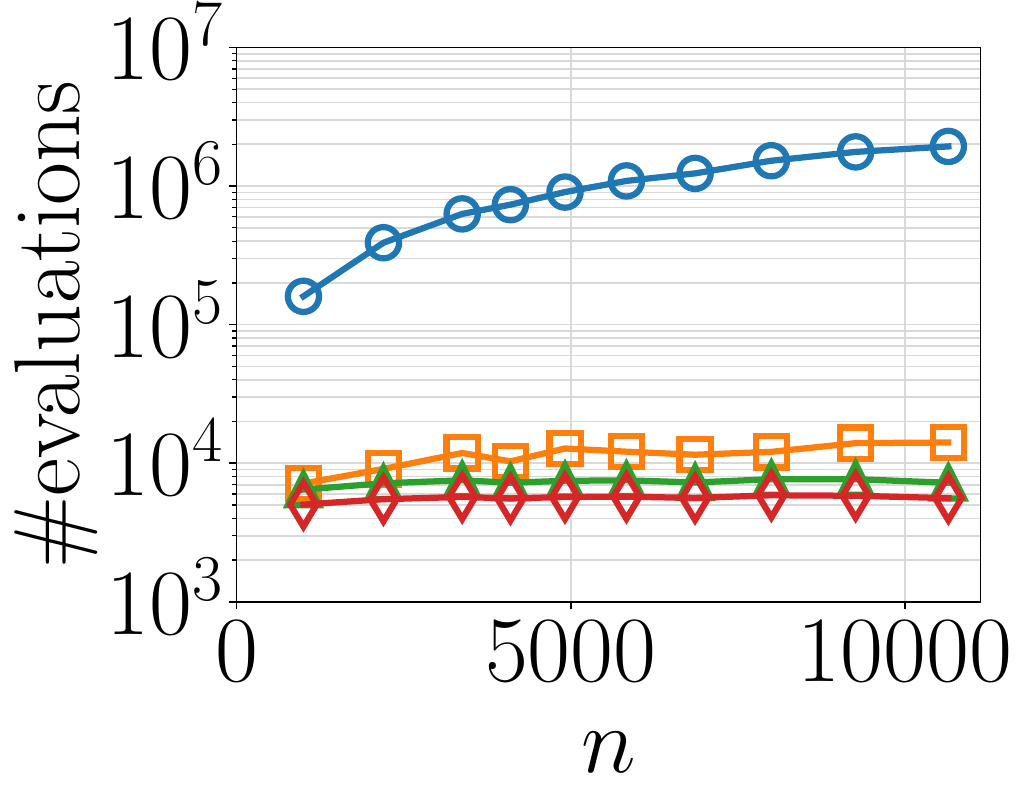}
 }
 \\
\subfloat[HV (Time)]{  
\includegraphics[width=\widthvar\textwidth]{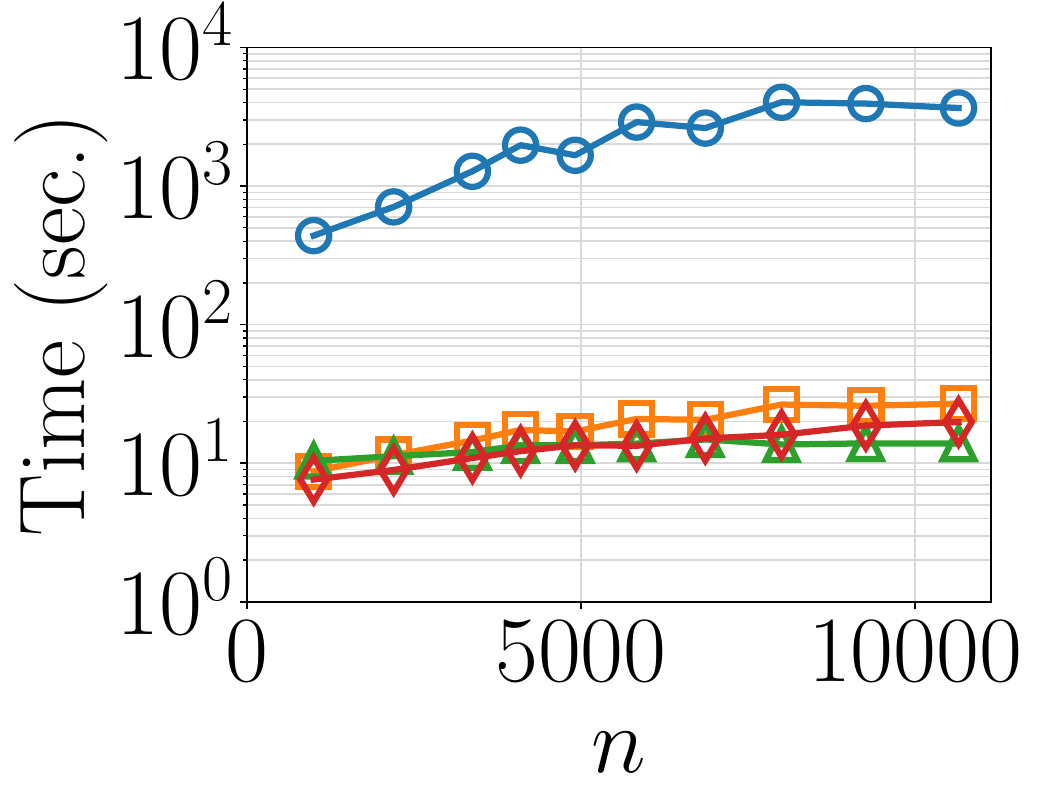}
}
\subfloat[IGD (Time)]{  
\includegraphics[width=\widthvar\textwidth]{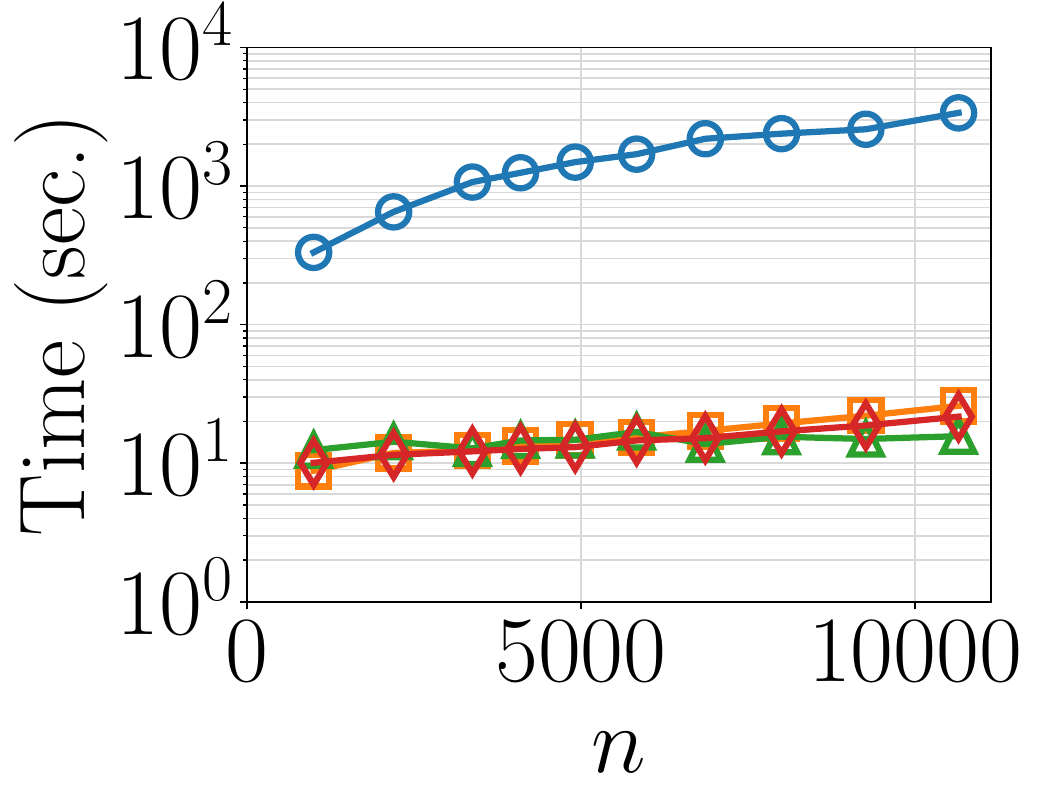}
}
\subfloat[R2 (Time)]{  
\includegraphics[width=\widthvar\textwidth]{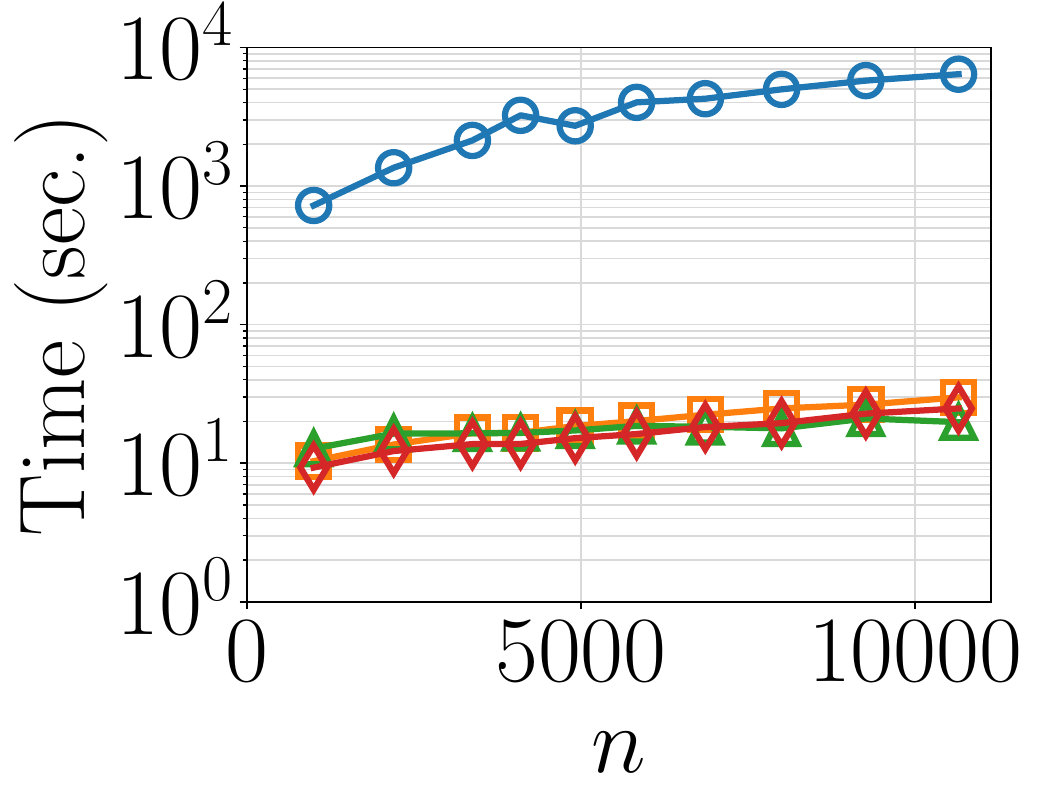}
}
 \subfloat[$\epsilon$ (Time)]{  
 \includegraphics[width=\widthvar\textwidth]{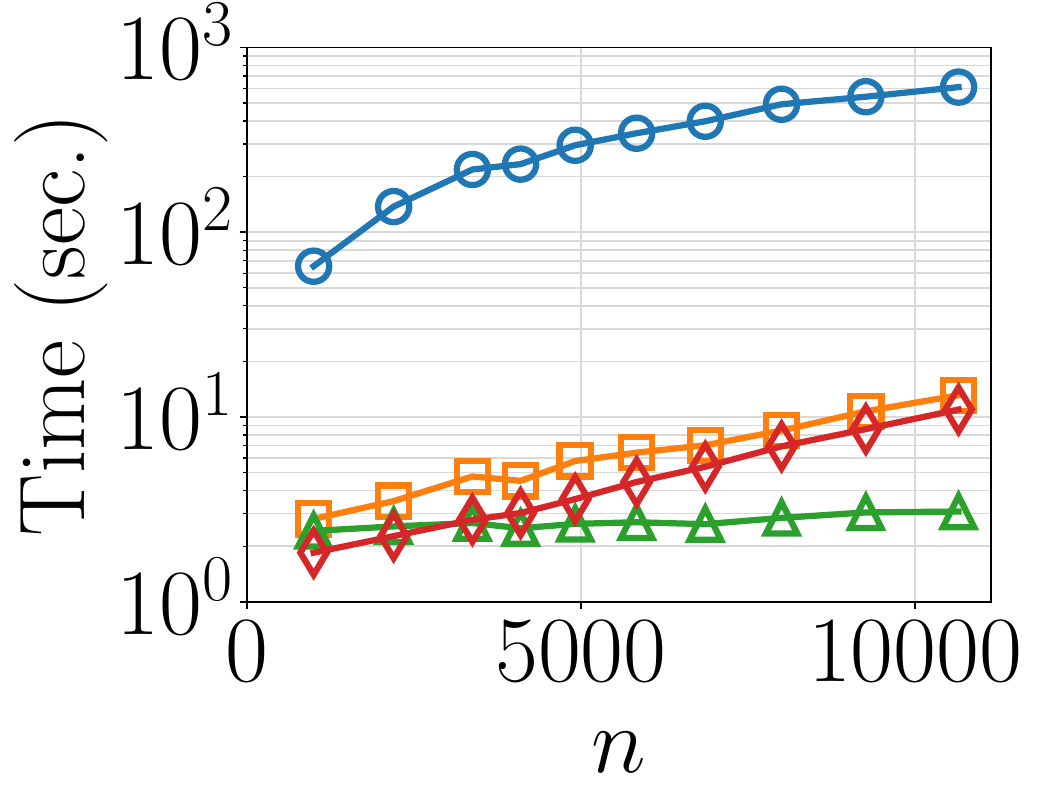}
 }
 \caption{
 Results of the four LS methods on the ISSP with the discontinuous PF using HV, IGD, R2, and $\epsilon$.
 %, where (\tabblue{$\bigcirc$}: LS, \tabgreen{$\triangle$}: LS-R, \taborange{$\square$}: LS-N, and \tabred{$\Diamond$}: LS-RN.
 %The results for $d=4$ are shown.
 }
   \label{fig:disconneted_effect}
\end{figure*}

\begin{table*}[t]
\setlength{\tabcolsep}{5pt} % Default value: 6pt
  \renewcommand{\arraystretch}{0.9} 
\centering
  \caption{\small Comparison of the four LS methods on the ISSP instances with the discontinuous PF using HV. The average quality indicator values of subsets found by the four methods are shown.
  }
  \label{tab:comparison_qi_discontinuous}
%{\scriptsize
{\footnotesize
%{\small
%%%%%%%%%%%
%\subfloat[HV]{
\begin{tabular}{ccccc}
\toprule
$n$ & LS & LS-N & LS-R & LS-RN\\  
\midrule
 $1$K & \cellcolor{c1}\Enote{5.90}{-01} & \Enote{5.83}{-01}\MSsymbol (\sgnPS\Enote{1.09}{+00}\%) & \Enote{5.88}{-01}\MSsymbol (\sgnPS\Enote{3.55}{-01}\%) & \cellcolor{c2}\Enote{5.89}{-01}\MSsymbol (\sgnPS\Enote{7.39}{-02}\%) \\
 $2$K & \cellcolor{c1}\Enote{5.94}{-01} & \Enote{5.89}{-01}\MSsymbol (\sgnPS\Enote{8.84}{-01}\%) & \Enote{5.91}{-01}\MSsymbol (\sgnPS\Enote{5.25}{-01}\%) & \cellcolor{c2}\Enote{5.93}{-01}\MSsymbol (\sgnPS\Enote{8.98}{-02}\%) \\
 $3$K & \cellcolor{c1}\Enote{5.94}{-01} & \Enote{5.89}{-01}\MSsymbol (\sgnPS\Enote{8.53}{-01}\%) & \Enote{5.91}{-01}\MSsymbol (\sgnPS\Enote{5.61}{-01}\%) & \cellcolor{c2}\Enote{5.94}{-01}\MSsymbol (\sgnPS\Enote{7.23}{-02}\%) \\
 $4$K & \cellcolor{c1}\Enote{5.94}{-01} & \Enote{5.86}{-01}\MSsymbol (\sgnPS\Enote{1.29}{+00}\%) & \Enote{5.90}{-01}\MSsymbol (\sgnPS\Enote{6.72}{-01}\%) & \cellcolor{c2}\Enote{5.93}{-01}\MSsymbol (\sgnPS\Enote{7.77}{-02}\%) \\
 $4$K & \cellcolor{c1}\Enote{5.95}{-01} & \Enote{5.89}{-01}\MSsymbol (\sgnPS\Enote{9.63}{-01}\%) & \Enote{5.91}{-01}\MSsymbol (\sgnPS\Enote{6.36}{-01}\%) & \cellcolor{c2}\Enote{5.94}{-01}\MSsymbol (\sgnPS\Enote{7.10}{-02}\%) \\
 $5$K & \cellcolor{c1}\Enote{5.94}{-01} & \Enote{5.87}{-01}\MSsymbol (\sgnPS\Enote{1.20}{+00}\%) & \Enote{5.90}{-01}\MSsymbol (\sgnPS\Enote{7.16}{-01}\%) & \cellcolor{c2}\Enote{5.94}{-01}\MSsymbol (\sgnPS\Enote{8.93}{-02}\%) \\
 $6$K & \cellcolor{c1}\Enote{5.95}{-01} & \Enote{5.89}{-01}\MSsymbol (\sgnPS\Enote{1.05}{+00}\%) & \Enote{5.91}{-01}\MSsymbol (\sgnPS\Enote{6.77}{-01}\%) & \cellcolor{c2}\Enote{5.94}{-01}\MSsymbol (\sgnPS\Enote{7.63}{-02}\%) \\
 $8$K & \cellcolor{c1}\Enote{5.94}{-01} & \Enote{5.87}{-01}\MSsymbol (\sgnPS\Enote{1.27}{+00}\%) & \Enote{5.90}{-01}\MSsymbol (\sgnPS\Enote{7.46}{-01}\%) & \cellcolor{c2}\Enote{5.94}{-01}\MSsymbol (\sgnPS\Enote{8.06}{-02}\%) \\
 $9$K & \cellcolor{c1}\Enote{5.95}{-01} & \Enote{5.89}{-01}\MSsymbol (\sgnPS\Enote{1.02}{+00}\%) & \Enote{5.91}{-01}\MSsymbol (\sgnPS\Enote{7.23}{-01}\%) & \cellcolor{c2}\Enote{5.94}{-01}\MSsymbol (\sgnPS\Enote{8.16}{-02}\%) \\
$10$K & \cellcolor{c1}\Enote{5.95}{-01} & \Enote{5.90}{-01}\MSsymbol (\sgnPS\Enote{7.39}{-01}\%) & \Enote{5.91}{-01}\MSsymbol (\sgnPS\Enote{7.13}{-01}\%) & \cellcolor{c2}\Enote{5.94}{-01}\MSsymbol (\sgnPS\Enote{6.88}{-02}\%) \\
\toprule
\end{tabular}
%}
%%%%%%%%%%%
}
\end{table*}

\subsubsection{Scalability to the subset size  $k$}

Fig. \ref{fig:scale_k} shows the average results of the four LS methods on the ISSP with $d=4$, $n=5\,000$, and $k \in \{50, 60, \dots, 150\}$ using HV and $\epsilon$ in terms of the number of subset evaluations. % and computation time. 
Fig. \ref{supfig:scale_k} shows the results for other quality indicators.
The results for IGD, IGD$^+$, R2, NR2, and $s$-energy are similar to those for HV in Fig. \ref{fig:scale_k}.
Table \ref{suptab:scale_k} shows the quality of subsets for $k \in \{50, 60, \ldots, 150\}$.
Similar to Table \ref{tab:comparison_qi}, Table \ref{suptab:scale_k} shows that LS-N and LS-RN find subsets with acceptable quality.

As seen from Fig. \ref{fig:scale_k}(a), the number of subset evaluations in all four LS methods increases as $k$ increases.
However, Fig. \ref{fig:scale_k}(a) demonstrates that the candidate list strategy significantly reduces the number of subset evaluations in LS for any $k$.
%This is simply because the size of the search space of the ISSP depends on $k$ as described in Section \ref{sec:issp}.
As shown in Fig. \ref{fig:scale_k}(b), for $\epsilon$, the number of subset evaluations in LS decreases as $k$ increases from $70$ to $150$.
LS with the candidate list strategy also shows a similar trend.
However, for any $k$, the number of subset evaluations in LS with the candidate list strategy is significantly smaller than that in conventional LS.

%\begin{tcolorbox}[title=Answers to RQ3, sharpish corners, top=2pt, bottom=2pt, left=4pt, right=4pt, boxrule=0.5pt]
%\textbf{Answers to RQ1:}
\begin{tcolorbox}[sharpish corners, top=2pt, bottom=2pt, left=4pt, right=4pt, boxrule=0.0pt, colback=black!4!white,leftrule=0.75mm,]
\textbf{Answers to RQ3:}
Our results show that LS-RN is well-scalable to the number of objectives $d$ and the subset size $k$.
Our results suggest that LS-RN is effective especially for a large $d$.
LS-RN can significantly reduce the computation time especially on the ISSP using HV.
We observed that LS-N performs worse than LS-R in terms of the number of subset evaluations and computation time when $d$ is small.
We also found that these differences between LS-N, LS-R, and LS-RN become small as $d$ increases.
Overall, our results show that LS-RN generally obtains a good subset for any $\mathcal{I}$, $d$, and $k$ compared to LS-R and LS-N.
The results indicate the importance of using the two neighbor lists sequentially.

\end{tcolorbox}

\subsection{Effectiveness of the candidate list strategy on the discontinuous PF}
\label{sec:eval_Is_for_disconnected}

%Section \ref{sec:eval_scalability} shows the effectiveness of the candidate list strategy on the ISSP with the continuous PF.
Unlike Section \ref{sec:eval_scalability}, we here investigate  the effectiveness of the candidate list strategy on an ISSP  with a discontinuous PF.
We used the discontinuous PF of the DTLZ7 problem~\cite{DebTLZ05}.
We normalized the PF of the DTLZ problem into $[0,1]^d$ so that the scale of objectives is the same as in other PFs.
We uniformly generated Pareto optimal points on the PF by using the method proposed in \cite{TianXZCJ18}.
%Since this generation method
Due to the combinatorial property of this generation method, we set the point set size $n$ to $1\,000$, $2\,197$, $3\,375$, $4\,096$, $4\,913$, $5\,832$, $6\,859$, $8\,000$, $9\,261$, and $10\,648$.
Here, we set $d$ to $4$.
Table \ref{subtab:lsrn-based_stat_disconnected} shows the results of the statistical tests for LS-RN with respect to LS-N and LS-R.
Although we do not explain Table \ref{subtab:lsrn-based_stat_disconnected} in detail, it shows that LS-RN performs significantly better than LS-N and LS-R, except for the results for $\epsilon$.

Fig. \ref{fig:disconneted_effect} shows the average results of the four LS methods on the ISSP with the discontinuous PF using HV, IGD, R2, and $\epsilon$ in terms of the number of subset evaluations and computation time.
We show the results for IGD$^+$, NR2, and $s$-energy in Fig. \ref{supfig:disconnected_effect} for the same reason as in Fig. \ref{fig:indicator_effect}.
Fig. \ref{fig:disconneted_effect} is similar to Fig. \ref{fig:indicator_effect}.
Thus, as shown in Fig. \ref{fig:disconneted_effect}, the candidate list strategy can speed up LS even on the discontinuous PF.

Table \ref{tab:comparison_qi_discontinuous} shows the comparison of the four LS methods in terms of the quality of the subsets on the ISSP using HV.
Table \ref{suptab:comparison_qi_discontinuous} shows the results for other quality indicators, but they are similar to Table \ref{tab:comparison_qi_discontinuous}.
As shown in Table \ref{tab:comparison_qi_discontinuous}, LS-N performs significantly worse than conventional LS in terms of the quality of subsets.
For example, as shown in Table \ref{tab:comparison_qi}(a), the relative error of LS-N compared to LS is only $6.42 \times 10^{-3} \%$ on the ISSP with the linear PF and $n=1\,000$ using HV.
In contrast, that is $1.09 \%$ on the ISSP with the discontinuous PF.
Interestingly, LS-N obtains worse subsets than LS-R on the ISSP using HV, where similar results are found on the ISSP using R2 and $\epsilon$.

As discussed in Section \ref{sec:rand_list}, the poor performance of LS-N for the discontinuous PF comes from the difficulty in swapping points on different subsets of the PF.
LS-RN addresses this issue of LS-N by using the random neighbor and the nearest neighbor lists sequentially.
As seen from Table \ref{tab:comparison_qi_discontinuous} and Fig. \ref{fig:disconneted_effect}, LS-RN finds better subsets than LS-N while reducing the computation cost.

%\begin{tcolorbox}[title=Answers to RQ4, sharpish corners, top=2pt, bottom=2pt, left=4pt, right=4pt, boxrule=0.5pt]
%\textbf{Answers to RQ1:}
\begin{tcolorbox}[sharpish corners, top=2pt, bottom=2pt, left=4pt, right=4pt, boxrule=0.0pt, colback=black!4!white,leftrule=0.75mm,]
\textbf{Answers to RQ4:}
    Our results show that LS-N and LS-R perform poorly on the ISSP with the discontinuous PF.
    In contrast, we observed that LS-RN can address the issue of LS-N and LS-R for the discontinuous PF.
    Our results show that LS-RN obtains better subsets than LS-N and LS-R while reducing the number of subset evaluations and computation time. 
    These observations indicate the importance of using two neighbor lists with different properties sequentially to handle discontinuous PFs.
\end{tcolorbox}

\subsection{Effects of $l$}
\label{sec:analysis_l}

%, especially the nearest neighbor list size $l^{\mathrm{N}}$. 

Here, we investigate the effects of the candidate list size, including the nearest neighbor list size $l^{\mathrm{N}}$ and the random neighbor list size $l^{\mathrm{R}}$.
First, this section investigates the influence of $l^{\mathrm{N}}$ on the performance of LS-N.
Then, this section investigates the effects of the ratio of $l^{\mathrm{N}}$ and $l^{\mathrm{R}}$ in LS-RN.

\subsubsection{Effects of $l^{\mathrm{N}}$ in LS-N}

%We set the list size $l^{\mathrm{N}}$ in LS-N to $40$ in Sections \ref{sec:eval_scalability}--\ref{sec:eval_Is_for_disconnected}.
%Here, we investigate the influence of $l^{\mathrm{N}}$ on the performance of LS-N.

\begin{figure}[t]
\centering
\subfloat[HV (Num. evals.)]{  
\includegraphics[width=0.23\textwidth]{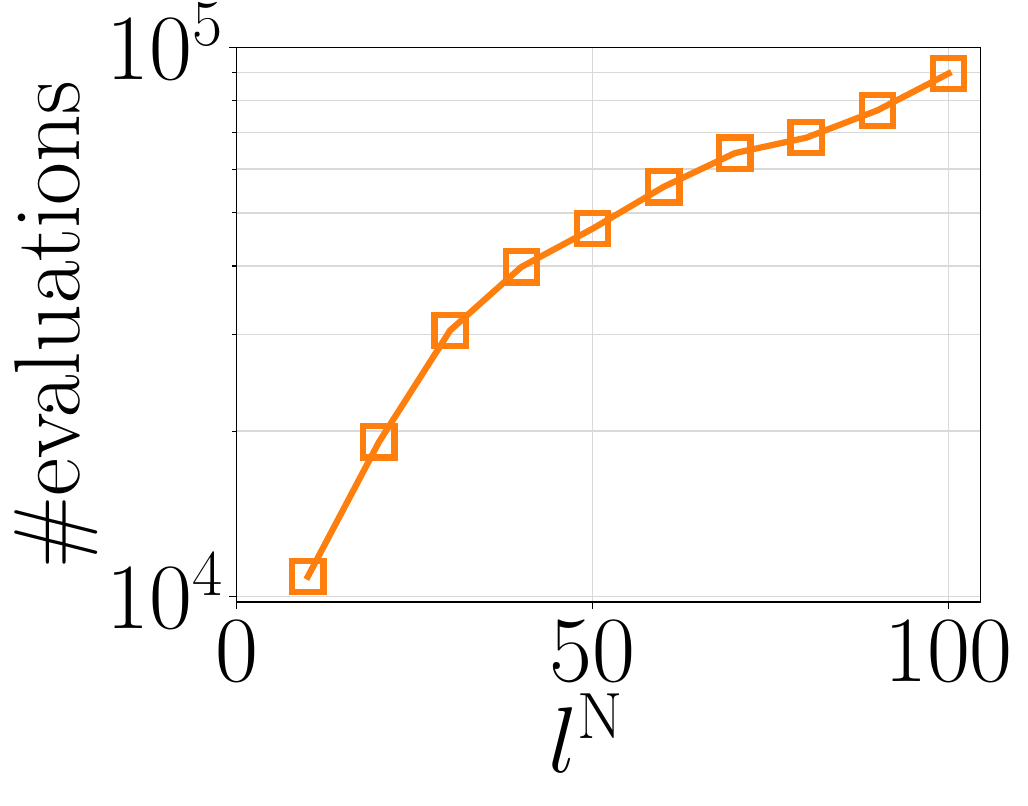}
}
\subfloat[$\epsilon$ (Num. evals.)]{  
\includegraphics[width=0.23\textwidth]{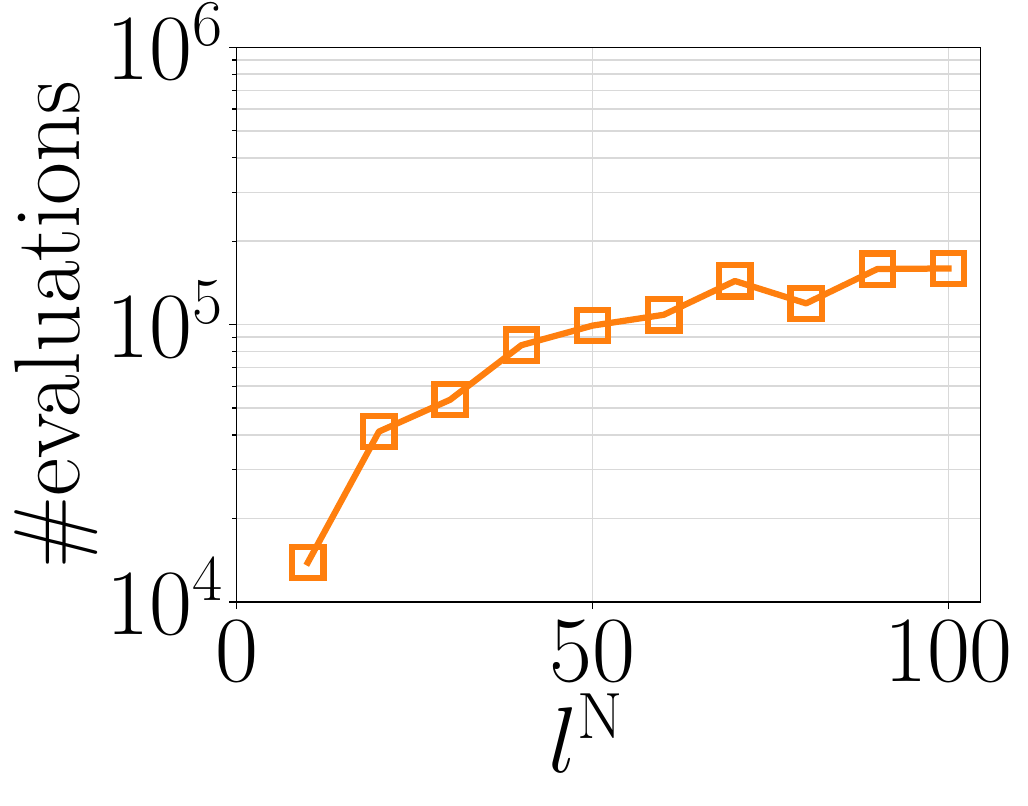}
}
% \\
% \subfloat[HV ($\mathcal{I}(S)$)]{  
% \includegraphics[width=0.235\textwidth]{graph/clist_size/clist_d2_I-hv_v-qi.pdf}
% }
% \subfloat[HV (Num. evals.)]{  
% \includegraphics[width=0.235\textwidth]{graph/clist_size/clist_d2_I-hv_v-n_qi_calls.pdf}
% }
% \subfloat[HV (Time)]{  
% \includegraphics[width=0.235\textwidth]{graph/clist_size/clist_d2_I-hv_v-time.pdf}
% }
\caption{Results of LS-N with $l^{\mathrm{N}} \in \{10, 20, \dots, 100\}$ on the ISSP  using HV and $\epsilon$. %The results for the linear PF, $d = 4, n = 5\,000,$ and $ k = 100$ are shown.
}
  \label{fig:clist_size}
\end{figure}

Fig. \ref{fig:clist_size} shows the results of LS-N with $l^{\mathrm{N}} \in \{10, 20, \dots, 100\}$ on the ISSP with $d = 4$, $n = 5\,000$, and $k = 100$ using HV and $\epsilon$.
Fig. \ref{fig:clist_size} shows the average number of subset evaluations required to complete the search.
Fig. \ref{supfig:clist_size} shows the results for all seven quality indicators.
Table \ref{subtab:clist_size} also shows the quality of subsets found by LS-N.

We do not describe Table \ref{subtab:clist_size} in detail, but it shows that LS-N with $l^{\mathrm{N}} = 10$ achieves the poorest performance in terms of the quality of subsets.
We can see that LS-N obtains better subsets as $l^{\mathrm{N}}$ increases.
In contrast, as shown in Fig. \ref{fig:clist_size}, the number of subset evaluations in LS-N increases as $l^{\mathrm{N}}$ increases.
These observations indicate that $l^{\mathrm{N}}$ in LS-N controls the trade-off between the quality of subsets and computational cost (i.e., the number of subset evaluations and computation time).
As seen from Table \ref{subtab:clist_size}, the quality of subsets is not drastically improved when setting $l^{\mathrm{N}}$ to more than $30$.
Although we set $l^{\mathrm{N}}$ to $40$ throughout this paper, this observation shows that this setting is reasonable.
Note that similar trends were observed for $l^{\mathrm{R}}$ in LS-R.

\subsubsection{Effects of the ratio of $l^{\mathrm{N}}$ and $l^{\mathrm{R}}$ in LS-RN}

This paper set the sizes of the nearest neighbor list $l^{\mathrm{N}}$ and random neighbor list $l^{\mathrm{R}}$ to $20$.
Here, we investigate how the ratio of $l^{\mathrm{N}}$ and $l^{\mathrm{R}}$ influences the performance of LS-RN.
In this experiment, we fixed the sum of $l^{\mathrm{N}}$ and $l^{\mathrm{R}}$ at $80$ in LS-RN.
Then, we changed $l^{\mathrm{N}}$ as follows: $l^{\mathrm{N}} \in \{0, 10, \dots, 80\}$.
For example, $l^{\mathrm{R}}=70$ when $l^{\mathrm{N}}=10$.
Note that LS-RN with $l^{\mathrm{N}} = 0$ and $l^{\mathrm{N}} = 80$ are exactly the same as LS-R and LS-N, respectively. 

Table \ref{subtab:list_size_ratio} shows the quality of subsets found by LS and LS-RN with various ratios of $l^{\mathrm{N}}$ and $l^{\mathrm{R}}$ on the ISSP using the seven quality indicators.
Table \ref{subtab:list_size_ratio} shows the results for the linear PF, $d = 4, n = 5\,000$, and $ k = 100$.
We do not provide a detailed explanation of Table \ref{subtab:list_size_ratio}, but the results show that extremely small and large $l^{\mathrm{N}}$ values deteriorate the performance of LS-RN, e.g., $l^{\mathrm{N}} = 0$ and $l^{\mathrm{N}}=80$.
%we can summarize Table \ref{subtab:list_size_ratio} as follows.
%The worst performance of LS-RN on the ISSP using HV is observed when setting $l^{\mathrm{N}}$ to $0$. 

\begin{figure}[t]
\centering
\newcommand{\widthvar}{0.23}
%\includegraphics[width=0.16\textwidth]{graph/legend2.pdf}
% \subfloat[HV (RE)]{  
% \includegraphics[width=\widthvar\textwidth]{graph/list_size_ratio/ratio_pf-linear_v-qi_ratio_I-hv.pdf}
% }
%  \subfloat[$\epsilon$ (RE)]{  
%  \includegraphics[width=\widthvar\textwidth]{graph/list_size_ratio/ratio_pf-linear_v-qi_ratio_I-epsilon.pdf}
%  }
\subfloat[HV (Num. evals.)]{  
\includegraphics[width=\widthvar\textwidth]{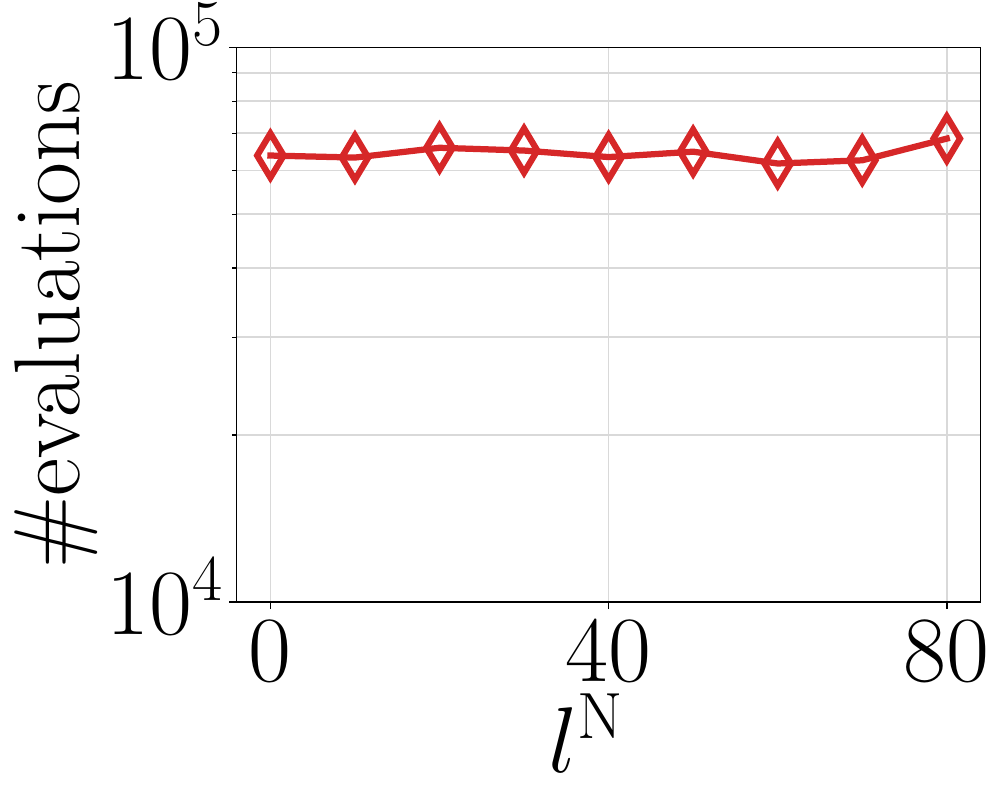}
}
 \subfloat[$\epsilon$ (Num. evals.)]{  
 \includegraphics[width=\widthvar\textwidth]{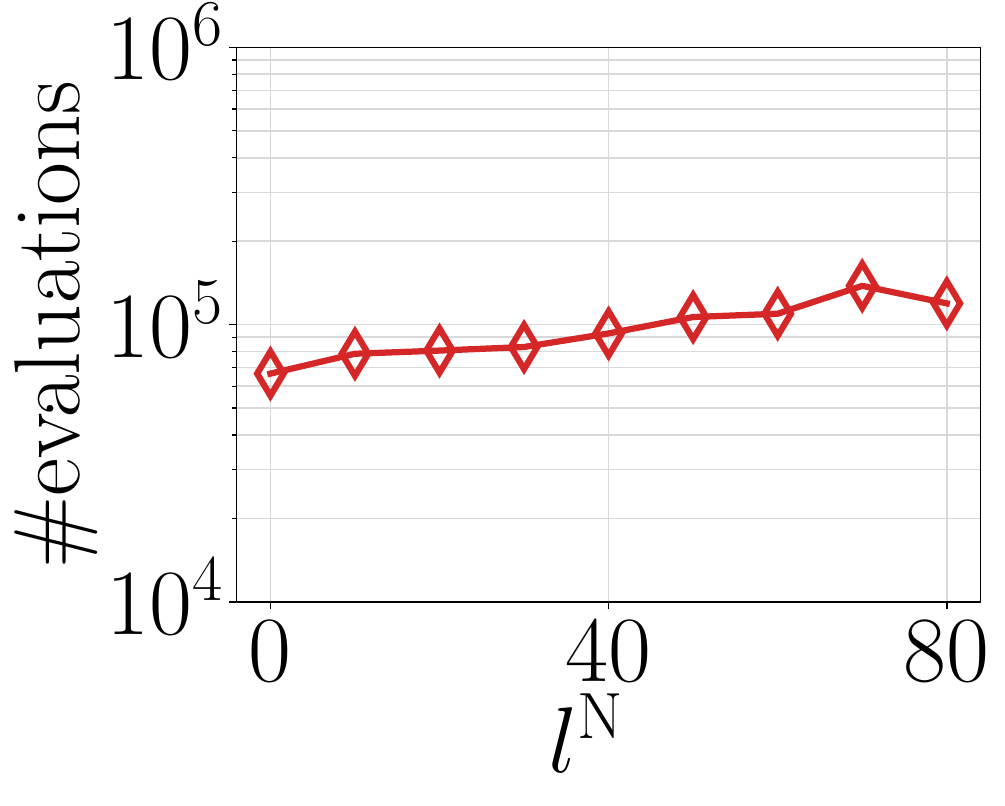}
 }
%\caption{The results of LS and LS-RN with $l^{\mathrm{N}} \in \{0, 10, \dots, 80\}$ on the ISSP using HV and $\epsilon$, where $l^{\mathrm{R}} = 80 - l^{\mathrm{N}}$ in LS-RN. The results for the linear PF, $d = 4, n = 5\,000$, and $ k = 100$ are shown.}
\caption{Results of LS-RN with $l^{\mathrm{N}} \in \{0, 10, \dots, 80\}$ and $l^{\mathrm{R}} = 80 - l^{\mathrm{N}}$ on the ISSP using HV and $\epsilon$. %The results for the linear PF, $d = 4, n = 5\,000$, and $k = 100$ are shown.
}
   \label{fig:list_size_ratio}
\end{figure}

Fig. \ref{fig:list_size_ratio} shows the results of LS-RN in terms of the average number of subset evaluations and computation time.
As seen from Fig. \ref{fig:list_size_ratio}, the setting of $l^{\mathrm{N}}$ does not significantly influence the two measurements of LS-RN.
This observation suggests that setting $l^{\mathrm{N}}$ and $l^{\mathrm{R}}$ to the same size as in this work is a reasonable first choice on an unseen real-world ISSP.

%\begin{tcolorbox}[title=Answers to RQ5, sharpish corners, top=2pt, bottom=2pt, left=4pt, right=4pt, boxrule=0.5pt]
%\textbf{Answers to RQ1:}
\begin{tcolorbox}[sharpish corners, top=2pt, bottom=2pt, left=4pt, right=4pt, boxrule=0.0pt, colback=black!4!white,leftrule=0.75mm,]
\textbf{Answers to RQ5:}
    Our results suggest that the nearest neighbor list size $l^{\mathrm{N}}$ in LS-N can control the trade-off between the quality of subsets and computational budget used in the search.
    %For both LS-N and LS-R, the use of a large $l$ value requires high computational cost but leads to improvement of the quality of subsets.
    For LS-N, the use of a large $l^{\mathrm{N}}$ value requires high computational cost but leads to improvement of the quality of subsets.
    We also observed that the performance of LS-RN can be improved by setting $l^{\mathrm{N}}$ and $l^{\mathrm{R}}$ to almost the same value when fixing the sum of $l^{\mathrm{N}}$ and $l^{\mathrm{R}}$ to $80$.
\end{tcolorbox}

\subsection{Further investigations}
\label{sec:further_analysis}

\subsubsection{Comparison to greedy search methods}

This paper has investigated the effectiveness of LS with the candidate list strategy by comparing it to conventional LS.
This section compares the proposed methods to advanced greedy search (GS) methods. 
Here, we consider the following four GS methods:
\begin{description}
    \item[\textbf{GS}] \hspace{0.3em} Greedy Subset Selection \cite{Lopez-IbanezKL11}
    \item[\textbf{GS-L}] \hspace{0.3em} Lazy GS for HV, IGD, IGD$^+$ \cite{ChenIS22}
    \item[\textbf{GS-A}] \hspace{0.3em} Approximated GS for HV \cite{ShangIC21}
    \item[\textbf{GS-AL}] \hspace{0.3em} Lazy Approximated GS for HV \cite{ShangIC21}
\end{description}

GS is the most basic and general method, which can be applied to the ISSP using any quality indicator.
In contrast, GS-A and GS-AL can be applied only to the ISSP using HV, and GS-L was designed only for the ISSP using HV, IGD, and IGD$^+$.
For this reason, we performed the comparison only on the ISSP using HV.
The \texttt{MATLAB} implementations of GS-L, GS-A, and GS-AL are publicly available.
However, \texttt{C++} code is generally faster than  \texttt{MATLAB} codes.
For fair comparison, we implemented all four GS methods in \texttt{C++} by referring to the original \texttt{MATLAB} implementations.
As in Section \ref{sec:analysis_l}, we used the following settings:  $d=4$, $n=5\,000$, $k=100$.

Table \ref{tab:vs-greedy} shows the results of the four GS methods and the four LS methods (LS, LS-N, LS-R, and LS-RN) on the ISSP using HV in terms of the quality of subsets.
Tables \ref{tab:vs-greedy} (a) and (b) show the results for the linear and discontinuous PFs, respectively.
We performed a single run for each GS method due to its deterministic nature.
For this reason, the standard deviation is not available for the four GS methods.
The right columns of Tables \ref{tab:vs-greedy} (a) and (b) show the relative error based on the results of GS.
Since GS-L is a faster version of GS, GS-L and GS find exactly the same subset.
For this reason, the relative errors for GS and GS-L are not shown in Table \ref{tab:vs-greedy}.
Tables \ref{suptab:vs-greedy_pf-linear} and \ref{suptab:vs-greedy_pf-disconnected} show the comparison of the two GS methods (GS and GS-L) and the four LS methods on the ISSPs using the other six quality indicators.
We do not explain Tables \ref{suptab:vs-greedy_pf-linear} and \ref{suptab:vs-greedy_pf-disconnected} in detail, but they are similar to the results in Table \ref{tab:vs-greedy}.

As seen from Tables \ref{tab:vs-greedy}(a) and (b), LS and LS-RN find better subsets than the four GS methods.
Although LS-N outperforms the four GS methods for the linear PF, LS-N is outperformed by GS and GS-L for the discontinuous PF.
This poor performance of LS-N for the discontinuous PF is consistent with the observation in Section \ref{sec:eval_Is_for_disconnected}.

\begin{figure}[t]
\captionsetup[subfloat]{farskip=2pt,captionskip=1pt}
\centering
\newcommand{\widthvar}{0.235}
\subfloat[linear PF]{
\includegraphics[width=0.35\textwidth]{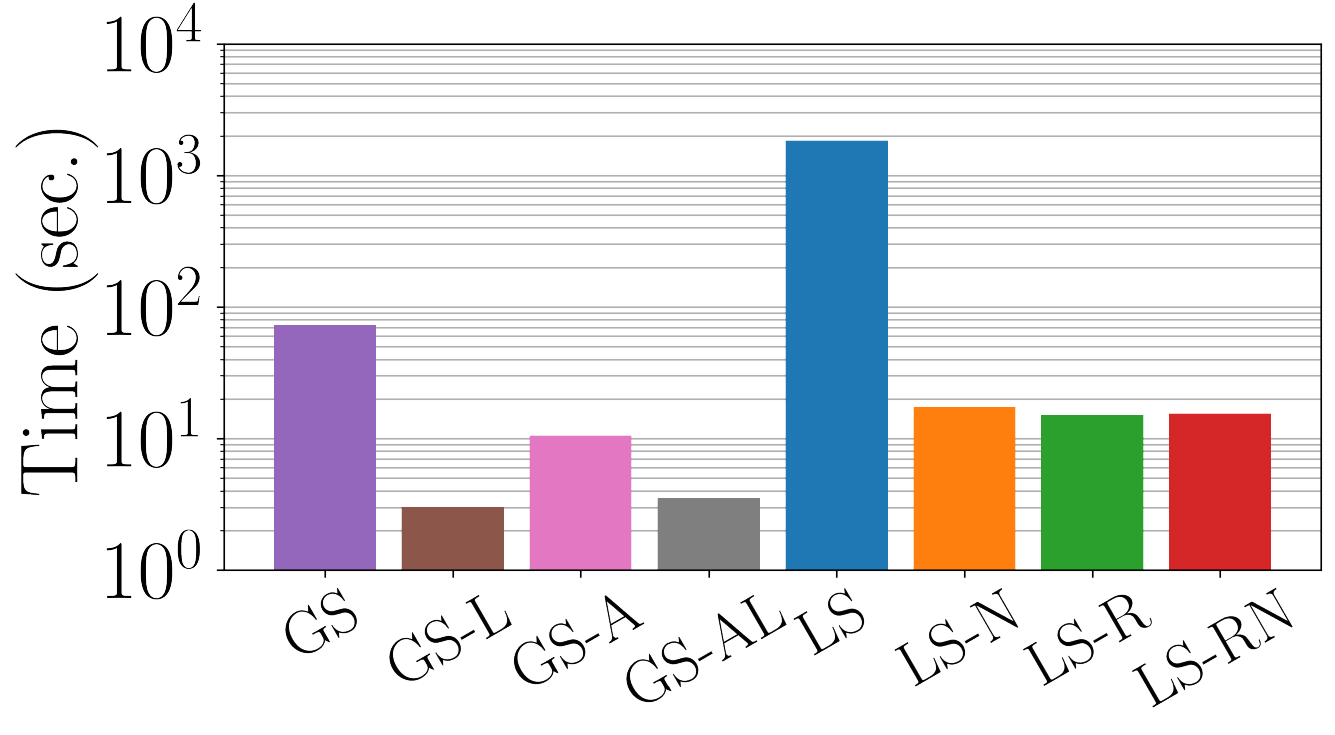}
} \\
\subfloat[discontinuous PF]{
\includegraphics[width=0.35\textwidth]{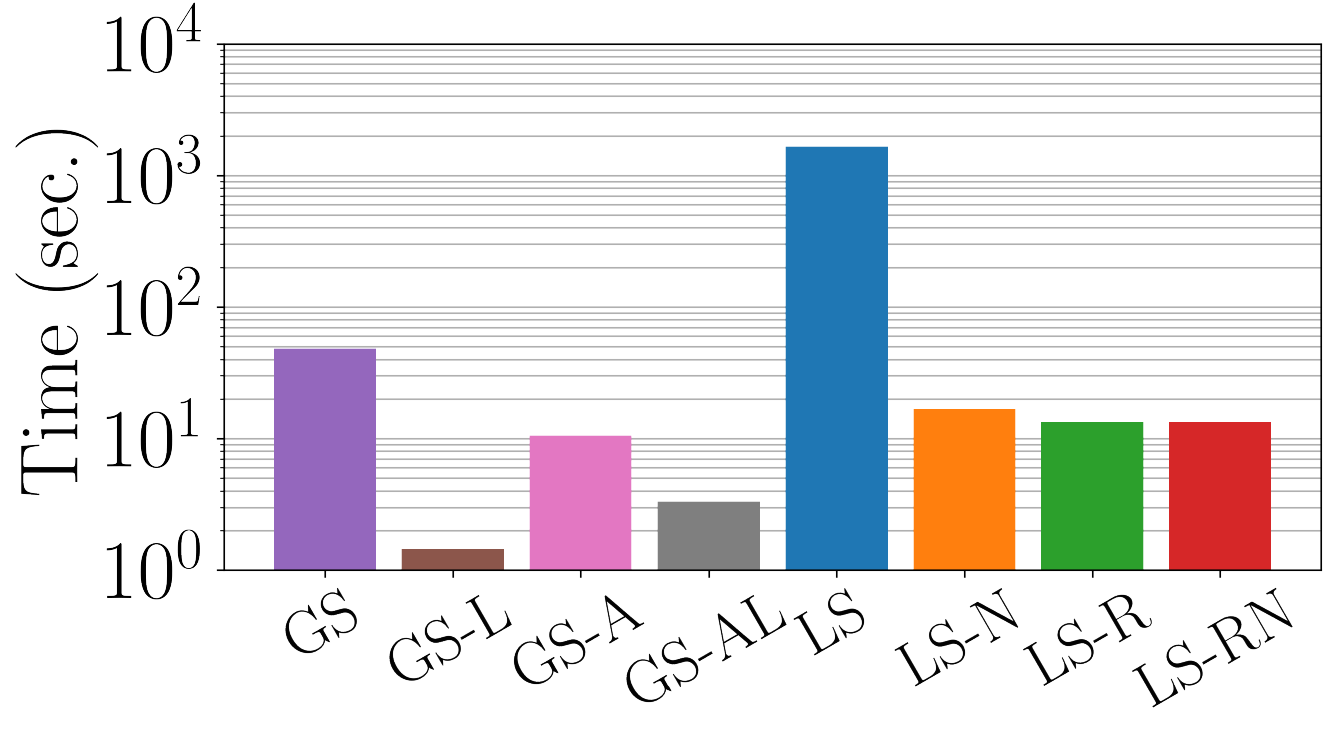}
}
\caption{Wall-clock time of the run of the four LS methods and the four GS methods on the ISSP instances using HV. }
\label{fig:greedy_time}
\end{figure}

Figs. \ref{fig:greedy_time} (a) and (b) show the wall-clock time of the run on the ISSPs with linear and discontinuous PFs, respectively.
As seen from Figs. \ref{fig:greedy_time} (a) and (b), as expected, conventional LS is the slowest in terms of the computation time.
GS-L is the fastest, followed by GS-AL.
Interestingly, LS-N, LS-R, and LS-RN are faster than conventional GS and slightly slower than GS-A.
In summary, our results showed the effectiveness of the LS-RN compared to the four GS methods.

\begin{table}[t]
    \centering
    \captionsetup[subfloat]{farskip=2pt,captionskip=1pt}
    \setlength{\tabcolsep}{5.0pt} % Default value: 6pt
    \renewcommand{\arraystretch}{0.9}
    \caption{\small
        Comparison of the four LS methods and the four GS methods on the ISSPs using HV. 
        The mean, standard deviation, and relative error based on GS of the quality indicator values of the subsets found by the other methods are shown. 
    }
  \label{tab:vs-greedy}  
{\footnotesize 
%\color{blue}
    \subfloat[linear PF] {
        \begin{tabular}{lccc}
        \toprule
         & mean & std. & rel. error \\
        \midrule
         GS    & \en{1.373}{+00} & NA              &                         \\
         GS-L  & \en{1.373}{+00} & NA              &                         \\
         GS-A  & \en{1.372}{+00} & NA              & \sgnPS\en{3.471}{-02}\% \\
         GS-AL & \en{1.372}{+00} & NA              & \sgnPS\en{3.471}{-02}\% \\
         LS    & \cellcolor{c1}\en{1.374}{+00} & \en{6.142}{-05} & \sgnMS\en{7.976}{-02}\% \\
         LS-N  & \cellcolor{c2}\en{1.374}{+00} & \en{7.190}{-05} & \sgnMS\en{7.522}{-02}\% \\
         LS-R  & \en{1.372}{+00} & \en{2.111}{-04} & \sgnPS\en{4.672}{-02}\% \\
         LS-RN & \en{1.374}{+00} & \en{8.525}{-05} & \sgnMS\en{7.213}{-02}\% \\
        \bottomrule
        \end{tabular}
    } \\
    \vspace{1em}
    \subfloat[discontinuous PF]{
        \begin{tabular}{lccc}
        \toprule
          & mean & std. & rel. error                                     \\
        \midrule
         GS    & \en{5.930}{-01} &  NA             &                         \\
         GS-L  & \en{5.930}{-01} &  NA             &                         \\
         GS-A  & \en{5.742}{-01} &  NA             & \sgnPS\en{3.159}{+00}\% \\
         GS-AL & \en{5.742}{-01} &  NA             & \sgnPS\en{3.159}{+00}\% \\
         LS    & \cellcolor{c1}\en{5.947}{-01} & \en{9.490}{-05} & \sgnMS\en{2.886}{-01}\% \\
         LS-N  & \en{5.889}{-01} & \en{1.520}{-03} & \sgnPS\en{6.768}{-01}\% \\
         LS-R  & \en{5.909}{-01} & \en{5.500}{-04} & \sgnPS\en{3.491}{-01}\% \\
         LS-RN & \cellcolor{c2}\en{5.942}{-01} & \en{1.649}{-04} & \sgnMS\en{2.174}{-01}\% \\
        \bottomrule
        \end{tabular}
    }
}
\end{table}

\subsubsection{Comparison on real-world problems}

This section investigates the performance of the proposed methods on 14 real-world problems (RE problems) \cite{TanabeI20}.
%The XXX RE problems \cite{TanabeI20} are considered.
Here, we consider selecting $k=100$ points from the approximated PF $P$ of size $n$ of each RE problem provided by \cite{TanabeI20}, where $n$ is $500 \times d$ in the RE problems, except for the RE3-4-6 problem.
Since $n$ of the RE3-4-6 problem is only 28, we do not use it.
For each RE problem, the approximated PF is normalized using the approximated ideal and nadir points.
%Unlike the experimental setup in other sections, 

Table \ref{tab:comparison_qi_reproblems} shows the comparison of LS, LS-N, LS-R, and LS-RN on the ISSP using HV and IGD$^+$ in terms of the quality of the subsets.
Table \ref{suptab:reproblems} shows the results for IGD, R2, NR2, $\epsilon$, and $s$-energy.

As seen from Table \ref{tab:comparison_qi_reproblems}(a), LS-RN is competitive to LS for HV.
The relative error of LS-RN is $0.0735\%$ even in the worst case.
As shown in Table \ref{tab:comparison_qi_reproblems}(b), LS-RN finds much worse subsets than LS for the RE2-3-5 and RE3-4-3 problems.
On these two problems, some points in the approximated PFs are very far from other points, where this kind of approximated PF can be viewed as a special case of the discontinuous PF.
In addition, such isolated points are unlikely to be included in the nearest neighbor lists of other points.
These are the reasons why LS-RN performs poorly on the ISSPs based on the RE2-3-5 and RE3-4-3 problems.

The results for IGD, R2, NR2, $s$-energy in Table \ref{suptab:reproblems} are consistent with the results in Section \ref{sec:eval_scalability}.
That is, LS-RN is competitive to LS for most quality indicators, but LS-RN is outperformed by LS for $\epsilon$.

\begin{figure}[t]
\captionsetup[subfloat]{farskip=2pt,captionskip=1pt}
\centering
\newcommand{\widthvar}{0.235}
\includegraphics[width=0.3\textwidth]{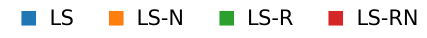}
\includegraphics[width=0.49\textwidth]{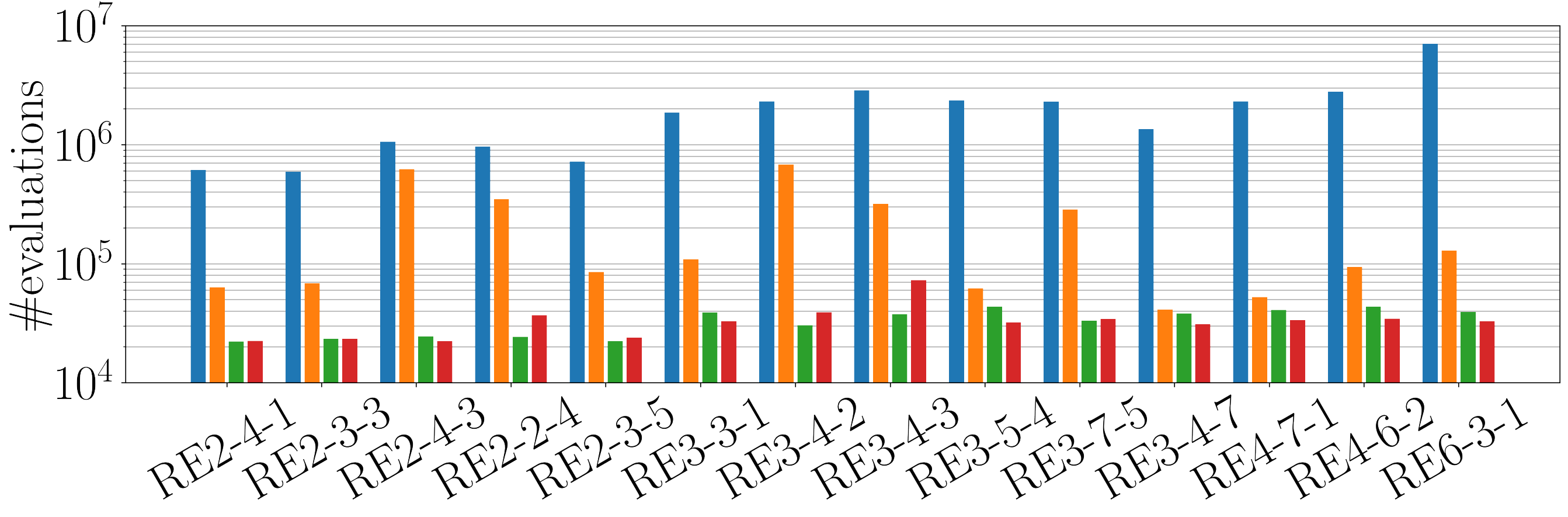}
 \caption{Number of subset evaluations by $\mathcal{I}$ for the four LS methods on the ISSPs using HV and the approximated PFs of the 14 real-world (RE) problems.}
\label{fig:reproblems_calls}
\end{figure}

Fig. \ref{fig:reproblems_calls} shows the number of subset evaluations for the four LS methods.
Similar to the results in Section \ref{sec:eval_scalability}, we can see that the number of subset evaluations for LS-R and LS-RN is significantly lower than that for LS.
In contrast, LS-N performs significantly worse than LS-R and LS-RN on the three ISSPs based on the RE2-4-3, RE2-2-4, RE3-4-2 problems in terms of the number of subset evaluations.
Fig. \ref{supfig:reproblems_calls} shows the results for IGD, IGD$^+$, R2, NR2, $\epsilon$, and $s$-energy.
Similar to Fig. \ref{fig:reproblems_calls}, Fig. \ref{supfig:reproblems_calls} shows that LS-RN performs significantly better than LS in terms of the number of subset evaluations.
In summary, the results on the ISSPs based on the RE problems show the effectiveness of LS-RN.

\begin{table*}[t]
\setlength{\tabcolsep}{5pt} % Default value: 6pt
  \renewcommand{\arraystretch}{0.9} 
\centering
  \caption{\small Comparison of the four LS methods on the ISSPs using HV and 
  the approximated PFs of  the 14 real-world (RE) problems. The mean and (relative error based on LS) of the quality indicator values of the subsets found by the four methods are shown.
  }
  \label{tab:comparison_qi_reproblems}
%{\scriptsize
{\footnotesize 
%\color{blue}
%{\small
%%%%%%%%%%%
\subfloat[HV]{
\begin{tabular}{ccccc}
\toprule
   Problem      & LS                       & LS-N                                                 & LS-R                                                 & LS-RN                                                \\
\midrule
 RE2-4-1 & \cellcolor{c1}\en{8.85}{-01} & \cellcolor{c3}  \en{8.84}{-01}\MSsymbol (\sgnPS\en{2.77}{-02}\%) & \cellcolor{c4}  \en{8.84}{-01}\MSsymbol (\sgnPS\en{2.79}{-02}\%) & \cellcolor{c2}  \en{8.85}{-01}\MSsymbol (\sgnPS\en{5.91}{-04}\%) \\
 RE2-3-3 & \cellcolor{c1}\en{7.59}{-01} & \cellcolor{c3}  \en{7.59}{-01}\MSsymbol (\sgnPS\en{3.48}{-02}\%) & \cellcolor{c4}  \en{7.59}{-01}\MSsymbol (\sgnPS\en{3.52}{-02}\%) & \cellcolor{c2}  \en{7.59}{-01}\MSsymbol (\sgnPS\en{1.08}{-03}\%) \\
 RE2-4-3 & \cellcolor{c1}\en{1.16}{+00} & \cellcolor{c4}  \en{1.16}{+00}\MSsymbol (\sgnPS\en{6.74}{-03}\%) & \cellcolor{c3}  \en{1.16}{+00}\MSsymbol (\sgnPS\en{1.71}{-03}\%) & \cellcolor{c2}  \en{1.16}{+00}\MSsymbol (\sgnPS\en{4.88}{-05}\%) \\
 RE2-2-4 & \cellcolor{c2}\en{1.17}{+00} & \cellcolor{c4}  \en{1.17}{+00}\MSsymbol (\sgnPS\en{3.90}{-02}\%) & \cellcolor{c3}  \en{1.17}{+00}\MSsymbol (\sgnPS\en{3.32}{-03}\%) & \cellcolor{c1}  \en{1.17}{+00}\PSsymbol (\sgnMS\en{2.72}{-04}\%) \\
 RE2-3-5 & \cellcolor{c1}\en{1.08}{+00} & \cellcolor{c4}  \en{7.74}{-01}\MSsymbol (\sgnPS\en{2.84}{+01}\%) & \cellcolor{c2}  \en{1.08}{+00}\EQsymbol (\sgnPS\en{5.67}{-06}\%) & \cellcolor{c3}  \en{1.08}{+00}\MSsymbol (\sgnPS\en{7.35}{-02}\%) \\
 RE3-3-1 & \cellcolor{c1}\en{1.33}{+00} & \cellcolor{c2}  \en{1.33}{+00}\MSsymbol (\sgnPS\en{8.24}{-06}\%) & \cellcolor{c3}  \en{1.33}{+00}\MSsymbol (\sgnPS\en{8.72}{-06}\%) & \cellcolor{c1}  \en{1.33}{+00}\EQsymbol (\sgnPS\en{0.00}{+00}\%) \\
 RE3-4-2 & \cellcolor{c1}\en{1.33}{+00} & \cellcolor{c4}  \en{1.33}{+00}\MSsymbol (\sgnPS\en{5.59}{-04}\%) & \cellcolor{c3}  \en{1.33}{+00}\MSsymbol (\sgnPS\en{4.80}{-05}\%) & \cellcolor{c2}  \en{1.33}{+00}\MSsymbol (\sgnPS\en{1.53}{-05}\%) \\
 RE3-4-3 & \cellcolor{c1}\en{1.31}{+00} & \cellcolor{c4}  \en{1.31}{+00}\MSsymbol (\sgnPS\en{3.04}{-02}\%) & \cellcolor{c2}  \en{1.31}{+00}\MSsymbol (\sgnPS\en{1.20}{-03}\%) & \cellcolor{c3}  \en{1.31}{+00}\MSsymbol (\sgnPS\en{3.00}{-03}\%) \\
 RE3-5-4 & \cellcolor{c1}\en{1.05}{+00} & \cellcolor{c4}  \en{1.04}{+00}\MSsymbol (\sgnPS\en{1.04}{-01}\%) & \cellcolor{c3}  \en{1.04}{+00}\MSsymbol (\sgnPS\en{4.40}{-02}\%) & \cellcolor{c2}  \en{1.05}{+00}\MSsymbol (\sgnPS\en{9.78}{-03}\%) \\
 RE3-7-5 & \cellcolor{c1}\en{1.31}{+00} & \cellcolor{c4}  \en{1.31}{+00}\MSsymbol (\sgnPS\en{8.42}{-03}\%) & \cellcolor{c3}  \en{1.31}{+00}\MSsymbol (\sgnPS\en{3.07}{-03}\%) & \cellcolor{c2}  \en{1.31}{+00}\MSsymbol (\sgnPS\en{5.66}{-04}\%) \\
 RE3-4-7 & \cellcolor{c1}\en{8.89}{-01} & \cellcolor{c3}  \en{8.89}{-01}\MSsymbol (\sgnPS\en{8.57}{-02}\%) & \cellcolor{c4}  \en{8.88}{-01}\MSsymbol (\sgnPS\en{1.45}{-01}\%) & \cellcolor{c2}  \en{8.89}{-01}\MSsymbol (\sgnPS\en{2.45}{-02}\%) \\
 RE4-7-1 & \cellcolor{c1}\en{8.65}{-01} & \cellcolor{c4}  \en{8.58}{-01}\MSsymbol (\sgnPS\en{8.04}{-01}\%) & \cellcolor{c3}  \en{8.63}{-01}\MSsymbol (\sgnPS\en{3.16}{-01}\%) & \cellcolor{c2}  \en{8.65}{-01}\MSsymbol (\sgnPS\en{6.62}{-02}\%) \\
 RE4-6-2 & \cellcolor{c1}\en{8.49}{-01} & \cellcolor{c4}  \en{8.46}{-01}\MSsymbol (\sgnPS\en{3.95}{-01}\%) & \cellcolor{c3}  \en{8.47}{-01}\MSsymbol (\sgnPS\en{2.05}{-01}\%) & \cellcolor{c2}  \en{8.49}{-01}\MSsymbol (\sgnPS\en{4.26}{-02}\%) \\
 RE6-3-1 & \cellcolor{c1}\en{1.50}{+00} & \cellcolor{c4}  \en{1.50}{+00}\MSsymbol (\sgnPS\en{9.82}{-02}\%) & \cellcolor{c3}  \en{1.50}{+00}\MSsymbol (\sgnPS\en{6.47}{-02}\%) & \cellcolor{c2}  \en{1.50}{+00}\MSsymbol (\sgnPS\en{2.01}{-02}\%) \\
\bottomrule
\end{tabular}
} \\
\subfloat[IGD$^+$]{
\begin{tabular}{ccccc}
\toprule
    Problem     & LS                       & LS-N                                                 & LS-R                                                 & LS-RN                                                \\
\midrule
 RE2-4-1 & \cellcolor{c1}\en{2.10}{-03} & \cellcolor{c4}  \en{2.28}{-03}\MSsymbol (\sgnPS\en{8.38}{+00}\%) & \cellcolor{c3}  \en{2.27}{-03}\MSsymbol (\sgnPS\en{7.75}{+00}\%) & \cellcolor{c2}  \en{2.12}{-03}\MSsymbol (\sgnPS\en{6.85}{-01}\%) \\
 RE2-3-3 & \cellcolor{c1}\en{2.07}{-03} & \cellcolor{c4}  \en{2.24}{-03}\MSsymbol (\sgnPS\en{8.04}{+00}\%) & \cellcolor{c3}  \en{2.22}{-03}\MSsymbol (\sgnPS\en{7.35}{+00}\%) & \cellcolor{c2}  \en{2.08}{-03}\MSsymbol (\sgnPS\en{5.24}{-01}\%) \\
 RE2-4-3 & \cellcolor{c1}\en{2.68}{-04} & \cellcolor{c4}  \en{3.34}{-04}\MSsymbol (\sgnPS\en{2.46}{+01}\%) & \cellcolor{c3}  \en{2.94}{-04}\MSsymbol (\sgnPS\en{9.66}{+00}\%) & \cellcolor{c2}  \en{2.71}{-04}\MSsymbol (\sgnPS\en{1.23}{+00}\%) \\
 RE2-2-4 & \cellcolor{c1}\en{4.11}{-04} & \cellcolor{c4}  \en{6.85}{-04}\MSsymbol (\sgnPS\en{6.68}{+01}\%) & \cellcolor{c3}  \en{4.39}{-04}\MSsymbol (\sgnPS\en{6.93}{+00}\%) & \cellcolor{c2}  \en{4.11}{-04}\EQsymbol (\sgnPS\en{5.27}{-02}\%) \\
 RE2-3-5 & \cellcolor{c1}\en{1.12}{-09} & \cellcolor{c4}  \en{2.25}{-03}\MSsymbol (\sgnPS\en{2.01}{+08}\%) & \cellcolor{c2}  \en{1.12}{-06}\MSsymbol (\sgnPS\en{1.00}{+05}\%) & \cellcolor{c3}  \en{1.22}{-05}\MSsymbol (\sgnPS\en{1.09}{+06}\%) \\
 RE3-3-1 & \cellcolor{c1}\en{7.93}{-08} & \cellcolor{c4}  \en{1.76}{-07}\MSsymbol (\sgnPS\en{1.21}{+02}\%) & \cellcolor{c3}  \en{9.84}{-08}\MSsymbol (\sgnPS\en{2.40}{+01}\%) & \cellcolor{c2}  \en{8.72}{-08}\MSsymbol (\sgnPS\en{9.96}{+00}\%) \\
 RE3-4-2 & \cellcolor{c1}\en{5.57}{-06} & \cellcolor{c4}  \en{9.86}{-06}\MSsymbol (\sgnPS\en{7.72}{+01}\%) & \cellcolor{c3}  \en{6.15}{-06}\MSsymbol (\sgnPS\en{1.06}{+01}\%) & \cellcolor{c2}  \en{5.71}{-06}\MSsymbol (\sgnPS\en{2.60}{+00}\%) \\
 RE3-4-3 & \cellcolor{c1}\en{4.51}{-09} & \cellcolor{c4}  \en{1.23}{-05}\MSsymbol (\sgnPS\en{2.72}{+05}\%) & \cellcolor{c2}  \en{1.82}{-07}\MSsymbol (\sgnPS\en{3.93}{+03}\%) & \cellcolor{c3}  \en{1.15}{-06}\MSsymbol (\sgnPS\en{2.53}{+04}\%) \\
 RE3-5-4 & \cellcolor{c1}\en{5.92}{-03} & \cellcolor{c3}  \en{6.19}{-03}\MSsymbol (\sgnPS\en{4.68}{+00}\%) & \cellcolor{c4}  \en{6.31}{-03}\MSsymbol (\sgnPS\en{6.58}{+00}\%) & \cellcolor{c2}  \en{5.96}{-03}\MSsymbol (\sgnPS\en{7.04}{-01}\%) \\
 RE3-7-5 & \cellcolor{c1}\en{3.05}{-04} & \cellcolor{c4}  \en{3.50}{-04}\MSsymbol (\sgnPS\en{1.47}{+01}\%) & \cellcolor{c3}  \en{3.32}{-04}\MSsymbol (\sgnPS\en{8.70}{+00}\%) & \cellcolor{c2}  \en{3.12}{-04}\MSsymbol (\sgnPS\en{2.23}{+00}\%) \\
 RE3-4-7 & \cellcolor{c1}\en{1.10}{-02} & \cellcolor{c3}  \en{1.13}{-02}\MSsymbol (\sgnPS\en{3.01}{+00}\%) & \cellcolor{c4}  \en{1.18}{-02}\MSsymbol (\sgnPS\en{7.21}{+00}\%) & \cellcolor{c2}  \en{1.11}{-02}\MSsymbol (\sgnPS\en{9.36}{-01}\%) \\
 RE4-7-1 & \cellcolor{c1}\en{2.24}{-02} & \cellcolor{c3}  \en{2.30}{-02}\MSsymbol (\sgnPS\en{2.64}{+00}\%) & \cellcolor{c4}  \en{2.37}{-02}\MSsymbol (\sgnPS\en{5.90}{+00}\%) & \cellcolor{c2}  \en{2.27}{-02}\MSsymbol (\sgnPS\en{1.29}{+00}\%) \\
 RE4-6-2 & \cellcolor{c1}\en{1.15}{-02} & \cellcolor{c3}  \en{1.22}{-02}\MSsymbol (\sgnPS\en{6.13}{+00}\%) & \cellcolor{c4}  \en{1.22}{-02}\MSsymbol (\sgnPS\en{6.78}{+00}\%) & \cellcolor{c2}  \en{1.16}{-02}\MSsymbol (\sgnPS\en{1.39}{+00}\%) \\
 RE6-3-1 & \cellcolor{c1}\en{8.56}{-03} & \cellcolor{c4}  \en{9.18}{-03}\MSsymbol (\sgnPS\en{7.22}{+00}\%) & \cellcolor{c3}  \en{9.18}{-03}\MSsymbol (\sgnPS\en{7.20}{+00}\%) & \cellcolor{c2}  \en{8.69}{-03}\MSsymbol (\sgnPS\en{1.49}{+00}\%) \\
\bottomrule
\end{tabular}
}
%%%%%%%%%%%
}
\end{table*}

\section{Conclusion}
\label{sec:conclusion}

First, this paper investigated the property of the ISSP (Section \ref{sec:analysis_cand_list}).
Our results showed that swapping two points far away from each other and close to each other is effective at early and later stages of LS, respectively.
Then, based on this observation, we proposed the candidate list strategy to speed up LS for the ISSP (Section \ref{sec:proposed_method}).
This paper introduced two types of candidate lists: the nearest neighbor  and random neighbor lists.
This paper also presented the idea of using the two neighbor lists sequentially.
We investigated the performance of the proposed candidate list strategy on the ISSPs with the seven quality indicators and various PFs (Section \ref{sec:results}).
The results for the continuous PF showed the effectiveness of the nearest neighbor list, except on the ISSP using $\epsilon$.
%and the two lists.
In contrast, our results indicated that the sequential use of the two neighbor lists is effective on the ISSP with both the continuous and discontinuous PFs.
We demonstrated that the sequential use of the two neighbor lists can drastically speed up LS in terms of both the number of subset evaluations and the computation time while maintaining the quality of subsets. % found.
We found that the proposed candidate list strategy is more effective as the point set size $n$ increases.

%Our results showed that the ISSP using R2 and $\epsilon$ have different properties compared to the ISSP using other quality indicators.
Our results showed that the properties of the ISSP using $\epsilon$ are different from those using other quality indicators (e.g., HV and IGD).
For this reason, the proposed candidate list strategy does not work well on the ISSP using $\epsilon$.
Based on these observations, we do not recommend applying the candidate list strategy to the real-world $\epsilon$ subset selection problem.
Future work should address this issue.
%Thus, further investigation is needed.

% Throughout this paper, we used the Euclidean distance to measure the distance between two points in the objective space.
% However, the effectiveness of the candidate list strategy may be improved by using other distance measures (e.g., the Manhattan distance) depending on the shape of the PF and the property of a quality indicator. % $\mathcal{I}$.
% An investigation of this concept is another future work.
% %It is also promising to design a method to adaptively adjust the size of the candidate list and its elements.

Our findings indicate the effectiveness of reducing the neighborhood size in LS, which has been previously  overlooked.
The proposed candidate list strategy can be a useful clue to  design an efficient LS method for the ISSP.
As described in Section \ref{sec:introduction}, the ISSP can be found in environmental selection in indicator-based EMO algorithms.
It is interesting to conduct a systematic benchmarking of subset selection methods for the ISSP for this setting.
Our results showed that the best candidate list strategy depends on various factors in the ISSP, including the distribution of points in $P$ and the choice of quality indicator.
Fitness landscape analysis of the ISSP is needed for a better understanding of the performance of subset selection methods.

\bibliographystyle{IEEEtran}
\bibliography{reference}

% \newpage

 

% \end{IEEEbiography}



%is a master course student with Graduate School of Environment and Information Sciences, Yokohama National University, Yokohama, Japan.

\begin{IEEEbiography}[{\includegraphics[width=1in,height=1.25in,keepaspectratio]{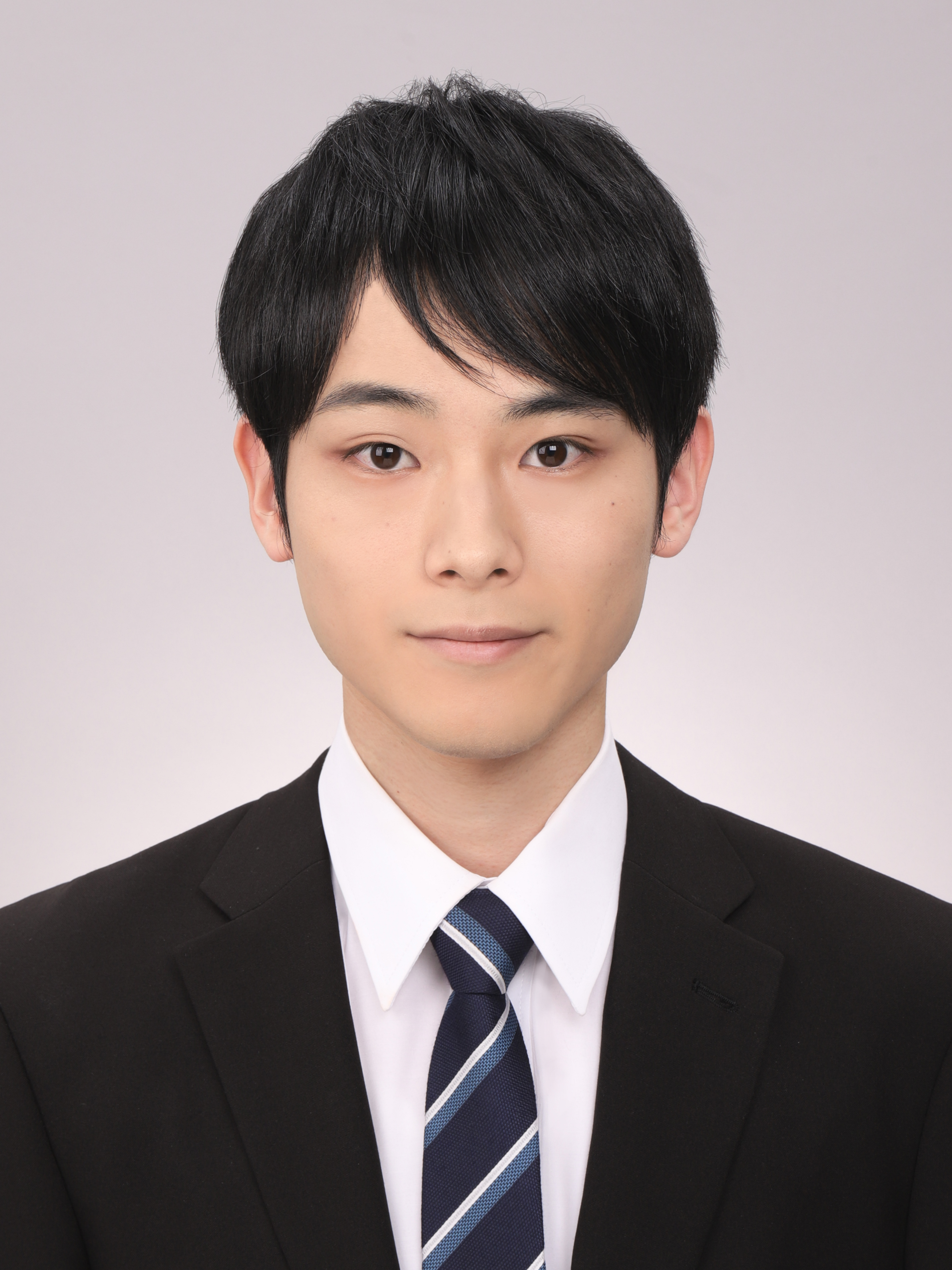}}]{Keisuke Korogi} is a master course student at Yokohama National University, Yokohama, Japan.
He received the B.S. degree in engineering from Yokohama National University in 2023.
His research interests include combinatorial optimization and multi-objective optimization.

  \end{IEEEbiography}
  
\begin{IEEEbiography}[{\includegraphics[width=1in,height=1.25in,keepaspectratio]{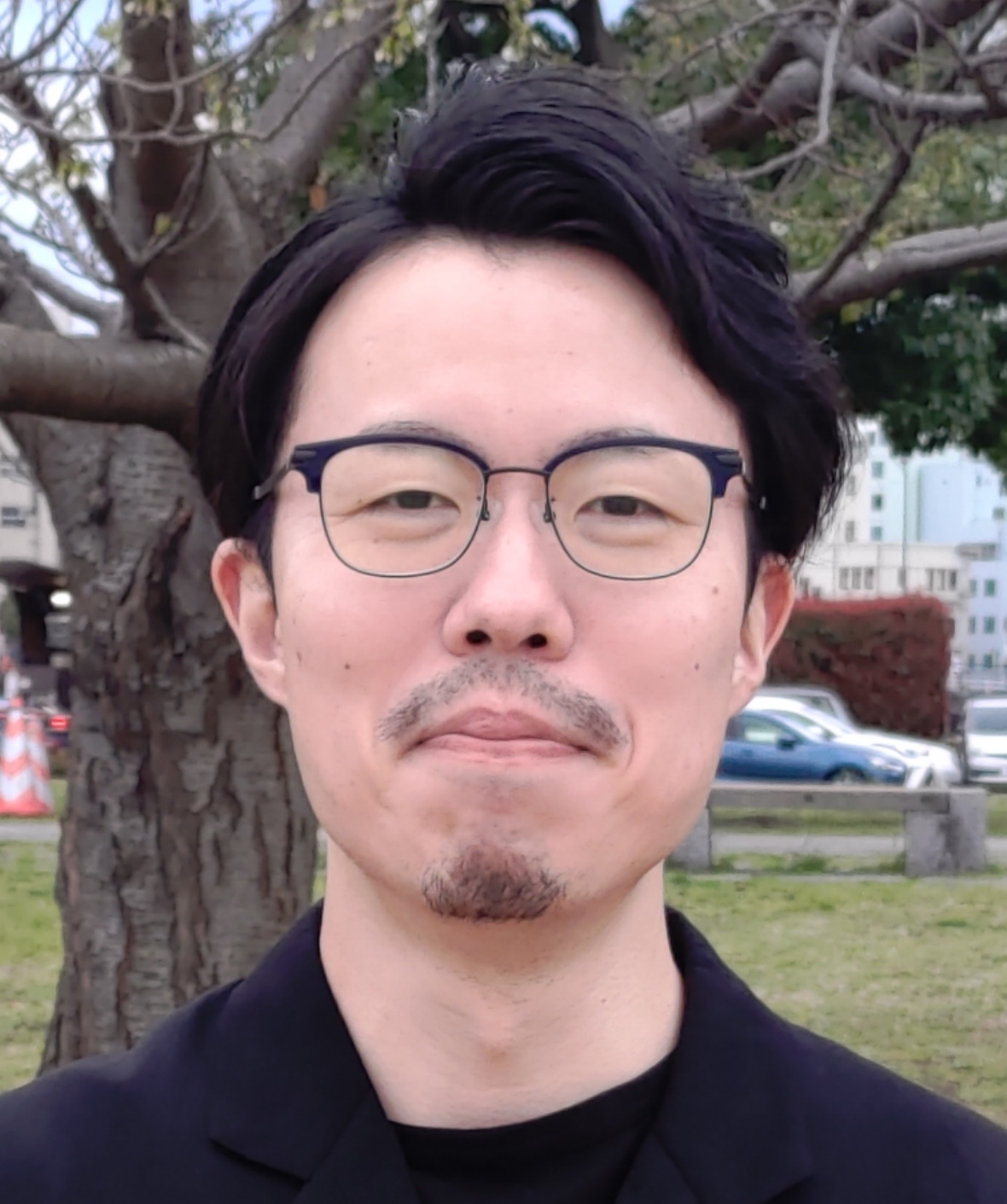}}]{Ryoji Tanabe}
is an Associate Professor at Yokohama National University, Yokohama, Japan (2019--).
Previously, he was a Research Assistant Professor at Southern University of Science and Technology, China (2017--2019).
He was also a Post-Doctoral Researcher at Japan Aerospace Exploration Agency, Japan (2016--2017).
He received the Ph.D. degree in Science from The University of Tokyo, Tokyo, Japan, in 2016.
His research interests include single- and multi-objective black-box optimization, benchmarking, and automated algorithm selection.
  \end{IEEEbiography}

\vfill

\end{document}